\pdfoutput=1

\documentclass[11pt]{article}

\usepackage[preprint]{acl}

\usepackage{times}
\usepackage{latexsym}

\usepackage[T1]{fontenc}

\usepackage[utf8]{inputenc}

\usepackage{microtype}

\usepackage{inconsolata}

\usepackage{graphicx}

\usepackage{multirow}
\usepackage{booktabs}
\usepackage{colortbl}
\usepackage{xcolor}
\usepackage{soul}
\usepackage{subcaption}
\usepackage{pifont}
\usepackage{tcolorbox}

\usepackage{graphicx}  
\usepackage{tikz}      

\usepackage{titlesec}   
\usepackage{titletoc}   

\usepackage{mathtools}
\usepackage{amsmath}
\usepackage{amssymb}  

\usepackage{algorithm}      
\usepackage{algpseudocode}  

\usepackage{cleveref}

\definecolor{mypurple}{rgb}{0.6, 0, 0.6} 
\definecolor{myred}{rgb}{0.7, 0.3, 0.0}
\definecolor{myblue}{rgb}{0.2, 0.3, 0.6}
\definecolor{mygreen}{rgb}{0.0, 0.4, 0.2}
\definecolor{mybrown}{rgb}{0.65, 0.16, 0.16} 

\definecolor{lightgreen}{rgb}{0.6, 1, 0.6}
\definecolor{lightred}{rgb}{1, 0.6, 0.6}
\definecolor{lightyellow}{rgb}{1, 1, 0.6}
\definecolor{lightblue}{rgb}{0.6, 0.8, 1}
\definecolor{lightgrey}{rgb}{0.93, 0.93, 0.93}

\newcommand{\hlgreen}[1]{{\sethlcolor{lightgreen}\hl{#1}}}
\newcommand{\hlred}[1]{{\sethlcolor{lightred}\hl{#1}}}

\title{DeepNote: Note-Centric Deep Retrieval-Augmented Generation}

\author{
 Ruobing Wang$^{1,2}$,
 Qingfei Zhao$^{1,2}$,
 Yukun Yan$^{3\dagger}$,
 Daren Zha$^{1}$,
 Yuxuan Chen$^{4}$,
 Shi Yu$^{3}$, \\
 \textbf{Zhenghao Liu$^{5}$,
 Yixuan Wang$^{3}$,
 Shuo Wang$^{3}$,
 Xu Han$^{3}$,
 Zhiyuan Liu$^{3}$,
 Maosong Sun$^{3\dagger}$}
\\
\\
 $^{1}$Institute of Information Engineering,
Chinese Academy of Sciences; \\
 $^{2}$School of Cyber Security,
University of Chinese Academy of Sciences; \\
 $^{3}$Department of Computer Science and Technology, Institute for AI,
Tsinghua University; \\
 $^{4}$South China University of Technology;
 \quad
 $^{5}$
Northeastern University \\
 \texttt{\{wangruobing\}@iie.ac.cn}
}

\begin{document}
\maketitle
\renewcommand{\thefootnote}{\fnsymbol{footnote}}
    \footnotetext[2]{Corresponding authors
    }
\renewcommand{\thefootnote}{\arabic{footnote}}
\begin{abstract}
Retrieval-Augmented Generation (RAG) mitigates factual errors and hallucinations in Large Language Models (LLMs) for question-answering (QA) by incorporating external knowledge.
However, existing adaptive RAG methods rely on LLMs to predict retrieval timing and directly use retrieved information for generation, often failing to reflect real information needs and fully leverage retrieved knowledge.
We develop \textbf{DeepNote}, an adaptive RAG framework that achieves in-depth and robust exploration of knowledge sources through note-centric adaptive retrieval.
DeepNote employs notes as carriers for refining and accumulating knowledge. During in-depth exploration, it uses these notes to determine retrieval timing, formulate retrieval queries, and iteratively assess knowledge growth, ultimately leveraging the best note for answer generation.
Extensive experiments and analyses demonstrate that DeepNote significantly outperforms all baselines (+10.2\% to +20.1\%) and exhibits the ability to gather knowledge with both high density and quality. Additionally, DPO further improves the performance of DeepNote.
The code and data are available at \url{https://github.com/thunlp/DeepNote}.

\end{abstract}

\section{Introduction}
Large Language Models (LLMs) (\citealp{DBLP:journals/corr/abs-2303-08774}; \citealp{DBLP:journals/corr/abs-2302-13971})
capture versatile knowledge~\citep{DBLP:journals/corr/abs-2403-17196} through billions of parameters, boosting performance in question-answering (QA) tasks.
\begin{figure}[ht!]
\includegraphics[width=\columnwidth]{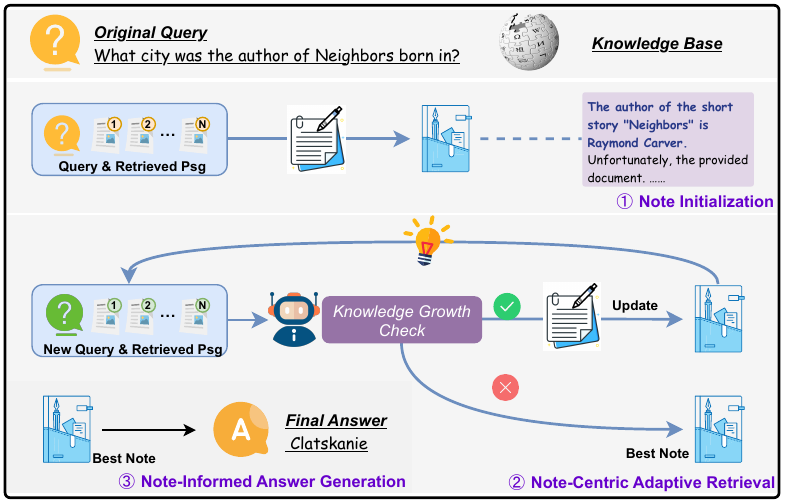}
\caption{\textbf{Illustration of DeepNote.} DeepNote fully integrates knowledge retrieved across multiple iterations using notes as the knowledge carrier and employs the best note to formulate retrieval decisions.
}
\label{fig:PrefRAG-Figure1}
\end{figure}
However, even state-of-the-art LLMs can encounter
hallucinations
\citep{DBLP:conf/cikm/ChenFYWFL0LX23}
and factual errors (\citealp{DBLP:conf/acl/MallenAZDKH23}; \citealp{DBLP:conf/emnlp/MinKLLYKIZH23}).
Retrieval-Augmented Generation (RAG)~\citep{DBLP:conf/nips/LewisPPPKGKLYR020}
is a widely used technique that leverages external non-parameterized knowledge resources to help LLMs push their inherent parameter knowledge boundaries to mitigate these issues.
However, Vanilla RAG usually fails to gather sufficient information for complex QA tasks w.r.t. long-form QA
(\citealp{DBLP:conf/emnlp/StelmakhLDC22};
~\citealp{DBLP:journals/corr/abs-2401-17043}), and multi-hop QA~\citep{DBLP:conf/emnlp/Yang0ZBCSM18}.
These complex QA tasks often involve broad or in-depth information retrieval needs, which may not be explicitly reflected in the initial query or easily fulfilled in a single retrieval attempt.

Recently, several works~(\citealp{DBLP:conf/emnlp/JiangXGSLDYCN23}; \citealp{DBLP:conf/iclr/AsaiWWSH24}) have proposed adaptive RAG (ARAG), which enables adaptively
capture more valuable knowledge for answering complex questions. 
Despite their success, they still have two limitations.
\textbf{\textit{First}}, each retrieval triggers an immediate generation. This approach may cause each output segment to reflect limited knowledge from a specific retrieval iteration, neglecting the integration and interaction of information across different retrieval iterations.
\textbf{\textit{Second}}, they leverage LLMs to actively predict retrieval timing; however, differences between the LLMs' internal cognition and the actual retrieval needs may lead to missing key knowledge.

To address them, we present \textbf{DeepNote}, an ARAG framework that utilizes notes as knowledge carriers to deeply and robustly explore knowledge bases for answering complex questions.
DeepNote comprises three key processes: note initialization, note-centric adaptive retrieval, and note-informed answer generation. \textbf{As depicted in Figure~\ref{fig:PrefRAG-Figure1}}, in the note initialization process, we first construct an initial note as the starting point for adaptive retrieval, treating it as the best note. In the note-centric adaptive retrieval process, we continuously use the best note to guide the system in making optimal forward retrieval decisions, and update the note with newly retrieved information from a view of knowledge growth. During each retrieval iteration, the model is encouraged to review and compare the latest note with the best note.
In the answer generation process, the system leverages the best note to generate comprehensive and accurate answers.

Extensive empirical experiments conducted on five datasets (including both complex and simple QA), demonstrate that DeepNote can effectively, robustly, and deeply explore knowledge bases. The overall performance of DeepNote significantly surpasses that of Vanilla RAG (up to +20.1\%) and a range of previous mainstream methods (up to +10.2\%), confirming its superiority.
We also develop an automated fine-tuning data construction pipeline and a training dataset, DNAlign, to enhance the model's instruction-following capabilities across multiple task stages and align with high-quality response preferences.
Empirical results on Llama3.1-8B and Qwen2.5-7B indicate that performing DPO with DNAlign further improves our framework's performance across all datasets. Additionally, multi-dimensional analysis demonstrates that our framework can gather high-quality and comprehensive information with higher knowledge density, while effectively balancing retrieval efficiency and performance.

\section{Related Work}
\subsection{Retrieval-Augmented Generation (RAG)}
Through knowledge augmentation, RAG~(\citealp{DBLP:journals/tacl/RamLDMSLS23};~\citealp{DBLP:conf/nips/LewisPPPKGKLYR020};~\citealp{DBLP:conf/icml/GuuLTPC20}) helps LLMs mitigate issues such as hallucinated outputs~(\citealp{DBLP:conf/cikm/ChenFYWFL0LX23}; \citealp{DBLP:conf/sigir-ap/ZucconKS23}), out-of-date knowledge and long-tail knowledge gaps~(\citealp{DBLP:journals/corr/abs-2301-00303}; \citealp{DBLP:conf/icml/KandpalDRWR23}), while extending LLMs beyond their knowledge boundaries~\cite{DBLP:conf/acl/YinSGWQH23}.
In QA tasks~(\citealp{DBLP:journals/corr/abs-2306-04136};~\citealp{DBLP:journals/tacl/SiriwardhanaWKWRN23};~\citealp{DBLP:conf/trec/Voorhees99}),
Vanilla RAG typically employs a retriever~\citep{DBLP:conf/emnlp/KarpukhinOMLWEC20} to fetch external knowledge from the corpus and incorporates it as text into the input space of LLMs, thereby enhancing the quality of answer.
Some previous methods~(\citealp{DBLP:journals/corr/abs-2311-09210}; \citealp{DBLP:journals/jmlr/IzacardLLHPSDJRG23}) adopt a single-step RAG method, where the retrieved passages are processed for knowledge refinement before generating the final answer.
However, they fail to directly retrieve sufficient information, especially in complex QA tasks.
One line of studies (\citealp{DBLP:conf/acl/TrivediBKS23}; \citealp{DBLP:conf/icml/BorgeaudMHCRM0L22}; \citealp{DBLP:journals/tacl/RamLDMSLS23}; \citealp{DBLP:conf/emnlp/PressZMSSL23};~\citealp{DBLP:journals/corr/abs-2403-05313}) attempt multi-step RAG during generation to alleviate this issue.
Another line of recent studies~(\citealp{DBLP:conf/emnlp/JiangXGSLDYCN23};~\citealp{DBLP:conf/iclr/YaoZYDSN023};~\citealp{DBLP:conf/iclr/AsaiWWSH24};~\citealp{DBLP:conf/naacl/JeongBCHP24}) propose ARAG systems, which can automatically determine ``\textit{when and what to retrieve}" via various feedbacks.
However, they may fail to actively predict true retrieval needs and timing through the LLM's parametric cognition and lack interaction with knowledge retrieved across multiple iterations.
Therefore, our work aims to establish a note-centric adaptive RAG that fully integrates knowledge retrieved across multiple iterations and uses the best note to guide retrieval decisions.

\subsection{Fine-Tuning for RAG}
Fine-tuning is widely used to improve the capabilities of LLM-augmented components in RAG systems~\citep{DBLP:journals/corr/abs-2401-08406}.
Early methods of fine-tuning to enhance LLM-based components in RAG primarily focused on training the retriever and the generator~(\citealp{DBLP:conf/acl/KeK00MB24};~\citealp{DBLP:conf/iclr/Lin0CSL00KSLZY24}).
Recent RAG methods have shifted toward modular designs~\citep{DBLP:journals/corr/abs-2312-10997}.
Particularly in complex QA tasks, adaptive RAG often requires base models to follow intricate instructions~(\citealp{yin2023llm};~\citealp{DBLP:conf/iclr/XuSZG0FTLJ24}) to enable the functionality of diverse components~\citep{DBLP:conf/iclr/AsaiWWSH24}.
Classic alignment training methods include supervised fine-tuning (SFT) and reinforcement learning from human feedback (RLHF). However, SFT lacks negative feedback and is prone to overfitting. Recently, ~\citeauthor{DBLP:conf/nips/RafailovSMMEF23} proposed a more efficient reinforcement learning algorithm, direct preference optimization (DPO), which aligns response preferences and enhances the model's instruction-following ability by learning the differences between positive and negative sample pairs. In our work, we focus on using DPO to enhance the model's capability in multiple processes.
\begin{figure*}[t]
\centering
\includegraphics[width=\textwidth]{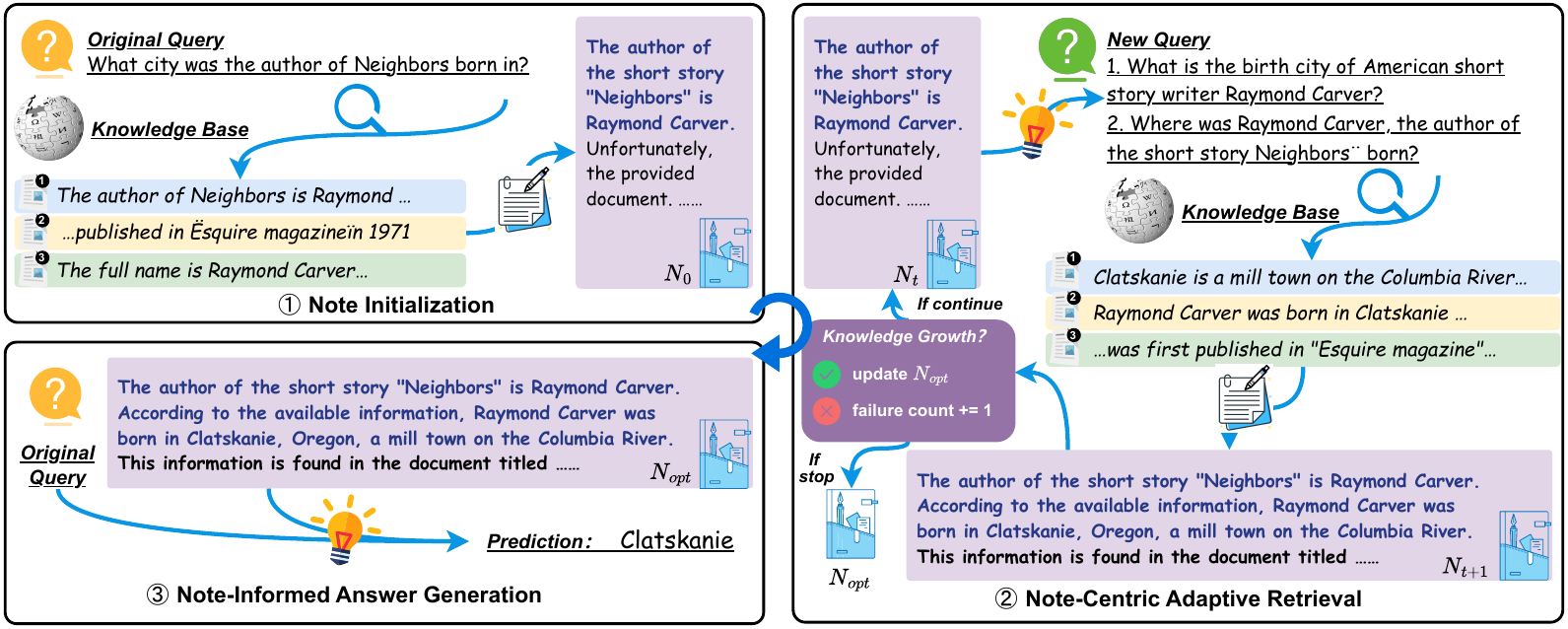}
\caption{\textbf{Overview of DeepNote.} DeepNote consists of three processes: Note Initialization, Note-Centric Adaptive Retrieval, and Note-Informed Answer Generation. We employ a note-centric strategy to formulate retrieval decisions (including "when and what to retrieve"), accumulate knowledge, and generate answers.}
\label{fig:DeepNote}
\end{figure*}
\section{Methodology}
In this section, we first introduce three key processes (\S~\ref{subsec:Note initialization}, \S~\ref{subsec:Note-centric Adaptive Retrieval}, and \S~\ref{subsec:Note-Informed Answer Generation}) of \textbf{DeepNote},  with an overview illustrated in Figure~\ref{fig:DeepNote}.
We then introduce our training dataset DNAlign, its automated construction pipeline~(\S~\ref{subsec:Data Construction for Training}), and the training process~(\S~\ref{subsec:Preference Optimization through DPO}).

\subsection{Note Initialization}
\label{subsec:Note initialization}
To enhance the model's awareness of useful knowledge while minimizing noise during adaptive exploration, we introduce a note as the knowledge carrier.
We start with an original query $q_{0}$, then retrieve top-$k$ passages $P_{k,0}=\left \{p_{1},p_{2},\dots,p_{k}\right \}$ as references.
We observe that since the system fails to foresee the characteristics and aspects of the retrieved knowledge, a fine-grained note construction approach, where notes are strictly summarized from predefined aspects or domains, often leads to misalignment between the collected knowledge and the actual relevant information.
Therefore, we delegate reasoning and decision-making entirely to the LLM, providing only the highest-level objective to facilitate its flexible and comprehensive collection of knowledge
that supports answering or reasoning about the $q_{0}$.
We now formalize this process:
\begin{equation}
N_{0}\sim \text{LLM}_{\text{Init}}(\text{Instruct}_{\text{Init}},q_{0}\Vert P_{k,0})
\end{equation}
where we use the prompt template $\text{Instruct}_{\text{Init}}$ to instruct LLM to generate the initial note $N_{0}$.
The $\text{LLM}_{\text{Init}}(\cdot)$ denotes the backbone model used in the note initialization.

\subsection{Note-Centric Adaptive Retrieval}
\label{subsec:Note-centric Adaptive Retrieval}
To effectively and deeply explore the unknown semantic space of the corpus, we develop a note-centric, three-stage adaptive retrieval process.

\noindent\textbf{Query Refinement}\quad
In this stage, we leverage the distilled knowledge stored in the note to formulate the new query $q_{t}$ for further retrieval.
Specifically, we only have the initial note $N_{0}$ as a reference after the note initialization process $\tau_{0}$.
Thus, in iteration $\tau_{1}$, we regard 
$N_{0}$ as $N_{\text{Opt}}$.
In each iteration $\tau_{t}$, the input consists of the $q_{0}$, the list of previously generated queries, and the best note so far.
Among them, the best note\footnotemark[1] so far refers to the note selected as the best choice by comparing it with the previous iteration’s best note, denoted as $N_{\text{Opt}}$.
\footnotetext[1]{The generation of the best note $N_{\text{Opt}}$ is a recursive process,
where $N_{\text{Opt}}$ in the current iteration $\tau_{t}$ is defined using the best note $N_{\text{Opt}}$ from the iteration $\tau_{t-1}$ along with other variables.
Therefore, we provide a detailed definition of $N_{\text{Opt}}$ in the adaptive retrieval decision stage in \S~\ref{subsec:Note-centric Adaptive Retrieval}.}
This recursive comparison process resembles how humans integrate and learn new knowledge, as they tend to formulate new questions based on their existing optimal understanding.
Additionally, the list of previously generated queries includes new queries generated in all previous iterations $\tau_{<t}$, denoted as $Q^{\text{Pre}}_{t} \coloneqq \{q_1, q_2, \ldots, q_{t-1}\}$.
This design stems from our observation that the LLM tends to repeatedly generate highly similar queries if issues raised in earlier iterations remain unresolved.
To prevent the system from getting trapped in localized exploration, we introduce $Q^{\text{Pre}}$ to eliminate the generation of redundant or ineffective queries.
To sum up, the process can be formalized as follows:
\begin{equation}
q_{t}\sim\text{LLM}_{\text{QR}}(\text{Instruct}_{\text{QR}}, q_{0}\Vert N_{\text{Opt}}\Vert Q^{\text{Pre}}_{t})
    \label{eq:new query}
\end{equation}
Equation~(\ref{eq:new query}) clearly illustrates the process of generating new queries $q_{t}$ for further retrieval in iteration $\tau_{t}$, where $t\ge 1$.
The $\text{Instruct}_{\text{QR}}$ and $\text{LLM}_{\text{QR}}(\cdot)$ represent the prompt template and backbone model of the process in the query refinement stage. 

\noindent\textbf{Knowledge Accumulation}\quad
Our goal is to leverage new queries to explore potential query-relevant semantic subspaces within the corpus for knowledge accumulation.
We guide the LLM from a view of "\textit{how to foster stable and effective knowledge growth}" for complex information collection, refinement, and updating.
Specifically, we first use a new query $q_ {t}$ to retrieve top-$k$ passages $P_{k,t}$.
Next, we construct a note-updating workflow informed by multi-dimensional guidance.
\begin{equation}
N_{t}\sim\text{LLM}_{\text{KA}}(\text{Instruct}_{\text{KA}}, q_{0}\Vert N_{\text{Opt}}\Vert P_{k,t})
    \label{eq:Update}
\end{equation}
Equation~(\ref{eq:Update}) presents the process of note updating for knowledge accumulation using the model $\text{LLM}_{\text{KA}}$.
The $\text{Instruct}_{\text{KA}}$ denotes the prompt template, where we provide a detailed workflow.
In this workflow, we require that the knowledge incorporated into updated notes $N_{t}$ remains faithful to the retrieved passages $P_{k,t}$, meaning that the collected information should follow their style and, whenever possible, use direct excerpts.
This strategy aims to minimize the introduction of parametric knowledge over deep iterative processes, which could otherwise lead to knowledge bias after multiple iterations.
Furthermore, we enforce knowledge validity, ensuring that the collected knowledge contributes to solving the $q_{0}$. This allows the system to remain focused on the $q_{0}$ throughout multiple iterations, mitigating noise interference.
Additionally, to avoid the accumulation of redundant knowledge over iterations, we perform a semantic review to assess whether the collected information is already present in $N_{\text{Opt}}$.

\noindent\textbf{Adaptive Retrieval Decision}\quad
An intuition is that retrieving relevant information from a corpus has an inherent upper bound.
Moreover, we observe that the model, limited by its ability to follow instructions, does not always accumulate knowledge effectively and may occasionally introduce noise.
Therefore, we focus on two key aspects in this stage. First, we determine whether to employ the next retrieval iteration by assessing whether the note updating leads to knowledge gain, achieving the adaptive retrieval process. Second, we identify the best note so far to improve retrieval decision, new query generation, and note update in the next iteration $\tau_{t+1}$.
Specifically, we first guide the LLM to carefully review the content of the updated note $N_{t}$ and the best note so far $N_{\text{Opt}}$, then assess their knowledge to get a status value $V_{t}$:
\begin{equation}
\begin{aligned}
V_{t}\sim \text{LLM}&_{\text{ARD}}(\text{Instruct}_{\text{ARD}},q_{0}\Vert N_{\text{Opt}}),\\
&V_{t}\in\left \{ \texttt{True,False} \right \}
\end{aligned}
\label{eq:Compare1}
\end{equation}
where the $\text{LLM}_{\text{ARD}}$ and the $\text{Instruct}_{\text{ARD}}$ refer to the backbone model and the prompt template in the assessment process.
In the assessment workflow, we have also designed multi-dimensional evaluation criteria, including
1) whether the content contains key information directly related to $q_{0}$, 2) whether the content has multiple aspects and sufficient details, and 3) whether the content is practical enough.
Next, we adopt $V_{t}$ to determine whether to update the best note $N_{\text{Opt}}$.
If $V_{t}=\texttt{True}$, the updated note $N_{t}$ generated in the current iteration $\tau_{t}$ is designated as the best note $N_{\text{Opt}}$.
If $V_{t}=\texttt{False}$, the content of the best note $N_{\text{Opt}}$ remains unchanged.

\subsection{Note-Informed Answer Generation}
\label{subsec:Note-Informed Answer Generation}
\noindent\textbf{Adaptive Stop Condition}\quad
If the LLM determines that an updated note $N_{t}$ is inferior to the best note $N_{\text{Opt}}$, the update is considered unsuccessful. Such a failed update indicates that the exploration has not contributed new knowledge and suggests low marginal returns from further retrieval. Based on this, we define two stopping criteria for adaptive retrieval. First, we set a threshold for the number of failure updates, termed "max failure"; once this limit is reached, the iteration terminates. Second, we impose a maximum number of iterations, termed "max step".

\noindent\textbf{Task-Oriented Generation}\quad
After terminating the iteration $\tau_{t}$, we input the $N_{\text{Opt}}$ from the final iteration along with the $q_{0}$ into the LLM to generate the final answer.
Due to the varying output styles of different question-answering tasks, we have customized generation instructions for each task (more details in Appendix~\ref{Appendix B.1: Prompt for Inference}).
\begin{equation}
\alpha\sim \text{LLM}_{\text{Ans}}(\text{Instruct}_{\text{Ans}},q_{0}\Vert N_{\text{Opt}})
\label{eq:answer}
\end{equation}
In Equation~(\ref{eq:answer}), $\text{Instruct}_{\text{Ans}}$ denotes the prompt template set of the task-oriented generation process, which includes a series of task-oriented instructions, and $\text{LLM}_{\text{Ans}}$ indicates the backbone model in task-oriented generation stage.

\subsection{Data Construction for Training}
\label{subsec:Data Construction for Training}
Previous studies have found that using state-of-the-art LLMs for automated sample annotation has high human correspondence~(\citealp{DBLP:conf/emnlp/LiuIXWXZ23};~\citealp{DBLP:conf/naacl/FuNJ024}). Therefore, we employ GPT-4o-mini for automated annotation for DPO training.
We developed an automated data construction pipeline and carefully curated a small but high-quality training dataset for multi-task training, named \textbf{DNAlign}.
This dataset $\mathcal{D}$ stems from four key task stages, including note initialization data $\mathcal{D}_{\text{Init}}$, query refinement data $\mathcal{D}_{\text{QR}}$, knowledge accumulation data $\mathcal{D}_{\text{KA}}$, and task-oriented generation data $\mathcal{D}_{\text{Ans}}$, which can be formulated as $\left \{ x,y^{+},y^{-} \right \} \sim\mathcal{D} =\left \langle \mathcal{D}_{\text{Init}},\mathcal{D}_{\text{QR}}, \mathcal{D}_{\text{KA}},\mathcal{D}_{\text{Ans}} \right \rangle$.
We provide a detailed description of the construction process and the statistics of DNAlign in Appendix~\ref{Appendix:Details of Training Dataset Construction}.

\subsection{Preference Optimization through DPO}
\label{subsec:Preference Optimization through DPO}
To enhance the instruction-following ability of the models used in each stage of DeepNote and align with higher-quality response preferences, we employ DPO to train the backbone models used in multiple stages, marked as $\text{M}_{\text{DN}}$. The training data comes from DNAlign.
\begin{multline}
\mathcal{L}_{DPO}(M^{\theta}_{\text{DN}};M^{ref}_{\text{DN}})=-\mathbb{E}_{\left \{ x,y^{+},y^{-} \right \}\sim \mathcal{D}} [log\sigma \\ [\beta log\frac{M^{\theta}_{\text{DN}}(y^{+}|x)}{M^{ref}_{\text{DN}}(y^{+}|x)}-\beta log\frac{M^{\theta}_{\text{DN}}(y^{-}|x)}{M^{ref}_{\text{DN}}(y^{-}|x)}]]
\label{eq:DPO}
\end{multline}
Equation~(\ref{eq:DPO}) defines the training objective, where $M^{\theta}_{\text{DN}}$ and $M^{ref}_{\text{DN}}$ represent  trained model and reference model frozen during training.
\begin{table*}[htbp]
  \centering
    \setlength{\tabcolsep}{1.5pt}
    \fontsize{8}{8}\selectfont
    \begin{tabular}{lcccccccccccccccc}
    \toprule
    \multirow{3}[3]{*}{\textbf{Methods \& LLMs}} & \multicolumn{12}{c}{\bf Multi-hop} & \multicolumn{3}{c}{\bf Long-form} & \multicolumn{1}{c}{\bf Short-form} \\
\cmidrule(lr){2-13}\cmidrule(lr){14-16}\cmidrule(lr){17-17}& \multicolumn{4}{c}{\bf HotpotQA} & \multicolumn{4}{c}{\bf 2WikiMQA} & \multicolumn{4}{c}{\bf MusiQue}  & \multicolumn{3}{c}{\bf ASQA} & \multicolumn{1}{c}{\bf StrategyQA}
\\
\cmidrule(lr){2-13}\cmidrule(lr){6-9}\cmidrule(lr){10-13}\cmidrule(lr){14-16}\cmidrule(lr){17-17} \multicolumn{1}{l}{} & \multicolumn{1}{c}{acc.} & \multicolumn{1}{c}{f1} & \multicolumn{1}{c}{em} & \multicolumn{1}{c}{avg.} & \multicolumn{1}{c}{acc.} & \multicolumn{1}{c}{f1} & \multicolumn{1}{c}{em} & \multicolumn{1}{c}{avg.} & \multicolumn{1}{c}{acc.} & \multicolumn{1}{c}{f1} & \multicolumn{1}{c}{em} & \multicolumn{1}{c}{avg.} & \multicolumn{1}{c}{str-em} & \multicolumn{1}{c}{str-hit} & \multicolumn{1}{c}{avg.} & \multicolumn{1}{c}{acc.}
\\
\midrule
\rowcolor[rgb]{ .851,  .851,  .851}\multicolumn{17}{c}{\textbf{\textit{LLMs without Retrieval}}}
\\
\multicolumn{1}{l}{Qwen2.5-7B-Instruct} & \multicolumn{1}{c}{19.2} & \multicolumn{1}{c}{25.7} & \multicolumn{1}{c}{18.2} & \multicolumn{1}{c}{21.0} & \multicolumn{1}{c}{25.0} & \multicolumn{1}{c}{29.0} & \multicolumn{1}{c}{24.2} & \multicolumn{1}{c}{26.1} & \multicolumn{1}{c}{2.8} & \multicolumn{1}{c}{9.8} & \multicolumn{1}{c}{2.4} & \multicolumn{1}{c}{5.0} & \multicolumn{1}{c}{24.9} & \multicolumn{1}{c}{8.3} & \multicolumn{1}{c}{12.7} & \multicolumn{1}{c}{67.2}
\\
\multicolumn{1}{l}{Llama3.1-8B-Instruct} & \multicolumn{1}{c}{22.6} & \multicolumn{1}{c}{27.7} & \multicolumn{1}{c}{22.0} & \multicolumn{1}{c}{24.1} & \multicolumn{1}{c}{29.2} & \multicolumn{1}{c}{32.5} & \multicolumn{1}{c}{28.2} & \multicolumn{1}{c}{30.0} & \multicolumn{1}{c}{3.2} & \multicolumn{1}{c}{9.2} & \multicolumn{1}{c}{3.2} & \multicolumn{1}{c}{5.2} & \multicolumn{1}{c}{32.4} & \multicolumn{1}{c}{10.2} & \multicolumn{1}{c}{15.9} & \multicolumn{1}{c}{69.2}
\\
GPT-4o-mini & \multicolumn{1}{c}{31.8} & \multicolumn{1}{c}{39.3} & \multicolumn{1}{c}{29.8} & \multicolumn{1}{c}{33.6} & \multicolumn{1}{c}{30.6} & \multicolumn{1}{c}{33.9} & \multicolumn{1}{c}{27.2} & \multicolumn{1}{c}{30.6} & \multicolumn{1}{c}{7.8} & \multicolumn{1}{c}{16.0} & \multicolumn{1}{c}{5.8} & \multicolumn{1}{c}{9.9} & \multicolumn{1}{c}{34.1} & \multicolumn{1}{c}{9.4} & \multicolumn{1}{c}{17.8} & \multicolumn{1}{c}{73.8}
\\
\multicolumn{1}{l}{Llama3.1-70B-Instruct} & \multicolumn{1}{c}{32.2} & \multicolumn{1}{c}{40.9} & \multicolumn{1}{c}{30.8} & \multicolumn{1}{c}{34.6} & \multicolumn{1}{c}{34.8} & \multicolumn{1}{c}{38.0} & \multicolumn{1}{c}{31.4} & \multicolumn{1}{c}{34.7} & \multicolumn{1}{c}{7.4} & \multicolumn{1}{c}{13.0} & \multicolumn{1}{c}{5.6} & \multicolumn{1}{c}{8.7} & \multicolumn{1}{c}{41.4} & \multicolumn{1}{c}{14.4} & \multicolumn{1}{c}{21.5} & \multicolumn{1}{c}{75.2}
\\
\midrule
\rowcolor[rgb]{ .851,  .851,  .851}\multicolumn{17}{c}{\textbf{\textit{ Vanilla RAG (Vanilla)}}}
\\
\multicolumn{1}{l}{Qwen2.5-7B-Instruct} & \multicolumn{1}{c}{37.4} & \multicolumn{1}{c}{44.0} & \multicolumn{1}{c}{33.6} & \multicolumn{1}{c}{38.3} & \multicolumn{1}{c}{33.2} & \multicolumn{1}{c}{36.3} & \multicolumn{1}{c}{31.8} & \multicolumn{1}{c}{33.8} & \multicolumn{1}{c}{7.6}  & \multicolumn{1}{c}{12.5} & \multicolumn{1}{c}{5.6}  & \multicolumn{1}{c}{8.6} & \multicolumn{1}{c}{42.1} & \multicolumn{1}{c}{15.9} & \multicolumn{1}{c}{22.2} & \multicolumn{1}{c}{68.4}
\\
\multicolumn{1}{l}{Llama3.1-8B-Instruct} & \multicolumn{1}{c}{37.6} & \multicolumn{1}{c}{46.4} & \multicolumn{1}{c}{35.0} & \multicolumn{1}{c}{39.7} & \multicolumn{1}{c}{33.4} & \multicolumn{1}{c}{36.3} & \multicolumn{1}{c}{32.0} & \multicolumn{1}{c}{33.9} & \multicolumn{1}{c}{6.8} & \multicolumn{1}{c}{12.1} & \multicolumn{1}{c}{6.0} & \multicolumn{1}{c}{8.3} & \multicolumn{1}{c}{39.3} & \multicolumn{1}{c}{13.3} & \multicolumn{1}{c}{20.3} & \multicolumn{1}{c}{71.4}
\\
\multicolumn{1}{l}{GPT-4o-mini} & \multicolumn{1}{c}{44.0} & \multicolumn{1}{c}{52.2} & \multicolumn{1}{c}{40.0} & \multicolumn{1}{c}{45.4} & \multicolumn{1}{c}{40.4} & \multicolumn{1}{c}{44.4} & \multicolumn{1}{c}{39.2} & \multicolumn{1}{c}{41.3} & \multicolumn{1}{c}{10.6} & \multicolumn{1}{c}{17.3} & \multicolumn{1}{c}{7.6} & \multicolumn{1}{c}{11.8} & \multicolumn{1}{c}{44.3} & \multicolumn{1}{c}{17.5} & \multicolumn{1}{c}{24.5} & \multicolumn{1}{c}{71.2}
\\
\multicolumn{1}{l}{Llama3.1-70B-Instruct} & \multicolumn{1}{c}{44.6} & \multicolumn{1}{c}{53.6} & \multicolumn{1}{c}{42.2} & \multicolumn{1}{c}{46.8} & \multicolumn{1}{c}{45.2} & \multicolumn{1}{c}{47.0} & \multicolumn{1}{c}{42.8} & \multicolumn{1}{c}{45.0} & \multicolumn{1}{c}{11.6} & \multicolumn{1}{c}{17.5} & \multicolumn{1}{c}{9.2}  & \multicolumn{1}{c}{12.8} & \multicolumn{1}{c}{42.0} & \multicolumn{1}{c}{15.3} & \multicolumn{1}{c}{23.4} & \multicolumn{1}{c}{73.8}
\\
\midrule
\rowcolor[rgb]{ .851,  .851,  .851}\multicolumn{17}{c}{\textbf{\textit{Baselines with GPT-4o-mini}}} \\
\multicolumn{1}{l}{FLARE~\citep{DBLP:conf/emnlp/JiangXGSLDYCN23}} & \multicolumn{1}{c}{45.8} & \multicolumn{1}{c}{52.9} & \multicolumn{1}{c}{39.2} & \multicolumn{1}{c}{46.0} & \multicolumn{1}{c}{54.8} & \multicolumn{1}{c}{53.6} & \multicolumn{1}{c}{42.4} & \multicolumn{1}{c}{50.3} & \multicolumn{1}{c}{18.6} & \multicolumn{1}{c}{24.9} & \multicolumn{1}{c}{15.6} & \multicolumn{1}{c}{19.7} & \multicolumn{1}{c}{36.8} & \multicolumn{1}{c}{9.9} & \multicolumn{1}{c}{23.4} & \multicolumn{1}{c}{70.0}
\\
\multicolumn{1}{l}{Self-RAG~\citep{DBLP:conf/iclr/AsaiWWSH24}} & \multicolumn{1}{c}{43.8} & \multicolumn{1}{c}{53.0} & \multicolumn{1}{c}{41.8} & \multicolumn{1}{c}{46.2} & \multicolumn{1}{c}{35.8} & \multicolumn{1}{c}{40.4} & \multicolumn{1}{c}{33.6} & \multicolumn{1}{c}{36.6} & \multicolumn{1}{c}{11.6} & \multicolumn{1}{c}{19.7} & \multicolumn{1}{c}{10.2} & \multicolumn{1}{c}{13.8} & \multicolumn{1}{c}{42.6} & \multicolumn{1}{c}{16.7} & \multicolumn{1}{c}{24.4} & \multicolumn{1}{c}{68.4}
\\
\multicolumn{1}{l}{CoN~\citep{DBLP:journals/corr/abs-2311-09210}} & \multicolumn{1}{c}{50.2} & \multicolumn{1}{c}{56.8} & \multicolumn{1}{c}{42.6} & \multicolumn{1}{c}{49.9} & \multicolumn{1}{c}{53.8} & \multicolumn{1}{c}{53.0} & \multicolumn{1}{c}{42.8} & \multicolumn{1}{c}{49.9} & \multicolumn{1}{c}{18.6} & \multicolumn{1}{c}{26.1} & \multicolumn{1}{c}{14.4} & \multicolumn{1}{c}{19.7} & \multicolumn{1}{c}{32.8} & \multicolumn{1}{c}{6.9} & \multicolumn{1}{c}{19.8} & \multicolumn{1}{c}{75.2}
\\
\multicolumn{1}{l}{RAT~\citep{DBLP:journals/corr/abs-2403-05313}} & \multicolumn{1}{c}{52.0} & \multicolumn{1}{c}{58.3} & \multicolumn{1}{c}{43.6} & \multicolumn{1}{c}{51.3} & \multicolumn{1}{c}{50.8} & \multicolumn{1}{c}{60.0} & \multicolumn{1}{c}{40.0} & \multicolumn{1}{c}{50.3} & \multicolumn{1}{c}{25.2} & \multicolumn{1}{c}{33.5} & \multicolumn{1}{c}{21.0} & \multicolumn{1}{c}{26.6} & \multicolumn{1}{c}{35.7} & \multicolumn{1}{c}{11.4} & \multicolumn{1}{c}{23.6} & \multicolumn{1}{c}{60.2}
\\
\multicolumn{1}{l}{ReAct~\citep{DBLP:conf/iclr/YaoZYDSN023}} & \multicolumn{1}{c}{56.0} & \multicolumn{1}{c}{56.8} & \multicolumn{1}{c}{40.4} & \multicolumn{1}{c}{51.1} & \multicolumn{1}{c}{63.6} & \multicolumn{1}{c}{52.6} & \multicolumn{1}{c}{35.6} & \multicolumn{1}{c}{50.6} & \multicolumn{1}{c}{27.0} & \multicolumn{1}{c}{29.3} & \multicolumn{1}{c}{16.6} & \multicolumn{1}{c}{24.3} & \multicolumn{1}{c}{39.4} & \multicolumn{1}{c}{15.1} & \multicolumn{1}{c}{27.3} & \multicolumn{1}{c}{72.0} \\
\midrule
\rowcolor[rgb]{0.85, 0.985, 0.985} \multicolumn{17}{c}{\textbf{\textit{ DeepNote (Ours)}}} \\
\multicolumn{1}{l}{DeepNote $_{\text{Qwen2.5-7B-Instruct}}$} & \multicolumn{1}{c}{50.6} & \multicolumn{1}{c}{59.2} & \multicolumn{1}{c}{48.0} & \multicolumn{1}{c}{52.6} & \multicolumn{1}{c}{50.0} & \multicolumn{1}{c}{51.4} & \multicolumn{1}{c}{41.8} & \multicolumn{1}{c}{47.7} & \multicolumn{1}{c}{14.6} & \multicolumn{1}{c}{19.8} & \multicolumn{1}{c}{11.6} & \multicolumn{1}{c}{15.3} & \multicolumn{1}{c}{44.4} & \multicolumn{1}{c}{19.4} & \multicolumn{1}{c}{26.4} & \multicolumn{1}{c}{71.6} \\
\multicolumn{1}{l}{\,\,\,\,\,\,\,\,+DPO $_{\text{Qwen2.5-7B-Instruct}}$} & \multicolumn{1}{c}{49.0} & \multicolumn{1}{c}{58.1} & \multicolumn{1}{c}{46.6} & \multicolumn{1}{c}{51.2} & \multicolumn{1}{c}{55.4} & \multicolumn{1}{c}{55.7} & \multicolumn{1}{c}{44.6} & \multicolumn{1}{c}{51.9} & \multicolumn{1}{c}{15.4} & \multicolumn{1}{c}{21.9} & \multicolumn{1}{c}{11.4} & \multicolumn{1}{c}{16.2} & \multicolumn{1}{c}{47.2} & \multicolumn{1}{c}{21.7} & \multicolumn{1}{c}{28.4} & \multicolumn{1}{c}{72.8} \\
\multicolumn{1}{l}{DeepNote $_{\text{Llama3.1-8B-Instruct}}$} & \multicolumn{1}{c}{48.0} & \multicolumn{1}{c}{54.3} & \multicolumn{1}{c}{41.2} & \multicolumn{1}{c}{47.8} & \multicolumn{1}{c}{58.0} & \multicolumn{1}{c}{58.2} & \multicolumn{1}{c}{48.2} & \multicolumn{1}{c}{54.8} & \multicolumn{1}{c}{17.0} & \multicolumn{1}{c}{21.3} & \multicolumn{1}{c}{13.2} & \multicolumn{1}{c}{17.2} & \multicolumn{1}{c}{43.4} & \multicolumn{1}{c}{17.9} & \multicolumn{1}{c}{26.2} & \multicolumn{1}{c}{70.8} \\
\multicolumn{1}{l}{\,\,\,\,\,\,\,\,+DPO $_{\text{Llama3.1-8B-Instruct}}$} & \multicolumn{1}{c}{54.6} & \multicolumn{1}{c}{58.9} & \multicolumn{1}{c}{44.0} & \multicolumn{1}{c}{52.5} & \multicolumn{1}{c}{63.8} & \multicolumn{1}{c}{60.5} & \multicolumn{1}{c}{47.4} & \multicolumn{1}{c}{57.2} & \multicolumn{1}{c}{24.4} & \multicolumn{1}{c}{27.3} & \multicolumn{1}{c}{14.4} & \multicolumn{1}{c}{22.0} & \multicolumn{1}{c}{46.4} & \multicolumn{1}{c}{19.8} & \multicolumn{1}{c}{29.4} & \multicolumn{1}{c}{74.2} \\
\multicolumn{1}{l}{DeepNote$_{\text{GPT-4o-mini}}$} & \multicolumn{1}{c}{56.8} & \multicolumn{1}{c}{64.3} & \multicolumn{1}{c}{50.2} & \multicolumn{1}{c}{57.1} & \multicolumn{1}{c}{66.2} & \multicolumn{1}{c}{63.7} & \multicolumn{1}{c}{52.6} & \multicolumn{1}{c}{60.8} & \multicolumn{1}{c}{24.8} & \multicolumn{1}{c}{31.3} & \multicolumn{1}{c}{18.4} & \multicolumn{1}{c}{24.8} & \multicolumn{1}{c}{\textbf{48.6}} & \multicolumn{1}{c}{\textbf{23.1}} & \multicolumn{1}{c}{\textbf{32.2}} & \multicolumn{1}{c}{\textbf{76.4}} \\
\multicolumn{1}{l}{DeepNote$_{\text{Llama3.1-70B-Instruct}}$} & \multicolumn{1}{c}{\textbf{59.2}} & \multicolumn{1}{c}{\textbf{67.2}} & \multicolumn{1}{c}{\textbf{54.2}} & \multicolumn{1}{c}{\textbf{60.2}} & \multicolumn{1}{c}{\textbf{72.4}} & \multicolumn{1}{c}{\textbf{67.1}} & \multicolumn{1}{c}{\textbf{55.8}} & \multicolumn{1}{c}{\textbf{65.1}} & \multicolumn{1}{c}{\textbf{32.6}} & \multicolumn{1}{c}{\textbf{35.0}} & \multicolumn{1}{c}{\textbf{23.0}} & \multicolumn{1}{c}{\textbf{30.2}} & \multicolumn{1}{c}{44.2} & \multicolumn{1}{c}{16.6} & \multicolumn{1}{c}{30.3} & \multicolumn{1}{c}{75.4} \\   \multicolumn{1}{l}{$\Delta_{\text{DeepNote}\rightarrow \text{Vanilla}}$} & 14.6$\uparrow$ & 13.6$\uparrow$ & 12.0$\uparrow$ & 13.4$\uparrow$ & 27.2$\uparrow$ & 20.1$\uparrow$ & 13.0$\uparrow$ & 20.1$\uparrow$ & 21.0$\uparrow$ & 17.5$\uparrow$ & 13.8$\uparrow$ & 17.4$\uparrow$  & 4.3$\uparrow$ & 5.6$\uparrow$ & 7.6$\uparrow$ & 5.2$\uparrow$ \\
    \bottomrule
    \end{tabular}%
    \caption{\textbf{Results (\%) of overall performance.}
    "\textbf{Bold}" denotes the highest value. Meanwhile, the symbol "$\uparrow$" indicates the increase in our highest value compared to the Vanilla baseline under the same backbone model setting.} 
  \label{tab:Overall Performance}%
\end{table*}%

\section{Experimental Setup}
In this section, we detail the experimental settings and summarize them in Appendix~\ref{Appendix C: Experimental Setup Details}.
\subsection{Datasets \& Metrics \& Corpora}
\textbf{Multi-hop QA task}\quad 
includes three challenging datasets: HotpotQA~\citep{DBLP:conf/emnlp/Yang0ZBCSM18}, 2WikiMultiHopQA (2WikiMQA)~\citep{DBLP:conf/coling/HoNSA20}, and MusiQue~\citep{DBLP:journals/tacl/TrivediBKS22}. They require the RAG system to retrieve multi-hop knowledge and provide accurate answers through multi-hop reasoning. For the evaluation data and retrieval corpus, we use the versions released by~\citet{DBLP:conf/acl/TrivediBKS23}.
For evaluation metrics, we follow ~\citeauthor{DBLP:conf/emnlp/JiangXGSLDYCN23} in using F1-Score (f1) and Exact Match (em).
Moreover, we also add Accuracy (acc.), a common metric for QA systems evaluation~\citep{vu2020ava}.


\noindent\textbf{Long-form QA task}\quad requires the system to gather diverse information and generate comprehensive answers.
We select the ASQA~\citep{DBLP:conf/emnlp/StelmakhLDC22} dataset to evaluate the system's ability to explore a wide range of relevant knowledge in response to the vague original question.
Specifically, we use the ASQA dataset with 948 queries recompiled by ALCE~\citep{DBLP:conf/emnlp/GaoYYC23} for evaluation and apply ALCE's official evaluation metrics, involving String Exact Match (str-em) and String Hit Rate (str-hit).

\noindent\textbf{Short-form QA task}\quad
aims to gather factual and commonsense information to produce brief responses, with relatively simple retrieval and reasoning requirements. We select StrategyQA~\citep{DBLP:journals/tacl/GevaKSKRB21} to evaluate the system's performance and robustness on simpler tasks. It requires the system to retrieve commonsense details and output a \texttt{Yes/No} answer.
We follow the test set from previous work~\citep{DBLP:journals/tmlr/SrivastavaRRSAF23}, randomly sampling 500 samples for evaluation, with accuracy (acc.) as the evaluation metric.

\subsection{Baselines \& LLMs}
We extensively compare five types of baselines: 1) LLMs without Retrieval, which directly feeds queries into LLMs to output answers;
2) Vanilla RAG (Vanilla), which employs one-time retrieval and directly inputs the retrieved passages along with the query to generate an answer;
3) Single-Step RAG (SSRAG), which involves additional processing of the retrieved knowledge, such as summarization, based on Vanilla RAG;
4) Multi-Step RAG (MSRAG), which employs multiple retrievals;
5) Adaptive RAG (ARAG), which leverages an adaptive forward exploration strategy to retrieve knowledge to enhance answer quality.
For SSRAG, we use Vanilla RAG, Chain-of-note (CoN) as counterparties.
For MSRAG, we select RAT for comparison. 
For ARAG, we select three recent mainstream methods for comparison, including FLARE, Self-RAG, and ReAct.
Additionally, we conduct experiments on a series of LLMs, including GPT-4o~\citep{DBLP:journals/corr/abs-2410-21276}
(OpenAI \texttt{gpt-4o-mini-0718}), Qwen2.5-7b~\citep{DBLP:journals/corr/abs-2412-15115}, Llama3.1-70B-Instruct and Llama3.1-8B~\citep{DBLP:journals/corr/abs-2407-21783}.

\subsection{Retrievers}
We conduct experiments on all multi-hop datasets using two types of retrievers: BM25, implemented in Elasticsearch as the sparse retriever, and bge-base-en-v1.5 as the dense retriever.
For ASQA and StrategyQA, we employ the dense retriever GTR-XXL~\citep{DBLP:conf/emnlp/Ni0LDAMZLHCY22} following~\citeauthor{DBLP:conf/emnlp/GaoYYC23}, and we use the corpus provided by ALCE.
In addition, we evaluate the performance of our framework under various top-$k$ settings, top-$k\in\left\{ 3,5,7 \right\}$, with a default of 5 (more results in Appendix~\ref{appsubsec:Impact of Different Top-k}).

\subsection{Implementation Details}
Our method conducts all inference and data construction under a zero-shot setting,
and we align the prompts for generation within the same dataset (cf. Appendix~\ref{Appendix B: Prompt Details}).
In practice, we utilize the vLLM~\citep{DBLP:conf/sosp/KwonLZ0ZY0ZS23} inference acceleration tool to speed up the inference of local open-source models.
Since our approach involves an adaptive iterative process, we also employ various iteration halt condition recipes to conduct a thorough analysis of our framework's performance and robustness (cf. Appendix~\ref{appsubsec:Adaptive Parameter Analysis}).
During DPO training, we perform full parameter fine-tuning on 8×A100 GPUs, using a batch size of 8, a learning rate of 5e-7, and $\beta$ set to 0.1, training the model for one epoch.


\begin{table*}[ht!]
  \centering
    \setlength{\tabcolsep}{2.2pt}
    \fontsize{8}{8}\selectfont
    \begin{tabular}{l|cccc|cccc|cccc|ccc|c}
    \toprule
    \multicolumn{1}{c}{\multirow{2}[4]{*}{\textbf{Methods}}} & \multicolumn{4}{|c}{\textbf{HotpotQA}} & \multicolumn{4}{c}{\textbf{2WikiMQA}} & \multicolumn{4}{c}{\textbf{MusiQue}} & \multicolumn{3}{c}{\textbf{ASQA}} & \textbf{StrategyQA } \\
\cmidrule(lr){2-5}\cmidrule(lr){6-9} \cmidrule(lr){10-13}\cmidrule(lr){14-16}  \cmidrule(lr){17-17}          & acc.  & f1    & em    & avg.  & acc.  & f1    & em    & avg.  & acc.  & f1    & em    & avg.  & str-em & str-hit & avg.  & acc. \\
    \midrule
    \multicolumn{17}{c}{\textit{GPT-4o-mini}} \\
    DeepNote & \cellcolor[rgb]{ .882,  .984,  .992}56.8  & \cellcolor[rgb]{ .882,  .984,  .992}64.3  & \cellcolor[rgb]{ .882,  .984,  .992}50.2  & \cellcolor[rgb]{ .882,  .984,  .992}57.1  & \cellcolor[rgb]{ .882,  .984,  .992}66.2  & \cellcolor[rgb]{ .882,  .984,  .992}63.7  & \cellcolor[rgb]{ .882,  .984,  .992}52.6  & \cellcolor[rgb]{ .882,  .984,  .992}60.8  & \cellcolor[rgb]{ .882,  .984,  .992}24.8  & \cellcolor[rgb]{ .882,  .984,  .992}31.3  & \cellcolor[rgb]{ .882,  .984,  .992}18.4  & \cellcolor[rgb]{ .882,  .984,  .992}24.8  & \cellcolor[rgb]{ .882,  .984,  .992}48.6  & \cellcolor[rgb]{ .882,  .984,  .992}23.1  & \cellcolor[rgb]{ .882,  .984,  .992}32.2  & \cellcolor[rgb]{ .882,  .984,  .992}76.4  \\
    w/o Adap. Retrieval & \cellcolor[rgb]{ .973,  .953,  .984}47.0  & \cellcolor[rgb]{ .973,  .953,  .984}54.6  & \cellcolor[rgb]{ .973,  .953,  .984}41.4  & \cellcolor[rgb]{ .973,  .953,  .984}47.7  & \cellcolor[rgb]{ .973,  .953,  .984}46.2  & \cellcolor[rgb]{ .973,  .953,  .984}48.8  & \cellcolor[rgb]{ .973,  .953,  .984}43.4  & \cellcolor[rgb]{ .973,  .953,  .984}46.1  & \cellcolor[rgb]{ .973,  .953,  .984}14.2  & \cellcolor[rgb]{ .973,  .953,  .984}20.8  & \cellcolor[rgb]{ .973,  .953,  .984}10.8  & \cellcolor[rgb]{ .973,  .953,  .984}15.3  & \cellcolor[rgb]{ .973,  .953,  .984}47.1  & \cellcolor[rgb]{ .973,  .953,  .984}21.0  & \cellcolor[rgb]{ .973,  .953,  .984}27.8  & \cellcolor[rgb]{ .973,  .953,  .984}74.8  \\
    w/o Adap. Retrieval \& Init. Note & \cellcolor[rgb]{ .953,  .886,  1}44.0  & \cellcolor[rgb]{ .953,  .886,  1}52.2  & \cellcolor[rgb]{ .953,  .886,  1}40.0  & \cellcolor[rgb]{ .953,  .886,  1}45.4  & \cellcolor[rgb]{ .953,  .886,  1}40.4  & \cellcolor[rgb]{ .953,  .886,  1}44.4  & \cellcolor[rgb]{ .953,  .886,  1}39.2  & \cellcolor[rgb]{ .953,  .886,  1}41.3  & \cellcolor[rgb]{ .953,  .886,  1}10.6  & \cellcolor[rgb]{ .953,  .886,  1}17.3  & \cellcolor[rgb]{ .953,  .886,  1}7.6  & \cellcolor[rgb]{ .953,  .886,  1}11.8  & \cellcolor[rgb]{ .953,  .886,  1}44.3  & \cellcolor[rgb]{ .953,  .886,  1}17.5  & \cellcolor[rgb]{ .953,  .886,  1}24.5  & \cellcolor[rgb]{ .953,  .886,  1}71.2  \\
    \midrule
    \multicolumn{17}{c}{\textit{Llama3.1-70B-Instruct}} \\
    DeepNote & \cellcolor[rgb]{ .882,  .984,  .992}59.2  & \cellcolor[rgb]{ .882,  .984,  .992}67.2  & \cellcolor[rgb]{ .882,  .984,  .992}54.2  & \cellcolor[rgb]{ .882,  .984,  .992}60.2  & \cellcolor[rgb]{ .882,  .984,  .992}72.4  & \cellcolor[rgb]{ .882,  .984,  .992}67.1  & \cellcolor[rgb]{ .882,  .984,  .992}55.8  & \cellcolor[rgb]{ .882,  .984,  .992}65.1  & \cellcolor[rgb]{ .882,  .984,  .992}32.6  & \cellcolor[rgb]{ .882,  .984,  .992}35.0  & \cellcolor[rgb]{ .882,  .984,  .992}23.0  & \cellcolor[rgb]{ .882,  .984,  .992}30.2  & \cellcolor[rgb]{ .882,  .984,  .992}44.2  & \cellcolor[rgb]{ .882,  .984,  .992}16.6  & \cellcolor[rgb]{ .882,  .984,  .992}30.3  & \cellcolor[rgb]{ .882,  .984,  .992}75.4  \\
    w/o Adap. Retrieval & \cellcolor[rgb]{ .953,  .886,  1}42.6  & \cellcolor[rgb]{ .953,  .886,  1}51.0  & \cellcolor[rgb]{ .953,  .886,  1}39.8  & \cellcolor[rgb]{ .953,  .886,  1}44.5  & \cellcolor[rgb]{ .953,  .886,  1}38.8  & \cellcolor[rgb]{ .953,  .886,  1}40.0  & \cellcolor[rgb]{ .953,  .886,  1}36.6  & \cellcolor[rgb]{ .953,  .886,  1}38.5  & \cellcolor[rgb]{ .973,  .953,  .984}12.6  & \cellcolor[rgb]{ .953,  .886,  1}16.3  & \cellcolor[rgb]{ .973,  .953,  .984}10.4  & \cellcolor[rgb]{ .973,  .953,  .984}13.1  & \cellcolor[rgb]{ .973,  .953,  .984}42.4  & \cellcolor[rgb]{ .973,  .953,  .984}15.5  & \cellcolor[rgb]{ .973,  .953,  .984}23.7  & \cellcolor[rgb]{ .973,  .953,  .984}73.8  \\
    w/o Adap. Retrieval \& Init. Note & \cellcolor[rgb]{ .973,  .953,  .984}44.6  & \cellcolor[rgb]{ .973,  .953,  .984}53.6  & \cellcolor[rgb]{ .973,  .953,  .984}42.2  & \cellcolor[rgb]{ .973,  .953,  .984}46.8  & \cellcolor[rgb]{ .973,  .953,  .984}45.2  & \cellcolor[rgb]{ .973,  .953,  .984}47.0  & \cellcolor[rgb]{ .973,  .953,  .984}42.8  & \cellcolor[rgb]{ .973,  .953,  .984}45.0  & \cellcolor[rgb]{ .953,  .886,  1}11.6  & \cellcolor[rgb]{ .973,  .953,  .984}17.5  & \cellcolor[rgb]{ .953,  .886,  1}9.2  & \cellcolor[rgb]{ .953,  .886,  1}12.8  & \cellcolor[rgb]{ .953,  .886,  1}42.0  & \cellcolor[rgb]{ .953,  .886,  1}15.3  & \cellcolor[rgb]{ .953,  .886,  1}23.4  & \cellcolor[rgb]{ .973,  .953,  .984}73.8  \\
    \bottomrule
    \end{tabular}%
    \caption{{\bf Results (\%) of the ablation study.}
    The "w/o Adap. Retrieval" denotes that DeepNote employs only the initial note without adaptive retrieval; the "w/o Adap. Retrieval \& Init. Note" means DeepNote employs neither adaptive retrieval nor initial note, which degenerates into Vanilla RAG. The "avg." denotes the arithmetic mean.
    \colorbox[rgb]{ .882,  .984,  .992}{"Blue"}, \colorbox[rgb]{ .973,  .953,  .984}{"light purple"} and \colorbox[rgb]{ .953,  .89,  1}{"dark purple"} represent the highest, second highest, and lowest values.}
  \label{tab:Results of the ablation study}%
\end{table*}%
\section{Results and Analysis}
\subsection{Overall Performance}
The overall performance of DeepNote in three types of QA tasks is shown in Table~\ref{tab:Overall Performance}.

\noindent\textbf{Vanilla RAG struggles to meet complex retrieval demands, while DeepNote shows significant improvement in complex QA tasks.}
As shown in Table~\ref{tab:Overall Performance}, we observe that Vanilla RAG performs well on relatively simple short-generation tasks but shows poor performance on complex multi-hop QA, highlighting that simple one-time retrieval fails to meet the demands of complex retrieval and reasoning. 
In contrast, DeepNote demonstrates significant performance improvements over Vanilla RAG on all datasets, regardless of whether using industry-leading closed-source models or small-size parameter open-source models. Our framework achieves a notable improvement by up to 20.1\%, which confirms the effectiveness and importance of the deep exploration of our framework.

\noindent\textbf{Even with information refinement, the single-step RAG remains limited by the knowledge boundary due to the one-time retrieval.}
DeepNote significantly outperforms the SSRAG method, CoN, on all complex QA tasks, while also showing performance advantages on simple short-form QA tasks. This trend indicates that although CoN summarizes retrieved documents to reduce noise, it still has a knowledge boundary.
Furthermore, we find that the performance of CoN decreases significantly on long-form tasks compared to other tasks. This suggests that note-centric adaptive exploration fosters more effective and stable knowledge growth than CoN while avoiding knowledge loss.

\noindent\textbf{DeepNote enables more effective and robust knowledge exploration and accumulation.}
Compared to the MSRAG and ARAG, DeepNote shows great performance advantages across all QA tasks, demonstrating its superiority and generalization.
We provide an in-depth analysis of the reasons behind this advantage. First, multi-step RAG (i.e. RAT) often introduces noise due to indiscriminate retrieval~\citep{DBLP:conf/iclr/AsaiWWSH24}. On the other hand, ARAG relies on limited retrieval data or previously generated segments to determine the next retrieval strategies.
The difference is that we use a note-centric approach to continuously accumulate knowledge from the perspective of information growth while avoiding noise during the adaptive iteration process.
The best note is used to make the next retrieval decision. This enables the system to ensure knowledge growth during exploration and make more effective and robust retrieval decisions based on the best knowledge.

\noindent\textbf{DPO effectively improves the model's ability to follow instructions in multi-stage tasks, leading to further performance gains of our framework.}
We find that DPO significantly improves the overall performance of DeepNote in most cases. Specifically, DPO improves the in-domain performance of our DeepNote by up to 4.2\%. This improvement also generalized to more challenging out-of-domain multi-hop QA data (i.e., MusiQue) and other types of out-of-domain tasks (i.e., long-form and short-form QA tasks), with an improvement of up to 4.8\%.
Importantly, we achieve broad performance improvements by training on data from a single dataset, 2WikiMQA. These results validate the effectiveness and generalization of our automated data construction pipeline, DNAlign training data, and multi-task training strategy.

\subsection{Ablation Study}
In the ablation study, we validate the effectiveness of the note-centric adaptive retrieval process and note initialization. Table~\ref{tab:Results of the ablation study} presents the main results of our ablation experiments, with additional results provided in Appendix~\ref{appsubsec:Ablation Study}.

We find that DeepNote significantly outperforms "w/o Adap. Retrieval", particularly on multi-hop datasets where the performance gap is more pronounced. These results validate the effectiveness of our note-centric adaptive retrieval process, which enables stable knowledge accumulation.
Notably, since the adaptive process is intrinsically built on notes, the initialization note and adaptive retrieval are interdependent.
Therefore, we further compare DeepNote with "w/o Adap. Retrieval \& Init. Note", which reveals that the initial note generally achieves superior performance over Vanilla RAG in most cases, though occasional performance degradation occurs. 
This suggests that the initial note is effective, but its performance can be unstable due to the inherent one-time summarization and refinement of information.

\begin{figure*}[ht!]
\includegraphics[width=\textwidth]{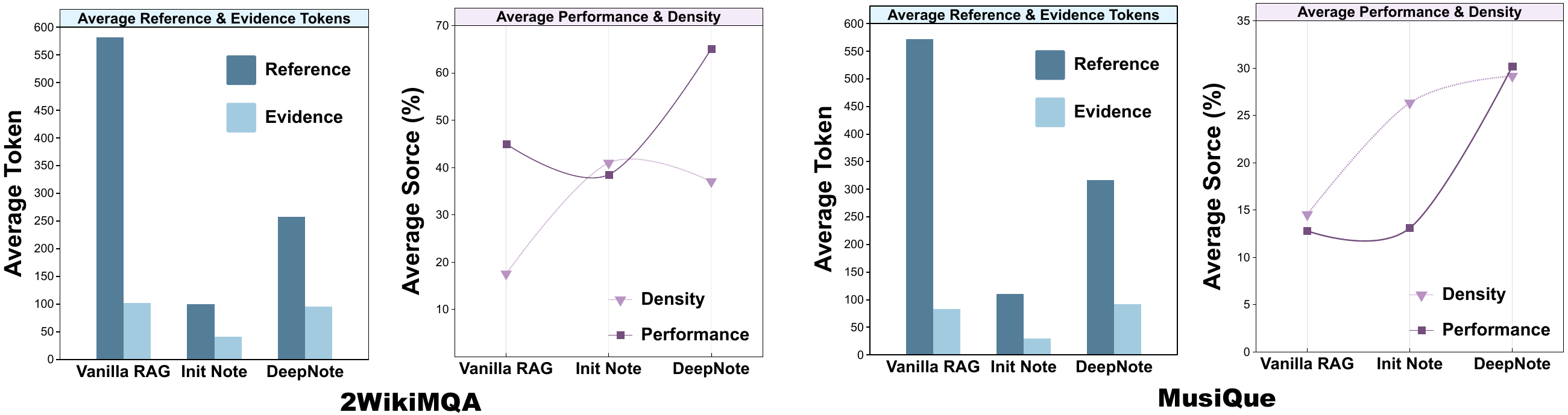}
\caption{\textbf{Knowledge Density Comparision on Llama3.1-70B-Instruct.} The "Init Note" means that the initial note. We calculated the arithmetic mean of token length, density, and performance.}
\label{fig:row-Knowledge Density Comparision on Llama3.1-70B-Instruct}
\end{figure*}

\begin{figure}[ht!]
\includegraphics[width=\columnwidth]{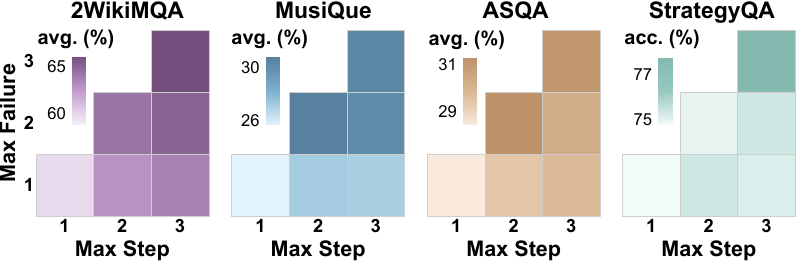}
\caption{{\bf Performance on different adaptive hyper-parameters} with Llama3.1-70B-Instruct.}
\label{fig:Adaptive Parameter Analysis}
\end{figure}

\begin{figure}[ht!]
\includegraphics[width=0.9\columnwidth]{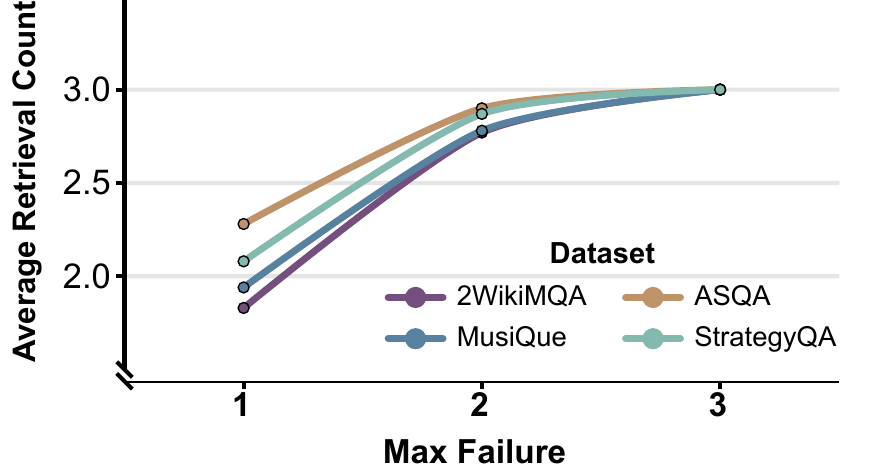}
\caption{{\bf Retrieval efficiency on different adaptive hyper-parameters} with Llama3.1-70B-Instruct.}
\label{fig:Retrieval Counts}
\end{figure}

\subsection{Analysis}
\noindent{\bf Knowledge Density and Performance Analysis}\quad
We conduct an in-depth analysis of how different processes in our framework affect the density of collected knowledge.
In Figure~\ref{fig:row-Knowledge Density Comparision on Llama3.1-70B-Instruct}, we refer to the retrieved documents or notes used in the final answers by Vanilla, initial note alone, and DeepNote as "References".
The portions of the "Reference" relevant to answering the original query are termed "Evidence". Specifically, we employ the model used in the answer generation stage to identify the "Evidence". Based on this, we also calculate the proportion of "Evidence" token length within the "Reference", referred to as knowledge density.
We find that the references in Vanilla are very lengthy but have low knowledge density, indicating significant noise in these references. The initial note improves knowledge density by summarizing and refining the information retrieved in a single pass.
However, this increase in density is mainly due to the sharp reduction in the total token length of the references.
In Figure~\ref{fig:row-Knowledge Density Comparision on Llama3.1-70B-Instruct} and Table~\ref{tab:Results of the ablation study}, we find that the initial note refines knowledge and reduces noise, thereby enhancing performance in most cases, although instability may arise due to the reduced total knowledge volume.
In contrast, our framework achieves a knowledge density comparable to the initial note and significantly higher than Vanilla, while showing substantial performance improvement. This suggests that note-centric adaptive retrieval can gather more comprehensive, refined, and accurate knowledge while minimizing noise.

\noindent{\bf Efficiency and Performance Trade-off}\quad
Using DeepNote, researchers can adjust the failure update threshold and total iteration threshold to control exploration depth.
In Figures~\ref{fig:Adaptive Parameter Analysis} and~\ref{fig:Retrieval Counts}, we investigate the impact of the adaptive stop threshold on both performance and retrieval counts. Figure~\ref{fig:Adaptive Parameter Analysis} suggests that performance improves as the total iteration threshold increases, while the maximum update failure threshold remains constant. This improvement arises from relaxing the total iteration constraint, which facilitates deeper exploration through additional retrieval attempts. Conversely, when the total iteration threshold is fixed, increasing the update failure threshold also enhances performance by allowing greater tolerance for errors during exploration. Notably, competitive performance is achieved when the two thresholds are set to similar values.
In Figure~\ref{fig:Retrieval Counts}, we further show the total number of retrievals used during the adaptive retrieval process (excluding the retrievals in the note initialization). We find that
increasing the threshold requires more retrieval counts, accompanied by diminishing marginal returns.
Therefore, when balancing retrieval efficiency and performance, it is advisable to choose a moderate or lower failure threshold and set the total iteration threshold slightly higher than it.

\section{Conclusion}
In this work, we identify two limitations in the existing studies and develop a novel ARAG framework--\textbf{DeepNote}. DeepNote uses notes as knowledge carriers for stable knowledge growth and devises optimal retrieval strategies based on the best available knowledge. Extensive empirical experiments, ablation studies, and multi-dimensional analyses confirm the superiority of DeepNote across various question-answering tasks and its flexibility in balancing retrieval efficiency and performance.

\section{Limitations}
Experiments demonstrate that DeepNote significantly advances RAG systems in tackling complex problems through robust and superior deep knowledge exploration and continuous information accumulation.
However, certain limitations still warrant attention. First, this work focuses on single-source retrieval; future efforts should explore dynamic knowledge integration in multi-source settings. Second, existing datasets prioritize early-stage exploration gains, leaving the performance of DeepNote in long-chain tasks unexplored. Building long-chain datasets could better align models with high-quality responses in later iterations.

\bibliography{acl}

\begin{thebibliography}{50}
\providecommand{\natexlab}[1]{#1}

\bibitem[{Asai et~al.(2024)Asai, Wu, Wang, Sil, and Hajishirzi}]{DBLP:conf/iclr/AsaiWWSH24}
Akari Asai, Zeqiu Wu, Yizhong Wang, Avirup Sil, and Hannaneh Hajishirzi. 2024.
\newblock \href {https://openreview.net/forum?id=hSyW5go0v8} {Self-rag: Learning to retrieve, generate, and critique through self-reflection}.
\newblock In \emph{The Twelfth International Conference on Learning Representations, {ICLR} 2024, Vienna, Austria, May 7-11, 2024}. OpenReview.net.

\bibitem[{Baek et~al.(2023)Baek, Aji, and Saffari}]{DBLP:journals/corr/abs-2306-04136}
Jinheon Baek, Alham~Fikri Aji, and Amir Saffari. 2023.
\newblock \href {https://doi.org/10.48550/ARXIV.2306.04136} {Knowledge-augmented language model prompting for zero-shot knowledge graph question answering}.
\newblock \emph{CoRR}, abs/2306.04136.

\bibitem[{Borgeaud et~al.(2022)Borgeaud, Mensch, Hoffmann, Cai, Rutherford, Millican, van~den Driessche, Lespiau, Damoc, Clark, de~Las~Casas, Guy, Menick, Ring, Hennigan, Huang, Maggiore, Jones, Cassirer, Brock, Paganini, Irving, Vinyals, Osindero, Simonyan, Rae, Elsen, and Sifre}]{DBLP:conf/icml/BorgeaudMHCRM0L22}
Sebastian Borgeaud, Arthur Mensch, Jordan Hoffmann, Trevor Cai, Eliza Rutherford, Katie Millican, George van~den Driessche, Jean{-}Baptiste Lespiau, Bogdan Damoc, Aidan Clark, Diego de~Las~Casas, Aurelia Guy, Jacob Menick, Roman Ring, Tom Hennigan, Saffron Huang, Loren Maggiore, Chris Jones, Albin Cassirer, Andy Brock, Michela Paganini, Geoffrey Irving, Oriol Vinyals, Simon Osindero, Karen Simonyan, Jack~W. Rae, Erich Elsen, and Laurent Sifre. 2022.
\newblock \href {https://proceedings.mlr.press/v162/borgeaud22a.html} {Improving language models by retrieving from trillions of tokens}.
\newblock In \emph{International Conference on Machine Learning, {ICML} 2022, 17-23 July 2022, Baltimore, Maryland, {USA}}, volume 162 of \emph{Proceedings of Machine Learning Research}, pages 2206--2240. {PMLR}.

\bibitem[{Chen et~al.(2023)Chen, Fu, Yuan, Wen, Fan, Liu, Zhang, Li, and Xiao}]{DBLP:conf/cikm/ChenFYWFL0LX23}
Yuyan Chen, Qiang Fu, Yichen Yuan, Zhihao Wen, Ge~Fan, Dayiheng Liu, Dongmei Zhang, Zhixu Li, and Yanghua Xiao. 2023.
\newblock \href {https://doi.org/10.1145/3583780.3614905} {Hallucination detection: Robustly discerning reliable answers in large language models}.
\newblock In \emph{Proceedings of the 32nd {ACM} International Conference on Information and Knowledge Management, {CIKM} 2023, Birmingham, United Kingdom, October 21-25, 2023}, pages 245--255. {ACM}.

\bibitem[{de~Luis~Balaguer et~al.(2024)de~Luis~Balaguer, Benara, de~Freitas~Cunha, de~M.~Estev{\~{a}}o~Filho, Hendry, Holstein, Marsman, Mecklenburg, Malvar, Nunes, Padilha, Sharp, Silva, Sharma, Aski, and Chandra}]{DBLP:journals/corr/abs-2401-08406}
Maria~Angels de~Luis~Balaguer, Vinamra Benara, Renato~Luiz de~Freitas~Cunha, Roberto de~M.~Estev{\~{a}}o~Filho, Todd Hendry, Daniel Holstein, Jennifer Marsman, Nick Mecklenburg, Sara Malvar, Leonardo~O. Nunes, Rafael Padilha, Morris Sharp, Bruno Silva, Swati Sharma, Vijay Aski, and Ranveer Chandra. 2024.
\newblock \href {https://doi.org/10.48550/ARXIV.2401.08406} {{RAG} vs fine-tuning: Pipelines, tradeoffs, and a case study on agriculture}.
\newblock \emph{CoRR}, abs/2401.08406.

\bibitem[{Dubey et~al.(2024)Dubey, Jauhri, Pandey, Kadian, Al{-}Dahle, Letman, Mathur, Schelten, Yang, Fan, Goyal, Hartshorn, Yang, Mitra, Sravankumar, Korenev, Hinsvark, Rao, Zhang, Rodriguez, Gregerson, Spataru, Rozi{\`{e}}re, Biron, Tang, Chern, Caucheteux, Nayak, Bi, Marra, McConnell, Keller, Touret, Wu, Wong, Ferrer, Nikolaidis, Allonsius, Song, Pintz, Livshits, Esiobu, Choudhary, Mahajan, Garcia{-}Olano, Perino, Hupkes, Lakomkin, AlBadawy, Lobanova, Dinan, Smith, Radenovic, Zhang, Synnaeve, Lee, Anderson, Nail, Mialon, Pang, Cucurell, Nguyen, Korevaar, Xu, Touvron, Zarov, Ibarra, Kloumann, Misra, Evtimov, Copet, Lee, Geffert, Vranes, Park, Mahadeokar, Shah, van~der Linde, Billock, Hong, Lee, Fu, Chi, Huang, Liu, Wang, Yu, Bitton, Spisak, Park, Rocca, Johnstun, Saxe, Jia, Alwala, Upasani, Plawiak, Li, Heafield, Stone, and et~al.}]{DBLP:journals/corr/abs-2407-21783}
Abhimanyu Dubey, Abhinav Jauhri, Abhinav Pandey, Abhishek Kadian, Ahmad Al{-}Dahle, Aiesha Letman, Akhil Mathur, Alan Schelten, Amy Yang, Angela Fan, Anirudh Goyal, Anthony Hartshorn, Aobo Yang, Archi Mitra, Archie Sravankumar, Artem Korenev, Arthur Hinsvark, Arun Rao, Aston Zhang, Aur{\'{e}}lien Rodriguez, Austen Gregerson, Ava Spataru, Baptiste Rozi{\`{e}}re, Bethany Biron, Binh Tang, Bobbie Chern, Charlotte Caucheteux, Chaya Nayak, Chloe Bi, Chris Marra, Chris McConnell, Christian Keller, Christophe Touret, Chunyang Wu, Corinne Wong, Cristian~Canton Ferrer, Cyrus Nikolaidis, Damien Allonsius, Daniel Song, Danielle Pintz, Danny Livshits, David Esiobu, Dhruv Choudhary, Dhruv Mahajan, Diego Garcia{-}Olano, Diego Perino, Dieuwke Hupkes, Egor Lakomkin, Ehab AlBadawy, Elina Lobanova, Emily Dinan, Eric~Michael Smith, Filip Radenovic, Frank Zhang, Gabriel Synnaeve, Gabrielle Lee, Georgia~Lewis Anderson, Graeme Nail, Gr{\'{e}}goire Mialon, Guan Pang, Guillem Cucurell, Hailey Nguyen, Hannah Korevaar, Hu~Xu, Hugo
  Touvron, Iliyan Zarov, Imanol~Arrieta Ibarra, Isabel~M. Kloumann, Ishan Misra, Ivan Evtimov, Jade Copet, Jaewon Lee, Jan Geffert, Jana Vranes, Jason Park, Jay Mahadeokar, Jeet Shah, Jelmer van~der Linde, Jennifer Billock, Jenny Hong, Jenya Lee, Jeremy Fu, Jianfeng Chi, Jianyu Huang, Jiawen Liu, Jie Wang, Jiecao Yu, Joanna Bitton, Joe Spisak, Jongsoo Park, Joseph Rocca, Joshua Johnstun, Joshua Saxe, Junteng Jia, Kalyan~Vasuden Alwala, Kartikeya Upasani, Kate Plawiak, Ke~Li, Kenneth Heafield, Kevin Stone, and et~al. 2024.
\newblock \href {https://doi.org/10.48550/ARXIV.2407.21783} {The llama 3 herd of models}.
\newblock \emph{CoRR}, abs/2407.21783.

\bibitem[{Fu et~al.(2024)Fu, Ng, Jiang, and Liu}]{DBLP:conf/naacl/FuNJ024}
Jinlan Fu, See{-}Kiong Ng, Zhengbao Jiang, and Pengfei Liu. 2024.
\newblock \href {https://doi.org/10.18653/V1/2024.NAACL-LONG.365} {Gptscore: Evaluate as you desire}.
\newblock In \emph{Proceedings of the 2024 Conference of the North American Chapter of the Association for Computational Linguistics: Human Language Technologies (Volume 1: Long Papers), {NAACL} 2024, Mexico City, Mexico, June 16-21, 2024}, pages 6556--6576. Association for Computational Linguistics.

\bibitem[{Gao et~al.(2023{\natexlab{a}})Gao, Yen, Yu, and Chen}]{DBLP:conf/emnlp/GaoYYC23}
Tianyu Gao, Howard Yen, Jiatong Yu, and Danqi Chen. 2023{\natexlab{a}}.
\newblock \href {https://doi.org/10.18653/V1/2023.EMNLP-MAIN.398} {Enabling large language models to generate text with citations}.
\newblock In \emph{Proceedings of the 2023 Conference on Empirical Methods in Natural Language Processing, {EMNLP} 2023, Singapore, December 6-10, 2023}, pages 6465--6488. Association for Computational Linguistics.

\bibitem[{Gao et~al.(2023{\natexlab{b}})Gao, Xiong, Gao, Jia, Pan, Bi, Dai, Sun, Guo, Wang, and Wang}]{DBLP:journals/corr/abs-2312-10997}
Yunfan Gao, Yun Xiong, Xinyu Gao, Kangxiang Jia, Jinliu Pan, Yuxi Bi, Yi~Dai, Jiawei Sun, Qianyu Guo, Meng Wang, and Haofen Wang. 2023{\natexlab{b}}.
\newblock \href {https://doi.org/10.48550/ARXIV.2312.10997} {Retrieval-augmented generation for large language models: {A} survey}.
\newblock \emph{CoRR}, abs/2312.10997.

\bibitem[{Geva et~al.(2021)Geva, Khashabi, Segal, Khot, Roth, and Berant}]{DBLP:journals/tacl/GevaKSKRB21}
Mor Geva, Daniel Khashabi, Elad Segal, Tushar Khot, Dan Roth, and Jonathan Berant. 2021.
\newblock \href {https://doi.org/10.1162/TACL\_A\_00370} {Did aristotle use a laptop? {A} question answering benchmark with implicit reasoning strategies}.
\newblock \emph{Trans. Assoc. Comput. Linguistics}, 9:346--361.

\bibitem[{Guu et~al.(2020)Guu, Lee, Tung, Pasupat, and Chang}]{DBLP:conf/icml/GuuLTPC20}
Kelvin Guu, Kenton Lee, Zora Tung, Panupong Pasupat, and Ming{-}Wei Chang. 2020.
\newblock \href {http://proceedings.mlr.press/v119/guu20a.html} {Retrieval augmented language model pre-training}.
\newblock In \emph{Proceedings of the 37th International Conference on Machine Learning, {ICML} 2020, 13-18 July 2020, Virtual Event}, volume 119 of \emph{Proceedings of Machine Learning Research}, pages 3929--3938. {PMLR}.

\bibitem[{He et~al.(2023)He, Zhang, and Roth}]{DBLP:journals/corr/abs-2301-00303}
Hangfeng He, Hongming Zhang, and Dan Roth. 2023.
\newblock \href {https://doi.org/10.48550/ARXIV.2301.00303} {Rethinking with retrieval: Faithful large language model inference}.
\newblock \emph{CoRR}, abs/2301.00303.

\bibitem[{Ho et~al.(2020)Ho, Nguyen, Sugawara, and Aizawa}]{DBLP:conf/coling/HoNSA20}
Xanh Ho, Anh{-}Khoa~Duong Nguyen, Saku Sugawara, and Akiko Aizawa. 2020.
\newblock \href {https://doi.org/10.18653/V1/2020.COLING-MAIN.580} {Constructing {A} multi-hop {QA} dataset for comprehensive evaluation of reasoning steps}.
\newblock In \emph{Proceedings of the 28th International Conference on Computational Linguistics, {COLING} 2020, Barcelona, Spain (Online), December 8-13, 2020}, pages 6609--6625. International Committee on Computational Linguistics.

\bibitem[{Hurst et~al.(2024)Hurst, Lerer, Goucher, Perelman, Ramesh, Clark, Ostrow, Welihinda, Hayes, Radford, Madry, Baker{-}Whitcomb, Beutel, Borzunov, Carney, Chow, Kirillov, Nichol, Paino, Renzin, Passos, Kirillov, Christakis, Conneau, Kamali, Jabri, Moyer, Tam, Crookes, Tootoonchian, Kumar, Vallone, Karpathy, Braunstein, Cann, Codispoti, Galu, Kondrich, Tulloch, Mishchenko, Baek, Jiang, Pelisse, Woodford, Gosalia, Dhar, Pantuliano, Nayak, Oliver, Zoph, Ghorbani, Leimberger, Rossen, Sokolowsky, Wang, Zweig, Hoover, Samic, McGrew, Spero, Giertler, Cheng, Lightcap, Walkin, Quinn, Guarraci, Hsu, Kellogg, Eastman, Lugaresi, Wainwright, Bassin, Hudson, Chu, Nelson, Li, Shern, Conger, Barette, Voss, Ding, Lu, Zhang, Beaumont, Hallacy, Koch, Gibson, Kim, Choi, McLeavey, Hesse, Fischer, Winter, Czarnecki, Jarvis, Wei, Koumouzelis, and Sherburn}]{DBLP:journals/corr/abs-2410-21276}
Aaron Hurst, Adam Lerer, Adam~P. Goucher, Adam Perelman, Aditya Ramesh, Aidan Clark, AJ~Ostrow, Akila Welihinda, Alan Hayes, Alec Radford, Aleksander Madry, Alex Baker{-}Whitcomb, Alex Beutel, Alex Borzunov, Alex Carney, Alex Chow, Alex Kirillov, Alex Nichol, Alex Paino, Alex Renzin, Alex~Tachard Passos, Alexander Kirillov, Alexi Christakis, Alexis Conneau, Ali Kamali, Allan Jabri, Allison Moyer, Allison Tam, Amadou Crookes, Amin Tootoonchian, Ananya Kumar, Andrea Vallone, Andrej Karpathy, Andrew Braunstein, Andrew Cann, Andrew Codispoti, Andrew Galu, Andrew Kondrich, Andrew Tulloch, Andrey Mishchenko, Angela Baek, Angela Jiang, Antoine Pelisse, Antonia Woodford, Anuj Gosalia, Arka Dhar, Ashley Pantuliano, Avi Nayak, Avital Oliver, Barret Zoph, Behrooz Ghorbani, Ben Leimberger, Ben Rossen, Ben Sokolowsky, Ben Wang, Benjamin Zweig, Beth Hoover, Blake Samic, Bob McGrew, Bobby Spero, Bogo Giertler, Bowen Cheng, Brad Lightcap, Brandon Walkin, Brendan Quinn, Brian Guarraci, Brian Hsu, Bright Kellogg, Brydon
  Eastman, Camillo Lugaresi, Carroll~L. Wainwright, Cary Bassin, Cary Hudson, Casey Chu, Chad Nelson, Chak Li, Chan~Jun Shern, Channing Conger, Charlotte Barette, Chelsea Voss, Chen Ding, Cheng Lu, Chong Zhang, Chris Beaumont, Chris Hallacy, Chris Koch, Christian Gibson, Christina Kim, Christine Choi, Christine McLeavey, Christopher Hesse, Claudia Fischer, Clemens Winter, Coley Czarnecki, Colin Jarvis, Colin Wei, Constantin Koumouzelis, and Dane Sherburn. 2024.
\newblock \href {https://doi.org/10.48550/ARXIV.2410.21276} {Gpt-4o system card}.
\newblock \emph{CoRR}, abs/2410.21276.

\bibitem[{Izacard et~al.(2023)Izacard, Lewis, Lomeli, Hosseini, Petroni, Schick, Dwivedi{-}Yu, Joulin, Riedel, and Grave}]{DBLP:journals/jmlr/IzacardLLHPSDJRG23}
Gautier Izacard, Patrick S.~H. Lewis, Maria Lomeli, Lucas Hosseini, Fabio Petroni, Timo Schick, Jane Dwivedi{-}Yu, Armand Joulin, Sebastian Riedel, and Edouard Grave. 2023.
\newblock \href {http://jmlr.org/papers/v24/23-0037.html} {Atlas: Few-shot learning with retrieval augmented language models}.
\newblock \emph{J. Mach. Learn. Res.}, 24:251:1--251:43.

\bibitem[{Jeong et~al.(2024)Jeong, Baek, Cho, Hwang, and Park}]{DBLP:conf/naacl/JeongBCHP24}
Soyeong Jeong, Jinheon Baek, Sukmin Cho, Sung~Ju Hwang, and Jong Park. 2024.
\newblock \href {https://doi.org/10.18653/V1/2024.NAACL-LONG.389} {Adaptive-rag: Learning to adapt retrieval-augmented large language models through question complexity}.
\newblock In \emph{Proceedings of the 2024 Conference of the North American Chapter of the Association for Computational Linguistics: Human Language Technologies (Volume 1: Long Papers), {NAACL} 2024, Mexico City, Mexico, June 16-21, 2024}, pages 7036--7050. Association for Computational Linguistics.

\bibitem[{Jiang et~al.(2023)Jiang, Xu, Gao, Sun, Liu, Dwivedi{-}Yu, Yang, Callan, and Neubig}]{DBLP:conf/emnlp/JiangXGSLDYCN23}
Zhengbao Jiang, Frank~F. Xu, Luyu Gao, Zhiqing Sun, Qian Liu, Jane Dwivedi{-}Yu, Yiming Yang, Jamie Callan, and Graham Neubig. 2023.
\newblock \href {https://doi.org/10.18653/V1/2023.EMNLP-MAIN.495} {Active retrieval augmented generation}.
\newblock In \emph{Proceedings of the 2023 Conference on Empirical Methods in Natural Language Processing, {EMNLP} 2023, Singapore, December 6-10, 2023}, pages 7969--7992. Association for Computational Linguistics.

\bibitem[{Kandpal et~al.(2023)Kandpal, Deng, Roberts, Wallace, and Raffel}]{DBLP:conf/icml/KandpalDRWR23}
Nikhil Kandpal, Haikang Deng, Adam Roberts, Eric Wallace, and Colin Raffel. 2023.
\newblock \href {https://proceedings.mlr.press/v202/kandpal23a.html} {Large language models struggle to learn long-tail knowledge}.
\newblock In \emph{International Conference on Machine Learning, {ICML} 2023, 23-29 July 2023, Honolulu, Hawaii, {USA}}, volume 202 of \emph{Proceedings of Machine Learning Research}, pages 15696--15707. {PMLR}.

\bibitem[{Karpukhin et~al.(2020)Karpukhin, Oguz, Min, Lewis, Wu, Edunov, Chen, and Yih}]{DBLP:conf/emnlp/KarpukhinOMLWEC20}
Vladimir Karpukhin, Barlas Oguz, Sewon Min, Patrick S.~H. Lewis, Ledell Wu, Sergey Edunov, Danqi Chen, and Wen{-}tau Yih. 2020.
\newblock \href {https://doi.org/10.18653/V1/2020.EMNLP-MAIN.550} {Dense passage retrieval for open-domain question answering}.
\newblock In \emph{Proceedings of the 2020 Conference on Empirical Methods in Natural Language Processing, {EMNLP} 2020, Online, November 16-20, 2020}, pages 6769--6781. Association for Computational Linguistics.

\bibitem[{Ke et~al.(2024)Ke, Kong, Li, Zhang, Mei, and Bendersky}]{DBLP:conf/acl/KeK00MB24}
Zixuan Ke, Weize Kong, Cheng Li, Mingyang Zhang, Qiaozhu Mei, and Michael Bendersky. 2024.
\newblock \href {https://doi.org/10.18653/V1/2024.ACL-LONG.562} {Bridging the preference gap between retrievers and llms}.
\newblock In \emph{Proceedings of the 62nd Annual Meeting of the Association for Computational Linguistics (Volume 1: Long Papers), {ACL} 2024, Bangkok, Thailand, August 11-16, 2024}, pages 10438--10451. Association for Computational Linguistics.

\bibitem[{Kwon et~al.(2023)Kwon, Li, Zhuang, Sheng, Zheng, Yu, Gonzalez, Zhang, and Stoica}]{DBLP:conf/sosp/KwonLZ0ZY0ZS23}
Woosuk Kwon, Zhuohan Li, Siyuan Zhuang, Ying Sheng, Lianmin Zheng, Cody~Hao Yu, Joseph Gonzalez, Hao Zhang, and Ion Stoica. 2023.
\newblock \href {https://doi.org/10.1145/3600006.3613165} {Efficient memory management for large language model serving with pagedattention}.
\newblock In \emph{Proceedings of the 29th Symposium on Operating Systems Principles, {SOSP} 2023, Koblenz, Germany, October 23-26, 2023}, pages 611--626. {ACM}.

\bibitem[{Lewis et~al.(2020)Lewis, Perez, Piktus, Petroni, Karpukhin, Goyal, K{\"{u}}ttler, Lewis, Yih, Rockt{\"{a}}schel, Riedel, and Kiela}]{DBLP:conf/nips/LewisPPPKGKLYR020}
Patrick S.~H. Lewis, Ethan Perez, Aleksandra Piktus, Fabio Petroni, Vladimir Karpukhin, Naman Goyal, Heinrich K{\"{u}}ttler, Mike Lewis, Wen{-}tau Yih, Tim Rockt{\"{a}}schel, Sebastian Riedel, and Douwe Kiela. 2020.
\newblock \href {https://proceedings.neurips.cc/paper/2020/hash/6b493230205f780e1bc26945df7481e5-Abstract.html} {Retrieval-augmented generation for knowledge-intensive {NLP} tasks}.
\newblock In \emph{Advances in Neural Information Processing Systems 33: Annual Conference on Neural Information Processing Systems 2020, NeurIPS 2020, December 6-12, 2020, virtual}.

\bibitem[{Lin et~al.(2024)Lin, Chen, Chen, Shi, Lomeli, James, Rodriguez, Kahn, Szilvasy, Lewis, Zettlemoyer, and Yih}]{DBLP:conf/iclr/Lin0CSL00KSLZY24}
Xi~Victoria Lin, Xilun Chen, Mingda Chen, Weijia Shi, Maria Lomeli, Richard James, Pedro Rodriguez, Jacob Kahn, Gergely Szilvasy, Mike Lewis, Luke Zettlemoyer, and Wen{-}tau Yih. 2024.
\newblock \href {https://openreview.net/forum?id=22OTbutug9} {{RA-DIT:} retrieval-augmented dual instruction tuning}.
\newblock In \emph{The Twelfth International Conference on Learning Representations, {ICLR} 2024, Vienna, Austria, May 7-11, 2024}. OpenReview.net.

\bibitem[{Liu et~al.(2023)Liu, Iter, Xu, Wang, Xu, and Zhu}]{DBLP:conf/emnlp/LiuIXWXZ23}
Yang Liu, Dan Iter, Yichong Xu, Shuohang Wang, Ruochen Xu, and Chenguang Zhu. 2023.
\newblock \href {https://doi.org/10.18653/V1/2023.EMNLP-MAIN.153} {G-eval: {NLG} evaluation using gpt-4 with better human alignment}.
\newblock In \emph{Proceedings of the 2023 Conference on Empirical Methods in Natural Language Processing, {EMNLP} 2023, Singapore, December 6-10, 2023}, pages 2511--2522. Association for Computational Linguistics.

\bibitem[{Lyu et~al.(2024)Lyu, Li, Niu, Xiong, Tang, Wang, Wu, Liu, Xu, and Chen}]{DBLP:journals/corr/abs-2401-17043}
Yuanjie Lyu, Zhiyu Li, Simin Niu, Feiyu Xiong, Bo~Tang, Wenjin Wang, Hao Wu, Huanyong Liu, Tong Xu, and Enhong Chen. 2024.
\newblock \href {https://doi.org/10.48550/ARXIV.2401.17043} {{CRUD-RAG:} {A} comprehensive chinese benchmark for retrieval-augmented generation of large language models}.
\newblock \emph{CoRR}, abs/2401.17043.

\bibitem[{Mallen et~al.(2023)Mallen, Asai, Zhong, Das, Khashabi, and Hajishirzi}]{DBLP:conf/acl/MallenAZDKH23}
Alex Mallen, Akari Asai, Victor Zhong, Rajarshi Das, Daniel Khashabi, and Hannaneh Hajishirzi. 2023.
\newblock \href {https://doi.org/10.18653/V1/2023.ACL-LONG.546} {When not to trust language models: Investigating effectiveness of parametric and non-parametric memories}.
\newblock In \emph{Proceedings of the 61st Annual Meeting of the Association for Computational Linguistics (Volume 1: Long Papers), {ACL} 2023, Toronto, Canada, July 9-14, 2023}, pages 9802--9822. Association for Computational Linguistics.

\bibitem[{Min et~al.(2023)Min, Krishna, Lyu, Lewis, Yih, Koh, Iyyer, Zettlemoyer, and Hajishirzi}]{DBLP:conf/emnlp/MinKLLYKIZH23}
Sewon Min, Kalpesh Krishna, Xinxi Lyu, Mike Lewis, Wen{-}tau Yih, Pang~Wei Koh, Mohit Iyyer, Luke Zettlemoyer, and Hannaneh Hajishirzi. 2023.
\newblock \href {https://doi.org/10.18653/V1/2023.EMNLP-MAIN.741} {Factscore: Fine-grained atomic evaluation of factual precision in long form text generation}.
\newblock In \emph{Proceedings of the 2023 Conference on Empirical Methods in Natural Language Processing, {EMNLP} 2023, Singapore, December 6-10, 2023}, pages 12076--12100. Association for Computational Linguistics.

\bibitem[{Ni et~al.(2022)Ni, Qu, Lu, Dai, {\'{A}}brego, Ma, Zhao, Luan, Hall, Chang, and Yang}]{DBLP:conf/emnlp/Ni0LDAMZLHCY22}
Jianmo Ni, Chen Qu, Jing Lu, Zhuyun Dai, Gustavo~Hern{\'{a}}ndez {\'{A}}brego, Ji~Ma, Vincent~Y. Zhao, Yi~Luan, Keith~B. Hall, Ming{-}Wei Chang, and Yinfei Yang. 2022.
\newblock \href {https://doi.org/10.18653/V1/2022.EMNLP-MAIN.669} {Large dual encoders are generalizable retrievers}.
\newblock In \emph{Proceedings of the 2022 Conference on Empirical Methods in Natural Language Processing, {EMNLP} 2022, Abu Dhabi, United Arab Emirates, December 7-11, 2022}, pages 9844--9855. Association for Computational Linguistics.

\bibitem[{OpenAI(2023)}]{DBLP:journals/corr/abs-2303-08774}
OpenAI. 2023.
\newblock \href {https://doi.org/10.48550/ARXIV.2303.08774} {{GPT-4} technical report}.
\newblock \emph{CoRR}, abs/2303.08774.

\bibitem[{Press et~al.(2023)Press, Zhang, Min, Schmidt, Smith, and Lewis}]{DBLP:conf/emnlp/PressZMSSL23}
Ofir Press, Muru Zhang, Sewon Min, Ludwig Schmidt, Noah~A. Smith, and Mike Lewis. 2023.
\newblock \href {https://doi.org/10.18653/V1/2023.FINDINGS-EMNLP.378} {Measuring and narrowing the compositionality gap in language models}.
\newblock In \emph{Findings of the Association for Computational Linguistics: {EMNLP} 2023, Singapore, December 6-10, 2023}, pages 5687--5711. Association for Computational Linguistics.

\bibitem[{Rafailov et~al.(2023)Rafailov, Sharma, Mitchell, Manning, Ermon, and Finn}]{DBLP:conf/nips/RafailovSMMEF23}
Rafael Rafailov, Archit Sharma, Eric Mitchell, Christopher~D. Manning, Stefano Ermon, and Chelsea Finn. 2023.
\newblock \href {http://papers.nips.cc/paper\_files/paper/2023/hash/a85b405ed65c6477a4fe8302b5e06ce7-Abstract-Conference.html} {Direct preference optimization: Your language model is secretly a reward model}.
\newblock In \emph{Advances in Neural Information Processing Systems 36: Annual Conference on Neural Information Processing Systems 2023, NeurIPS 2023, New Orleans, LA, USA, December 10 - 16, 2023}.

\bibitem[{Ram et~al.(2023)Ram, Levine, Dalmedigos, Muhlgay, Shashua, Leyton{-}Brown, and Shoham}]{DBLP:journals/tacl/RamLDMSLS23}
Ori Ram, Yoav Levine, Itay Dalmedigos, Dor Muhlgay, Amnon Shashua, Kevin Leyton{-}Brown, and Yoav Shoham. 2023.
\newblock \href {https://doi.org/10.1162/TACL\_A\_00605} {In-context retrieval-augmented language models}.
\newblock \emph{Trans. Assoc. Comput. Linguistics}, 11:1316--1331.

\bibitem[{Shultz et~al.(2024)Shultz, Wise, and Nobandegani}]{DBLP:journals/corr/abs-2403-17196}
Thomas~R. Shultz, Jamie Wise, and Ardavan~Salehi Nobandegani. 2024.
\newblock \href {https://doi.org/10.48550/ARXIV.2403.17196} {{GPT-4} understands discourse at least as well as humans do}.
\newblock \emph{CoRR}, abs/2403.17196.

\bibitem[{Siriwardhana et~al.(2023)Siriwardhana, Weerasekera, Kaluarachchi, Wen, Rana, and Nanayakkara}]{DBLP:journals/tacl/SiriwardhanaWKWRN23}
Shamane Siriwardhana, Rivindu Weerasekera, Tharindu Kaluarachchi, Elliott Wen, Rajib Rana, and Suranga Nanayakkara. 2023.
\newblock \href {https://doi.org/10.1162/TACL\_A\_00530} {Improving the domain adaptation of retrieval augmented generation {(RAG)} models for open domain question answering}.
\newblock \emph{Trans. Assoc. Comput. Linguistics}, 11:1--17.

\bibitem[{Srivastava et~al.(2023)Srivastava, Rastogi, Rao, Shoeb, Abid, Fisch, Brown, Santoro, Gupta, Garriga{-}Alonso, Kluska, Lewkowycz, Agarwal, Power, Ray, Warstadt, Kocurek, Safaya, Tazarv, Xiang, Parrish, Nie, Hussain, Askell, Dsouza, Slone, Rahane, Iyer, Andreassen, Madotto, Santilli, Stuhlm{\"{u}}ller, Dai, La, Lampinen, Zou, Jiang, Chen, Vuong, Gupta, Gottardi, Norelli, Venkatesh, Gholamidavoodi, Tabassum, Menezes, Kirubarajan, Mullokandov, Sabharwal, Herrick, Efrat, Erdem, Karakas, Roberts, Loe, Zoph, Bojanowski, {\"{O}}zyurt, Hedayatnia, Neyshabur, Inden, Stein, Ekmekci, Lin, Howald, Orinion, Diao, Dour, Stinson, Argueta, Ram{\'{\i}}rez, Singh, Rathkopf, Meng, Baral, Wu, Callison{-}Burch, Waites, Voigt, Manning, Potts, Ramirez, Rivera, Siro, Raffel, Ashcraft, Garbacea, Sileo, Garrette, Hendrycks, Kilman, Roth, Freeman, Khashabi, Levy, Gonz{\'{a}}lez, Perszyk, Hernandez, Chen, Ippolito, Gilboa, Dohan, Drakard, Jurgens, Datta, Ganguli, Emelin, Kleyko, Yuret, Chen, Tam, Hupkes, Misra, Buzan, Mollo,
  Yang, Lee, Schrader, Shutova, Cubuk, Segal, Hagerman, Barnes, Donoway, Pavlick, Rodol{\`{a}}, Lam, Chu, Tang, Erdem, Chang, Chi, Dyer, Jerzak, Kim, Manyasi, Zheltonozhskii, Xia, Siar, Mart{\'{\i}}nez{-}Plumed, Happ{\'{e}}, Chollet, Rong, Mishra, Winata, de~Melo, Kruszewski, Parascandolo, Mariani, Wang, Jaimovitch{-}L{\'{o}}pez, Betz, Gur{-}Ari, Galijasevic, Kim, Rashkin, Hajishirzi, Mehta, Bogar, Shevlin, Sch{\"{u}}tze, Yakura, Zhang, Wong, Ng, Noble, Jumelet, Geissinger, Kernion, Hilton, Lee, Fisac, Simon, Koppel, Zheng, Zou, Kocon, Thompson, Wingfield, Kaplan, Radom, Sohl{-}Dickstein, Phang, Wei, Yosinski, Novikova, Bosscher, Marsh, Kim, Taal, Engel, Alabi, Xu, Song, Tang, Waweru, Burden, Miller, Balis, Batchelder, Berant, Frohberg, Rozen, Hern{\'{a}}ndez{-}Orallo, Boudeman, Guerr, Jones, Tenenbaum, Rule, Chua, Kanclerz, Livescu, Krauth, Gopalakrishnan, Ignatyeva, Markert, Dhole, Gimpel, Omondi, Mathewson, Chiafullo, Shkaruta, Shridhar, McDonell, Richardson, Reynolds, Gao, Zhang, Dugan, Qin, Ochando,
  Morency, Moschella, Lam, Noble, Schmidt, He, Col{\'{o}}n, Metz, Senel, Bosma, Sap, ter Hoeve, Farooqi, Faruqui, Mazeika, Baturan, Marelli, Maru, Ram{\'{\i}}rez{-}Quintana, Tolkiehn, Giulianelli, Lewis, Potthast, Leavitt, Hagen, Schubert, Baitemirova, Arnaud, McElrath, Yee, Cohen, Gu, Ivanitskiy, Starritt, Strube, Swedrowski, Bevilacqua, Yasunaga, Kale, Cain, Xu, Suzgun, Walker, Tiwari, Bansal, Aminnaseri, Geva, Gheini, T., Peng, Chi, Lee, Krakover, Cameron, Roberts, Doiron, Martinez, Nangia, Deckers, Muennighoff, Keskar, Iyer, Constant, Fiedel, Wen, Zhang, Agha, Elbaghdadi, Levy, Evans, Casares, Doshi, Fung, Liang, Vicol, Alipoormolabashi, Liao, Liang, Chang, Eckersley, Htut, Hwang, Milkowski, Patil, Pezeshkpour, Oli, Mei, Lyu, Chen, Banjade, Rudolph, Gabriel, Habacker, Risco, Milli{\`{e}}re, Garg, Barnes, Saurous, Arakawa, Raymaekers, Frank, Sikand, Novak, Sitelew, LeBras, Liu, Jacobs, Zhang, Salakhutdinov, Chi, Lee, Stovall, Teehan, Yang, Singh, Mohammad, Anand, Dillavou, Shleifer, Wiseman, Gruetter,
  Bowman, Schoenholz, Han, Kwatra, Rous, Ghazarian, Ghosh, Casey, Bischoff, Gehrmann, Schuster, Sadeghi, Hamdan, Zhou, Srivastava, Shi, Singh, Asaadi, Gu, Pachchigar, Toshniwal, Upadhyay, Debnath, Shakeri, Thormeyer, Melzi, Reddy, Makini, Lee, Torene, Hatwar, Dehaene, Divic, Ermon, Biderman, Lin, Prasad, Piantadosi, Shieber, Misherghi, Kiritchenko, Mishra, Linzen, Schuster, Li, Yu, Ali, Hashimoto, Wu, Desbordes, Rothschild, Phan, Wang, Nkinyili, Schick, Kornev, Tunduny, Gerstenberg, Chang, Neeraj, Khot, Shultz, Shaham, Misra, Demberg, Nyamai, Raunak, Ramasesh, Prabhu, Padmakumar, Srikumar, Fedus, Saunders, Zhang, Vossen, Ren, Tong, Zhao, Wu, Shen, Yaghoobzadeh, Lakretz, Song, Bahri, Choi, Yang, Hao, Chen, Belinkov, Hou, Hou, Bai, Seid, Zhao, Wang, Wang, Wang, and Wu}]{DBLP:journals/tmlr/SrivastavaRRSAF23}
Aarohi Srivastava, Abhinav Rastogi, Abhishek Rao, Abu Awal~Md Shoeb, Abubakar Abid, Adam Fisch, Adam~R. Brown, Adam Santoro, Aditya Gupta, Adri{\`{a}} Garriga{-}Alonso, Agnieszka Kluska, Aitor Lewkowycz, Akshat Agarwal, Alethea Power, Alex Ray, Alex Warstadt, Alexander~W. Kocurek, Ali Safaya, Ali Tazarv, Alice Xiang, Alicia Parrish, Allen Nie, Aman Hussain, Amanda Askell, Amanda Dsouza, Ambrose Slone, Ameet Rahane, Anantharaman~S. Iyer, Anders Andreassen, Andrea Madotto, Andrea Santilli, Andreas Stuhlm{\"{u}}ller, Andrew~M. Dai, Andrew La, Andrew~K. Lampinen, Andy Zou, Angela Jiang, Angelica Chen, Anh Vuong, Animesh Gupta, Anna Gottardi, Antonio Norelli, Anu Venkatesh, Arash Gholamidavoodi, Arfa Tabassum, Arul Menezes, Arun Kirubarajan, Asher Mullokandov, Ashish Sabharwal, Austin Herrick, Avia Efrat, Aykut Erdem, Ayla Karakas, B.~Ryan Roberts, Bao~Sheng Loe, Barret Zoph, Bartlomiej Bojanowski, Batuhan {\"{O}}zyurt, Behnam Hedayatnia, Behnam Neyshabur, Benjamin Inden, Benno Stein, Berk Ekmekci, Bill~Yuchen
  Lin, Blake Howald, Bryan Orinion, Cameron Diao, Cameron Dour, Catherine Stinson, Cedrick Argueta, C{\`{e}}sar~Ferri Ram{\'{\i}}rez, Chandan Singh, Charles Rathkopf, Chenlin Meng, Chitta Baral, Chiyu Wu, Chris Callison{-}Burch, Chris Waites, Christian Voigt, Christopher~D. Manning, Christopher Potts, Cindy Ramirez, Clara~E. Rivera, Clemencia Siro, Colin Raffel, Courtney Ashcraft, Cristina Garbacea, Damien Sileo, Dan Garrette, Dan Hendrycks, Dan Kilman, Dan Roth, Daniel Freeman, Daniel Khashabi, Daniel Levy, Daniel~Mosegu{\'{\i}} Gonz{\'{a}}lez, Danielle Perszyk, Danny Hernandez, Danqi Chen, Daphne Ippolito, Dar Gilboa, David Dohan, David Drakard, David Jurgens, Debajyoti Datta, Deep Ganguli, Denis Emelin, Denis Kleyko, Deniz Yuret, Derek Chen, Derek Tam, Dieuwke Hupkes, Diganta Misra, Dilyar Buzan, Dimitri~Coelho Mollo, Diyi Yang, Dong{-}Ho Lee, Dylan Schrader, Ekaterina Shutova, Ekin~Dogus Cubuk, Elad Segal, Eleanor Hagerman, Elizabeth Barnes, Elizabeth Donoway, Ellie Pavlick, Emanuele Rodol{\`{a}}, Emma
  Lam, Eric Chu, Eric Tang, Erkut Erdem, Ernie Chang, Ethan~A. Chi, Ethan Dyer, Ethan~J. Jerzak, Ethan Kim, Eunice~Engefu Manyasi, Evgenii Zheltonozhskii, Fanyue Xia, Fatemeh Siar, Fernando Mart{\'{\i}}nez{-}Plumed, Francesca Happ{\'{e}}, Fran{\c{c}}ois Chollet, Frieda Rong, Gaurav Mishra, Genta~Indra Winata, Gerard de~Melo, Germ{\'{a}}n Kruszewski, Giambattista Parascandolo, Giorgio Mariani, Gloria Wang, Gonzalo Jaimovitch{-}L{\'{o}}pez, Gregor Betz, Guy Gur{-}Ari, Hana Galijasevic, Hannah Kim, Hannah Rashkin, Hannaneh Hajishirzi, Harsh Mehta, Hayden Bogar, Henry Shevlin, Hinrich Sch{\"{u}}tze, Hiromu Yakura, Hongming Zhang, Hugh~Mee Wong, Ian Ng, Isaac Noble, Jaap Jumelet, Jack Geissinger, Jackson Kernion, Jacob Hilton, Jaehoon Lee, Jaime~Fern{\'{a}}ndez Fisac, James~B. Simon, James Koppel, James Zheng, James Zou, Jan Kocon, Jana Thompson, Janelle Wingfield, Jared Kaplan, Jarema Radom, Jascha Sohl{-}Dickstein, Jason Phang, Jason Wei, Jason Yosinski, Jekaterina Novikova, Jelle Bosscher, Jennifer Marsh,
  Jeremy Kim, Jeroen Taal, Jesse~H. Engel, Jesujoba Alabi, Jiacheng Xu, Jiaming Song, Jillian Tang, Joan Waweru, John Burden, John Miller, John~U. Balis, Jonathan Batchelder, Jonathan Berant, J{\"{o}}rg Frohberg, Jos Rozen, Jos{\'{e}} Hern{\'{a}}ndez{-}Orallo, Joseph Boudeman, Joseph Guerr, Joseph Jones, Joshua~B. Tenenbaum, Joshua~S. Rule, Joyce Chua, Kamil Kanclerz, Karen Livescu, Karl Krauth, Karthik Gopalakrishnan, Katerina Ignatyeva, Katja Markert, Kaustubh~D. Dhole, Kevin Gimpel, Kevin Omondi, Kory~W. Mathewson, Kristen Chiafullo, Ksenia Shkaruta, Kumar Shridhar, Kyle McDonell, Kyle Richardson, Laria Reynolds, Leo Gao, Li~Zhang, Liam Dugan, Lianhui Qin, Lidia~Contreras Ochando, Louis{-}Philippe Morency, Luca Moschella, Lucas Lam, Lucy Noble, Ludwig Schmidt, Luheng He, Luis~Oliveros Col{\'{o}}n, Luke Metz, L{\"{u}}tfi~Kerem Senel, Maarten Bosma, Maarten Sap, Maartje ter Hoeve, Maheen Farooqi, Manaal Faruqui, Mantas Mazeika, Marco Baturan, Marco Marelli, Marco Maru, Mar{\'{\i}}a~Jos{\'{e}}
  Ram{\'{\i}}rez{-}Quintana, Marie Tolkiehn, Mario Giulianelli, Martha Lewis, Martin Potthast, Matthew~L. Leavitt, Matthias Hagen, M{\'{a}}ty{\'{a}}s Schubert, Medina Baitemirova, Melody Arnaud, Melvin McElrath, Michael~A. Yee, Michael Cohen, Michael Gu, Michael~I. Ivanitskiy, Michael Starritt, Michael Strube, Michal Swedrowski, Michele Bevilacqua, Michihiro Yasunaga, Mihir Kale, Mike Cain, Mimee Xu, Mirac Suzgun, Mitch Walker, Mo~Tiwari, Mohit Bansal, Moin Aminnaseri, Mor Geva, Mozhdeh Gheini, Mukund~Varma T., Nanyun Peng, Nathan~A. Chi, Nayeon Lee, Neta~Gur{-}Ari Krakover, Nicholas Cameron, Nicholas Roberts, Nick Doiron, Nicole Martinez, Nikita Nangia, Niklas Deckers, Niklas Muennighoff, Nitish~Shirish Keskar, Niveditha Iyer, Noah Constant, Noah Fiedel, Nuan Wen, Oliver Zhang, Omar Agha, Omar Elbaghdadi, Omer Levy, Owain Evans, Pablo Antonio~Moreno Casares, Parth Doshi, Pascale Fung, Paul~Pu Liang, Paul Vicol, Pegah Alipoormolabashi, Peiyuan Liao, Percy Liang, Peter Chang, Peter Eckersley, Phu~Mon Htut,
  Pinyu Hwang, Piotr Milkowski, Piyush Patil, Pouya Pezeshkpour, Priti Oli, Qiaozhu Mei, Qing Lyu, Qinlang Chen, Rabin Banjade, Rachel~Etta Rudolph, Raefer Gabriel, Rahel Habacker, Ramon Risco, Rapha{\"{e}}l Milli{\`{e}}re, Rhythm Garg, Richard Barnes, Rif~A. Saurous, Riku Arakawa, Robbe Raymaekers, Robert Frank, Rohan Sikand, Roman Novak, Roman Sitelew, Ronan LeBras, Rosanne Liu, Rowan Jacobs, Rui Zhang, Ruslan Salakhutdinov, Ryan Chi, Ryan Lee, Ryan Stovall, Ryan Teehan, Rylan Yang, Sahib Singh, Saif~M. Mohammad, Sajant Anand, Sam Dillavou, Sam Shleifer, Sam Wiseman, Samuel Gruetter, Samuel~R. Bowman, Samuel~S. Schoenholz, Sanghyun Han, Sanjeev Kwatra, Sarah~A. Rous, Sarik Ghazarian, Sayan Ghosh, Sean Casey, Sebastian Bischoff, Sebastian Gehrmann, Sebastian Schuster, Sepideh Sadeghi, Shadi Hamdan, Sharon Zhou, Shashank Srivastava, Sherry Shi, Shikhar Singh, Shima Asaadi, Shixiang~Shane Gu, Shubh Pachchigar, Shubham Toshniwal, Shyam Upadhyay, Shyamolima~(Shammie) Debnath, Siamak Shakeri, Simon Thormeyer,
  Simone Melzi, Siva Reddy, Sneha~Priscilla Makini, Soo{-}Hwan Lee, Spencer Torene, Sriharsha Hatwar, Stanislas Dehaene, Stefan Divic, Stefano Ermon, Stella Biderman, Stephanie Lin, Stephen Prasad, Steven~T. Piantadosi, Stuart~M. Shieber, Summer Misherghi, Svetlana Kiritchenko, Swaroop Mishra, Tal Linzen, Tal Schuster, Tao Li, Tao Yu, Tariq Ali, Tatsu Hashimoto, Te{-}Lin Wu, Th{\'{e}}o Desbordes, Theodore Rothschild, Thomas Phan, Tianle Wang, Tiberius Nkinyili, Timo Schick, Timofei Kornev, Titus Tunduny, Tobias Gerstenberg, Trenton Chang, Trishala Neeraj, Tushar Khot, Tyler Shultz, Uri Shaham, Vedant Misra, Vera Demberg, Victoria Nyamai, Vikas Raunak, Vinay~V. Ramasesh, Vinay~Uday Prabhu, Vishakh Padmakumar, Vivek Srikumar, William Fedus, William Saunders, William Zhang, Wout Vossen, Xiang Ren, Xiaoyu Tong, Xinran Zhao, Xinyi Wu, Xudong Shen, Yadollah Yaghoobzadeh, Yair Lakretz, Yangqiu Song, Yasaman Bahri, Yejin Choi, Yichi Yang, Yiding Hao, Yifu Chen, Yonatan Belinkov, Yu~Hou, Yufang Hou, Yuntao Bai,
  Zachary Seid, Zhuoye Zhao, Zijian Wang, Zijie~J. Wang, Zirui Wang, and Ziyi Wu. 2023.
\newblock \href {https://openreview.net/forum?id=uyTL5Bvosj} {Beyond the imitation game: Quantifying and extrapolating the capabilities of language models}.
\newblock \emph{Trans. Mach. Learn. Res.}, 2023.

\bibitem[{Stelmakh et~al.(2022)Stelmakh, Luan, Dhingra, and Chang}]{DBLP:conf/emnlp/StelmakhLDC22}
Ivan Stelmakh, Yi~Luan, Bhuwan Dhingra, and Ming{-}Wei Chang. 2022.
\newblock \href {https://doi.org/10.18653/V1/2022.EMNLP-MAIN.566} {{ASQA:} factoid questions meet long-form answers}.
\newblock In \emph{Proceedings of the 2022 Conference on Empirical Methods in Natural Language Processing, {EMNLP} 2022, Abu Dhabi, United Arab Emirates, December 7-11, 2022}, pages 8273--8288. Association for Computational Linguistics.

\bibitem[{Touvron et~al.(2023)Touvron, Lavril, Izacard, Martinet, Lachaux, Lacroix, Rozi{\`{e}}re, Goyal, Hambro, Azhar, Rodriguez, Joulin, Grave, and Lample}]{DBLP:journals/corr/abs-2302-13971}
Hugo Touvron, Thibaut Lavril, Gautier Izacard, Xavier Martinet, Marie{-}Anne Lachaux, Timoth{\'{e}}e Lacroix, Baptiste Rozi{\`{e}}re, Naman Goyal, Eric Hambro, Faisal Azhar, Aur{\'{e}}lien Rodriguez, Armand Joulin, Edouard Grave, and Guillaume Lample. 2023.
\newblock \href {https://doi.org/10.48550/ARXIV.2302.13971} {Llama: Open and efficient foundation language models}.
\newblock \emph{CoRR}, abs/2302.13971.

\bibitem[{Trivedi et~al.(2022)Trivedi, Balasubramanian, Khot, and Sabharwal}]{DBLP:journals/tacl/TrivediBKS22}
Harsh Trivedi, Niranjan Balasubramanian, Tushar Khot, and Ashish Sabharwal. 2022.
\newblock \href {https://doi.org/10.1162/TACL\_A\_00475} {Musique: Multihop questions via single-hop question composition}.
\newblock \emph{Trans. Assoc. Comput. Linguistics}, 10:539--554.

\bibitem[{Trivedi et~al.(2023)Trivedi, Balasubramanian, Khot, and Sabharwal}]{DBLP:conf/acl/TrivediBKS23}
Harsh Trivedi, Niranjan Balasubramanian, Tushar Khot, and Ashish Sabharwal. 2023.
\newblock \href {https://doi.org/10.18653/V1/2023.ACL-LONG.557} {Interleaving retrieval with chain-of-thought reasoning for knowledge-intensive multi-step questions}.
\newblock In \emph{Proceedings of the 61st Annual Meeting of the Association for Computational Linguistics (Volume 1: Long Papers), {ACL} 2023, Toronto, Canada, July 9-14, 2023}, pages 10014--10037. Association for Computational Linguistics.

\bibitem[{Voorhees(1999)}]{DBLP:conf/trec/Voorhees99}
Ellen~M. Voorhees. 1999.
\newblock \href {http://trec.nist.gov/pubs/trec8/papers/qa\_report.pdf} {The {TREC-8} question answering track report}.
\newblock In \emph{Proceedings of The Eighth Text REtrieval Conference, {TREC} 1999, Gaithersburg, Maryland, USA, November 17-19, 1999}, volume 500-246 of \emph{{NIST} Special Publication}. National Institute of Standards and Technology {(NIST)}.

\bibitem[{Vu and Moschitti(2020)}]{vu2020ava}
Thuy Vu and Alessandro Moschitti. 2020.
\newblock Ava: an automatic evaluation approach to question answering systems.
\newblock \emph{arXiv preprint arXiv:2005.00705}.

\bibitem[{Wang et~al.(2024)Wang, Liu, Lin, Li, Ma, and Liang}]{DBLP:journals/corr/abs-2403-05313}
Zihao Wang, Anji Liu, Haowei Lin, Jiaqi Li, Xiaojian Ma, and Yitao Liang. 2024.
\newblock \href {https://doi.org/10.48550/ARXIV.2403.05313} {{RAT:} retrieval augmented thoughts elicit context-aware reasoning in long-horizon generation}.
\newblock \emph{CoRR}, abs/2403.05313.

\bibitem[{Xu et~al.(2024)Xu, Sun, Zheng, Geng, Zhao, Feng, Tao, Lin, and Jiang}]{DBLP:conf/iclr/XuSZG0FTLJ24}
Can Xu, Qingfeng Sun, Kai Zheng, Xiubo Geng, Pu~Zhao, Jiazhan Feng, Chongyang Tao, Qingwei Lin, and Daxin Jiang. 2024.
\newblock \href {https://openreview.net/forum?id=CfXh93NDgH} {Wizardlm: Empowering large pre-trained language models to follow complex instructions}.
\newblock In \emph{The Twelfth International Conference on Learning Representations, {ICLR} 2024, Vienna, Austria, May 7-11, 2024}. OpenReview.net.

\bibitem[{Yang et~al.(2024)Yang, Yang, Zhang, Hui, Zheng, Yu, Li, Liu, Huang, Wei, Lin, Yang, Tu, Zhang, Yang, Yang, Zhou, Lin, Dang, Lu, Bao, Yang, Yu, Li, Xue, Zhang, Zhu, Men, Lin, Li, Xia, Ren, Ren, Fan, Su, Zhang, Wan, Liu, Cui, Zhang, and Qiu}]{DBLP:journals/corr/abs-2412-15115}
An~Yang, Baosong Yang, Beichen Zhang, Binyuan Hui, Bo~Zheng, Bowen Yu, Chengyuan Li, Dayiheng Liu, Fei Huang, Haoran Wei, Huan Lin, Jian Yang, Jianhong Tu, Jianwei Zhang, Jianxin Yang, Jiaxi Yang, Jingren Zhou, Junyang Lin, Kai Dang, Keming Lu, Keqin Bao, Kexin Yang, Le~Yu, Mei Li, Mingfeng Xue, Pei Zhang, Qin Zhu, Rui Men, Runji Lin, Tianhao Li, Tingyu Xia, Xingzhang Ren, Xuancheng Ren, Yang Fan, Yang Su, Yichang Zhang, Yu~Wan, Yuqiong Liu, Zeyu Cui, Zhenru Zhang, and Zihan Qiu. 2024.
\newblock \href {https://doi.org/10.48550/ARXIV.2412.15115} {Qwen2.5 technical report}.
\newblock \emph{CoRR}, abs/2412.15115.

\bibitem[{Yang et~al.(2018)Yang, Qi, Zhang, Bengio, Cohen, Salakhutdinov, and Manning}]{DBLP:conf/emnlp/Yang0ZBCSM18}
Zhilin Yang, Peng Qi, Saizheng Zhang, Yoshua Bengio, William~W. Cohen, Ruslan Salakhutdinov, and Christopher~D. Manning. 2018.
\newblock \href {https://doi.org/10.18653/V1/D18-1259} {Hotpotqa: {A} dataset for diverse, explainable multi-hop question answering}.
\newblock In \emph{Proceedings of the 2018 Conference on Empirical Methods in Natural Language Processing, Brussels, Belgium, October 31 - November 4, 2018}, pages 2369--2380. Association for Computational Linguistics.

\bibitem[{Yao et~al.(2023)Yao, Zhao, Yu, Du, Shafran, Narasimhan, and Cao}]{DBLP:conf/iclr/YaoZYDSN023}
Shunyu Yao, Jeffrey Zhao, Dian Yu, Nan Du, Izhak Shafran, Karthik~R. Narasimhan, and Yuan Cao. 2023.
\newblock \href {https://openreview.net/forum?id=WE\_vluYUL-X} {React: Synergizing reasoning and acting in language models}.
\newblock In \emph{The Eleventh International Conference on Learning Representations, {ICLR} 2023, Kigali, Rwanda, May 1-5, 2023}. OpenReview.net.

\bibitem[{Yin et~al.(2023{\natexlab{a}})Yin, Ye, Liu, Ren, and Sch{\"u}tze}]{yin2023llm}
Wenpeng Yin, Qinyuan Ye, Pengfei Liu, Xiang Ren, and Hinrich Sch{\"u}tze. 2023{\natexlab{a}}.
\newblock Llm-driven instruction following: Progresses and concerns.
\newblock In \emph{Proceedings of the 2023 Conference on Empirical Methods in Natural Language Processing: Tutorial Abstracts}, pages 19--25.

\bibitem[{Yin et~al.(2023{\natexlab{b}})Yin, Sun, Guo, Wu, Qiu, and Huang}]{DBLP:conf/acl/YinSGWQH23}
Zhangyue Yin, Qiushi Sun, Qipeng Guo, Jiawen Wu, Xipeng Qiu, and Xuanjing Huang. 2023{\natexlab{b}}.
\newblock \href {https://doi.org/10.18653/V1/2023.FINDINGS-ACL.551} {Do large language models know what they don't know?}
\newblock In \emph{Findings of the Association for Computational Linguistics: {ACL} 2023, Toronto, Canada, July 9-14, 2023}, pages 8653--8665. Association for Computational Linguistics.

\bibitem[{Yu et~al.(2023)Yu, Zhang, Pan, Ma, Wang, and Yu}]{DBLP:journals/corr/abs-2311-09210}
Wenhao Yu, Hongming Zhang, Xiaoman Pan, Kaixin Ma, Hongwei Wang, and Dong Yu. 2023.
\newblock \href {https://doi.org/10.48550/ARXIV.2311.09210} {Chain-of-note: Enhancing robustness in retrieval-augmented language models}.
\newblock \emph{CoRR}, abs/2311.09210.

\bibitem[{Zuccon et~al.(2023)Zuccon, Koopman, and Shaik}]{DBLP:conf/sigir-ap/ZucconKS23}
Guido Zuccon, Bevan Koopman, and Razia Shaik. 2023.
\newblock \href {https://doi.org/10.1145/3624918.3625329} {Chatgpt hallucinates when attributing answers}.
\newblock In \emph{Annual International {ACM} {SIGIR} Conference on Research and Development in Information Retrieval in the Asia Pacific Region, {SIGIR-AP} 2023, Beijing, China, November 26-28, 2023}, pages 46--51. {ACM}.

\end{thebibliography}
\clearpage

\appendix
\section*{Appendix}
\label{sec:appendix}

\startcontents[sections]
\printcontents[sections]{l}{1}{\setcounter{tocdepth}{2}}

\section{Additional Experimental Results}
\label{Appendix A: Additional Experimental Results}

\subsection{Ablation Study}
\label{appsubsec:Ablation Study}
Table~\ref{tab-app:All results of Ablation Study} presents more ablation results across all models and datasets. We observe that on complex QA datasets (including multi-hop and long-form QA tasks), the performance with adaptive retrieval significantly surpasses that without adaptive retrieval, confirming the effectiveness of our note-centric adaptive retrieval. However, on the simpler StrategyQA dataset, the advantage diminishes, as straightforward reasoning tasks inherently require less retrieval.

\subsection{Adaptive Hyper-Parameter Analysis}
\label{appsubsec:Adaptive Parameter Analysis}
In Table~\ref{tab-app:Adaptive Parameter Analysis of DeepNote}, we present the impact of different hyper-parameters on DeepNote's performance across all datasets and models. We employ six sets of hyper-parameters,  $\left\{ \text{max step}, \text{max failure} \right \}=\left\{ (1,1),(2,1),(2,2),(3,1),(3,2),(3,3) \right \}$.  It is worth mentioning that the max failure value cannot exceed the max step value, as having failure updates exceed the total iteration threshold would render the max failure meaningless.
In Table~\ref{tab-app:Adaptive Parameter Analysis of DeepNote}, we observe conclusions similar to those in Figure~\ref{fig:Adaptive Parameter Analysis}. Increasing either max failure or max step can encourage the model to potentially perform deeper retrieval. Comparing the results of the $(2,2)$ and $(3,1)$ hyper-parameter sets, we find that $(2,2)$ often outperforms $(3,1)$ as reaching the max failure limit terminates the iteration, rendering an excessively high max step ineffective. Therefore, we recommend researchers use values for max failure and max step that are close to each other when running DeepNote.

Additionally, we find that models trained with DPO tend to achieve higher performance with smaller hyper-parameter settings. This is partly because the initial iteration of deep exploration typically yields the highest returns, with diminishing marginal gains as exploration continues. Furthermore, since our training data is derived from $\tau_{0}$ and $\tau_{1}$, the model effectively learns how to better explore the knowledge base in the early stages.

\subsection{Knowledge Density Analysis}

\begin{figure*}[ht!]
\includegraphics[width=\textwidth]{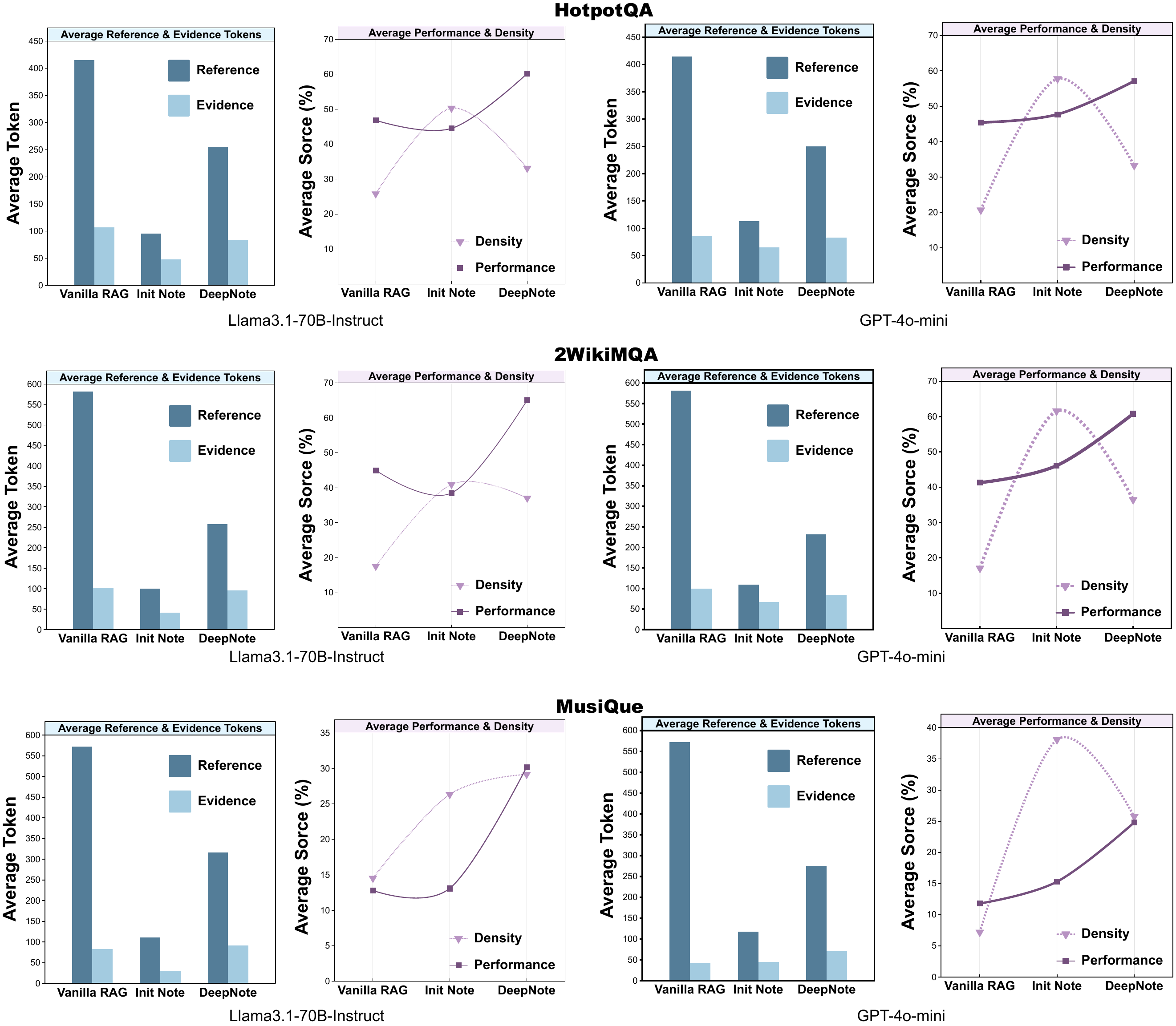}
\caption{\textbf{All results (\%) of knowledge density analysis.}}
\label{fig-app:All results of knowledge density analysis}
\end{figure*}

Figure~\ref{fig-app:All results of knowledge density analysis} presents additional results on knowledge density analysis. The trends and conclusions are consistent with those in Figure~\ref{fig:row-Knowledge Density Comparision on Llama3.1-70B-Instruct}.

\subsection{Impact of Different Top-\texorpdfstring{$k$}{k} Values and Retrievers}
\label{appsubsec:Impact of Different Top-k}
The top-$k$ and retriever settings significantly impact the overall performance of RAG systems. In Table~\ref{tab:Overall Performance}, we have already presented the main results of DeepNote based on the top-5 settings and the BM25-based retriever. Here, we further investigate the performance of DeepNote under different top-$k$ settings and evaluate its performance on two mainstream types of retrievers.

Table~\ref{tab-app:Top-k. GPT-4o-mini} and ~\ref{tab-app:Top-k. Llama3.1-70B-Instruct} present the performance of DeepNote with different top-$k$ settings. The results show that on complex datasets, using a higher top-$k$ (i.e., top-7) leads to better performance. On relatively simple commonsense QA datasets, top-5 achieves the best results. This indicates that complex datasets have higher and more intricate retrieval demands. Additionally, across various top-$k$ settings, DeepNote significantly outperforms Vanilla RAG, demonstrating its robustness.

For different retrievers, the results in Table~\ref{tab-app: Impacts of Different retrievers. Llama3.1-70B-Instruct} reveal that using dense retrievers achieves higher performance.
Overall, DeepNote's performance is similar using both types of retrievers, confirming the robustness of our framework.


\section{Prompt Details}
\label{Appendix B: Prompt Details}
In this section, we present all the prompts used in our framework.
\subsection{Prompts for Inference}
\label{Appendix B.1: Prompt for Inference}
For prompt in the inference stage,
we present the prompts used in all three key processes: note initialization~(Table~\ref{tab-app:Prompt of the Note Initialization Stage}), note-centric adaptive retrieval, and note-informed answer generation. The note-centric adaptive retrieval process consists of multiple stages, including the Query Refinement Stage~(Table~\ref{tab-app:Prompt of the Query Refinement Stage}), Knowledge Accumulation Stage~(Table~\ref{tab-app:Prompt of the Knowledge Accumulation Stage}), and Adaptive Retrieval Decision Stage~(Table~\ref{tab-app:Prompt of the Adaptive Retrieval Decision Stage}).
In addition, due to the varied output style (e.g., long- or short-form generations) of different QA tasks, we tailor the prompts to be task-oriented. 
For example, multi-hop QA tasks require short and precise outputs, often only a few words, while the knowledge in the best note appears as a long text. Therefore, we guide the LLM to output only key answers without including extraneous words~(Table~\ref{Prompts of the Note-Informed Answer Generation Process (Multi-hop QA)}). For the long-form QA task, we guide the response style instead of stringent limitations~(Table~\ref{tab-app:Prompts of the Note-Informed Answer Generation Process (ASQA)}).
Additionally, since StrategyQA requires the system to provide binary answers (Yes/No), our prompt instructs the model to output only Yes or No as the response~(Table~\ref{tab-app:Prompts of the Prompts of the Note-Informed Answer Generation Process (StrategyQA)}).

\subsection{Prompts for DPO}
Only constructing Note Initialization Data~(Table~\ref{tab-app:Prompts of Note Initialization Stage for DPO}) and Query Refinement Data~(Table~\ref{tab-app:Prompts of Query Refinement Stage for DPO}) require additional prompts. In building Knowledge Accumulation Data, we directly use the $\text{Instruct}_{\text{ARD}}$ from the inference process to determine whether knowledge has increased and construct positive-negative pairs based on this judgment. In building Task-Oriented Generation Data, we use the same prompt as in the inference process and employ task evaluation metrics as supervision signals to select positive-negative pairs.

\section{Experimental Setup Details}
\label{Appendix C: Experimental Setup Details}

\subsection{More Implementation Details}
\label{Appendix C.1: More Implementation Details}
In detail, we reproduce Self-RAG and ReAct via the langchain framework\footnotemark[2].
During the inference stage, we use a temperature value of 0.1.
In the data construction phase, we primarily adjust two parameters: temperature and top\_p. By combining them pairwise, we use nine parameter sets to construct the training data, temperature $\in\left\{0.1,0.5,0.9\right \}$ and top\_p $\in\left\{ 0.1,0.5,0.9\right \}$.
\footnotetext[2]{\url{https://github.com/langchain-ai}}
Plus, We summarize all experimental settings in Table~\ref{tab-app:All Experimental Settings}.
\section{Details of Training Dataset Construction}
\label{Appendix:Details of Training Dataset Construction}
We randomly sampled 15000 samples from the train set of the 2WikiMQA dataset to construct our DNAlign dataset. We present the statistics of DNAlign in Table~\ref{tab:Statistics of DNAlign Datasets}.

\begin{table}[t!]
  \centering
    \renewcommand{\arraystretch}{0.9}
    \setlength{\tabcolsep}{2pt}
    \fontsize{9}{10}\selectfont
    \begin{tabular}{c|cccc|c}
    \toprule
          & $\mathcal{D}_{\text{Init}}$   & $\mathcal{D}_{\text{QR}}$    & $\mathcal{D}_{\text{KA}}$    & $\mathcal{D}_{\text{Ans}}$    & $\mathcal{D}$ \\
    \midrule
    \textbf{\# Sample} & 1900  & 1900  & 1900  & 300   & 6000 \\
    \bottomrule
    \end{tabular}%
    \caption{\bf Statistics of DNAlign Datasets for DPO.}
  \label{tab:Statistics of DNAlign Datasets}%
\end{table}%

\subsection{Note Initialization Data}
For each sampled instance, we used the original query $q_{0}$, the retrieved document $P_{k,0}$, and the prompt template $\text{Instruct}_{\text{Init}}$ to form the input $x_{\text{Init}}$, which was fed into the LLM for the note initialization inference process. To improve the diversity of responses, we configured nine parameter settings (detailed in Appendix~\ref{Appendix C.1: More Implementation Details}) during inference. It is worth mentioning that we also use multiple top-$k$ values to simulate diverse retrieval scenarios in real-world settings. After inference, we employed GPT-4o-mini as the evaluation model to select the positive example $y^{+}_{\text{Init}}$ and negative example $y^{-}_{\text{Init}}$ from the nine generated initial notes. We filtered out instances that lacked either a positive or a negative example.
Finally, the constructed training data for the note initialization process is denoted as $\left \{ x_{\text{Init}},y^{+}_{\text{Init}},y^{-}_{\text{Init}} \right \} \sim\mathcal{D}_{\text{Init}}$.

\subsection{Query Refinement Data}
We perform inference with the same parameter settings, top-$k$ strategy, and apply the same filtering approach. Notably, this stage requires using the generated output from the initialization note as input, meaning the quality of the initial note affects the quality of the training data at this stage.
Based on this, we construct the input $x_{\text{QR}}$ using $y^{+}_{\text{Init}}$, $q_{0}$, and the prompt template $\text{Instruct}_{\text{QR}}$. We then employ GPT-4o-mini to select positive examples $y^{+}_{\text{QR}}$ and negative examples $y^{-}_{\text{QR}}$, forming the dataset $\left \{ x_{\text{QR}},y^{+}_{\text{QR}},y^{-}_{\text{QR}} \right \} \sim\mathcal{D}_{\text{QR}}$.

\subsection{Knowledge Accumulation Data}
At this stage, the data enhances the model's ability to update notes and maximize knowledge accumulation. We maintain the same inference parameters, top-$k$ strategy, and filtering strategies.
We retrieve the top-$k$ documents, $P_{k,1}$, using the new query labeled $y^{+}_{\text{QR}}$.
Next, we use $y^{+}_{\text{Init}}$, $q_{0}$, and $P_{k,1}$ as the input.
We directly apply the evaluation strategy from the adaptive retrieval decision stage to generate positive and negative labels.
We then randomly select one positive and one negative example from the respective sets as the final positive and negative samples. The final dataset is denoted as $\left \{ x_{\text{KA}},y^{+}_{\text{KA}},y^{-}_{\text{KA}} \right \} \sim\mathcal{D}_{\text{KA}}$.

\subsection{Task-Oriented Generation Data}
After obtaining a high-quality note, we aim to align the system's response style for specific tasks. We employ the inference process of Vanilla RAG to generate answers and use the task evaluation metric to identify positive and negative examples. We apply the same parameters, top-$k$ strategy, and positive-negative pairs selection strategy as in the knowledge accumulation stage. The dataset can be formulated as: $\left \{ x_{\text{Ans}},y^{+}_{\text{Ans}},y^{-}_{\text{Ans}} \right \} \sim\mathcal{D}_{\text{Ans}}$.

\section{Case Study}
\label{Appendix D: Case Study}
In Tables~\ref{tab-app: DeepNote examples on the 2WikiMQA dataset} and~\ref{tab-app: DeepNote examples on the MusiQue dataset}, we present examples of DeepNote and conduct a case study.
Given the query "\textit{Where was the place of death of Anna Of Pomerania’s father?}", 
Vanilla RAG and Self-RAG failed to explore effective information and outputted the response "\text{No information}." DeepNote, after the second note update, identified the key information about her father. Following the third update, it not only located his place of death but also found the time of her father's death within the same paragraph, ultimately outputting the correct information: "\textit{Stettin}." Importantly, we observe that our answer not only includes the correct response but also expands on closely related knowledge: "\textit{Stettin (also known as Szczecin in Polish)}".
This demonstrates DeepNote's superior knowledge integration capability and the ability to maintain logical coherence during the integration process.

Additionally, Table~\ref{tab-app: Badcase Analysis of DeepNote on the HotpotQA dataset} presents a highly challenging question, i.e., "\textit{A man who played in the 1986 FIFA world cup played for what team during the 1982 Scottish League Cup Final?}".
This case illustrates that errors are mainly due to the inability to retrieve relevant information.

\clearpage

\begin{table*}[ht!]
  \centering
    \setlength{\tabcolsep}{1.5pt}
    \fontsize{7}{8}\selectfont
    \begin{tabular}{l|cc|cccc|cccc|cccc|ccc|c}
    \toprule
    \toprule
    \multirow{2}[2]{*}{\textbf{LLMs}} & \multirow{2}[2]{*}{\textbf{Max Step}} & \multirow{2}[2]{*}{\textbf{Max Failure}} & \multicolumn{4}{c|}{\textbf{HotpotQA}} & \multicolumn{4}{c|}{\textbf{2WikiMQA}} & \multicolumn{4}{c|}{\textbf{MusiQue}} & \multicolumn{3}{c|}{\textbf{ASQA}} & \textbf{StrategyQA} \\
\cmidrule{4-7}\cmidrule(lr){8-11}\cmidrule(lr){12-15}\cmidrule(lr){16-18}\cmidrule(lr){19-19} &       &       & acc.  & f1    & em    & avg.  & acc.  & f1    & em    & avg.  & acc.  & f1    & em    & avg.  & str-em & str-hit & avg.  & acc \\
    \midrule
    \multirow{6}[2]{*}{\textbf{Qwen2.5-7B-Instruct}} & 1     & 1     & 47.2  & 56.2  & 44.4  & 49.3  & 45.4  & 47.4  & 39.8  & 44.2  & 12.2  & 17.5  & 9.8   & 13.2  & 44.5  & 19.7  & 25.8  & 72.2  \\
          & 2     & 1     & 46.2  & 54.6  & 42.8  & 47.9  & 47.2  & 48.7  & 39.8  & 45.2  & 12.8  & 17.4  & 9.8   & 13.3  & 44.6  & 19.7  & 25.9  & 69.4  \\
          & 2     & 2     & 49.0  & 57.3  & 44.8  & 50.4  & 48.8  & 50.0  & 40.8  & 46.5  & 12.2  & 17.7  & 9.8   & 13.2  & 44.8  & 19.3  & 25.8  & 71.2  \\
          & 3     & 1     & 46.8  & 55.5  & 43.6  & 48.6  & 45.8  & 47.6  & 38.8  & 44.1  & 11.8  & 16.1  & 8.6   & 12.2  & 44.2  & 19.5  & 25.3  & 71.6  \\
          & 3     & 2     & 50.6  & 59.2  & 48.0  & 52.6  & 50.0  & 51.4  & 41.8  & 47.7  & 14.6  & 19.8  & 11.6  & 15.3  & 44.4  & 19.4  & 26.4  & 71.6  \\
          & 3     & 3     & 48.2  & 57.5  & 45.6  & 50.4  & 51.2  & 52.0  & 42.2  & 48.5  & 14.6  & 19.8  & 11.8  & 15.4  & 44.5  & 19.8  & 26.6  & 72.0  \\
    \midrule
    \multirow{6}[2]{*}{\textbf{Qwen2.5-7B-Instruct+DPO}} & 1     & 1     & 46.4  & 56.7  & 44.4  & 49.2  & 54.0  & 54.9  & 45.0  & 51.3  & 16.8  & 23.6  & 13.6  & 18.0  & 46.2  & 20.7  & 28.3  & 70.8  \\
          & 2     & 1     & 46.4  & 56.8  & 45.0  & 49.4  & 54.4  & 55.1  & 45.4  & 51.6  & 14.0  & 21.9  & 11.6  & 15.8  & 47.1  & 21.2  & 28.0  & 70.2  \\
          & 2     & 2     & 47.4  & 57.3  & 44.8  & 49.8  & 57.4  & 57.7  & 48.0  & 54.4  & 15.8  & 23.9  & 13.0  & 17.6  & 47.1  & 21.8  & 28.8  & 70.8  \\
          & 3     & 1     & 47.4  & 57.4  & 44.8  & 49.9  & 53.2  & 54.1  & 43.4  & 50.2  & 13.6  & 21.3  & 10.2  & 15.0  & 47.0  & 21.9  & 28.0  & 70.2  \\
          & 3     & 2     & 49.0  & 58.1  & 46.6  & 51.2  & 55.4  & 55.7  & 44.6  & 51.9  & 15.4  & 21.9  & 11.4  & 16.2  & 47.2  & 21.7  & 28.4  & 72.8  \\
          & 3     & 3     & 46.2  & 57.3  & 44.6  & 49.4  & 55.2  & 55.6  & 45.0  & 51.9  & 16.6  & 23.4  & 12.6  & 17.5  & 47.4  & 22.7  & 29.2  & 70.2  \\
    \midrule
    \multirow{6}[2]{*}{\textbf{Llama3.1-8B-Instruct}} & 1     & 1     & 45.2  & 52.0  & 39.8  & 45.7  & 54.2  & 53.8  & 45.6  & 51.2  & 14.4  & 18.9  & 11.0  & 14.8  & 43.8  & 18.3  & 25.6  & 72.2  \\
          & 2     & 1     & 45.8  & 52.8  & 40.8  & 46.5  & 53.4  & 52.9  & 46.0  & 50.8  & 14.8  & 18.9  & 11.8  & 15.2  & 45.0  & 18.9  & 26.4  & 72.0  \\
          & 2     & 2     & 49.8  & 56.9  & 44.6  & 50.4  & 53.8  & 53.6  & 45.0  & 50.8  & 16.0  & 21.6  & 12.6  & 16.7  & 44.4  & 18.4  & 26.5  & 72.8  \\
          & 3     & 1     & 47.8  & 54.2  & 42.6  & 48.2  & 54.6  & 53.0  & 45.0  & 50.9  & 15.0  & 19.2  & 11.4  & 15.2  & 44.8  & 19.2  & 26.4  & 73.0  \\
          & 3     & 2     & 48.0  & 54.3  & 41.2  & 47.8  & 58.0  & 58.2  & 48.2  & 54.8  & 17.0  & 21.3  & 13.2  & 17.2  & 43.4  & 17.9  & 26.2  & 70.8  \\
          & 3     & 3     & 49.6  & 56.6  & 44.8  & 50.3  & 57.2  & 56.3  & 48.0  & 53.8  & 16.2  & 21.4  & 12.2  & 16.6  & 44.6  & 18.9  & 26.7  & 70.2  \\
    \midrule
    \multirow{6}[2]{*}{\textbf{Llama3.1-8B-Instruct+DPO}} & 1     & 1     & 53.2  & 60.1  & 44.8  & 52.7  & 60.2  & 57.3  & 47.6  & 55.0  & 21.6  & 24.9  & 13.2  & 19.9  & 46.7  & 20.8  & 29.1  & 74.2  \\
          & 2     & 1     & 54.2  & 58.1  & 41.2  & 51.2  & 61.8  & 60.0  & 49.6  & 57.1  & 21.8  & 26.4  & 15.2  & 21.1  & 46.3  & 20.4  & 29.3  & 73.8  \\
          & 2     & 2     & 53.6  & 58.1  & 42.6  & 51.4  & 63.6  & 59.9  & 48.4  & 57.3  & 21.4  & 26.9  & 15.2  & 21.2  & 46.6  & 20.6  & 29.5  & 72.4  \\
          & 3     & 1     & 54.0  & 58.5  & 42.8  & 51.8  & 64.8  & 61.9  & 50.0  & 58.9  & 25.4  & 27.7  & 15.8  & 23.0  & 46.5  & 19.2  & 29.6  & 73.2  \\
          & 3     & 2     & 54.6  & 58.9  & 44.0  & 52.5  & 63.8  & 60.5  & 47.4  & 57.2  & 24.4  & 27.3  & 14.4  & 22.0  & 46.4  & 19.8  & 29.4  & 74.2  \\
          & 3     & 3     & 55.6  & 59.4  & 43.0  & 52.7  & 65.6  & 62.3  & 50.0  & 59.3  & 22.4  & 26.5  & 14.4  & 21.1  & 47.1  & 20.2  & 29.5  & 72.2  \\
    \midrule
    \multirow{6}[2]{*}{\textbf{GPT-4o-mini}} & 1     & 1     & 56.2  & 63.2  & 49.8  & 56.4  & 60.6  & 59.8  & 50.0  & 56.8  & 22.0  & 28.3  & 16.2  & 22.2  & 48.4  & 22.9  & 31.2  & 75.4  \\
          & 2     & 1     & 57.0  & 64.0  & 49.2  & 56.7  & 64.0  & 62.6  & 52.4  & 59.7  & 22.4  & 28.1  & 16.2  & 22.2  & 48.7  & 22.7  & 31.2  & 75.4  \\
          & 2     & 2     & 58.0  & 64.9  & 50.0  & 57.6  & 65.8  & 64.3  & 53.0  & 61.0  & 23.4  & 29.6  & 17.2  & 23.4  & 48.7  & 22.4  & 31.5  & 77.4  \\
          & 3     & 1     & 57.0  & 63.4  & 49.0  & 56.5  & 63.4  & 61.6  & 51.6  & 58.9  & 22.8  & 28.8  & 16.4  & 22.7  & 48.4  & 21.8  & 31.0  & 76.2  \\
          & 3     & 2     & 56.8  & 64.3  & 50.2  & 57.1  & 66.2  & 63.7  & 52.6  & 60.8  & 24.8  & 31.3  & 18.4  & 24.8  & 48.6  & 23.1  & 32.2  & 76.4  \\
          & 3     & 3     & 58.4  & 65.4  & 49.8  & 57.9  & 64.0  & 62.3  & 51.2  & 59.2  & 25.6  & 31.0  & 19.4  & 25.3  & 49.4  & 23.1  & 32.6  & 77.0  \\
    \midrule
    \multirow{6}[2]{*}{\textbf{Llama3.1-70B-Instruct}} & 1     & 1     & 55.6  & 63.7  & 50.6  & 56.6  & 65.8  & 61.5  & 52.4  & 59.9  & 27.2  & 30.4  & 19.8  & 25.8  & 43.8  & 15.9  & 28.5  & 74.8  \\
          & 2     & 1     & 56.8  & 64.9  & 51.6  & 57.8  & 69.2  & 64.6  & 55.2  & 63.0  & 28.8  & 31.7  & 21.4  & 27.3  & 44.5  & 16.7  & 29.5  & 75.4  \\
          & 2     & 2     & 60.2  & 68.4  & 54.4  & 61.0  & 70.6  & 66.1  & 56.2  & 64.3  & 33.0  & 34.9  & 24.4  & 30.8  & 45.1  & 17.9  & 31.3  & 75.0  \\
          & 3     & 1     & 57.8  & 65.4  & 52.0  & 58.4  & 70.0  & 65.0  & 56.0  & 63.7  & 28.6  & 31.7  & 21.4  & 27.2  & 44.7  & 17.6  & 29.8  & 75.2  \\
          & 3     & 2     & 59.2  & 67.2  & 54.2  & 60.2  & 72.4  & 67.1  & 55.8  & 65.1  & 32.6  & 35.0  & 23.0  & 30.2  & 44.2  & 16.6  & 30.3  & 75.4  \\
          & 3     & 3     & 59.6  & 67.8  & 53.4  & 60.3  & 73.0  & 67.9  & 57.2  & 66.0  & 32.0  & 35.8  & 23.6  & 30.5  & 45.3  & 17.8  & 31.2  & 77.8  \\
    \bottomrule
    \bottomrule
    \end{tabular}%
    \caption{{\bf Results (\%) of performance on different adaptive hyper-parameter analysis} of DeepNote on all LLMs and datasets. We have set a total of six sets of hyper-parameters.}
  \label{tab-app:Adaptive Parameter Analysis of DeepNote}%
\end{table*}%
\clearpage

\begin{table*}[ht!]
  \centering
    \setlength{\tabcolsep}{2pt}
    \fontsize{8}{9}\selectfont
    \begin{tabular}{l|cccc|cccc|cccc|ccc|c}
    \toprule
    \toprule
    \multicolumn{1}{c}{\multirow{2}[4]{*}{\textbf{Methods}}} & \multicolumn{4}{|c|}{\textbf{HotpotQA}} & \multicolumn{4}{c|}{\textbf{2WikiMQA}} & \multicolumn{4}{c|}{\textbf{MusiQue}} & \multicolumn{3}{c|}{\textbf{ASQA}} & \textbf{StrategyQA } \\
\cmidrule(lr){2-5}\cmidrule(lr){6-9} \cmidrule(lr){10-13}\cmidrule(lr){14-16}  \cmidrule(lr){17-17}          & acc.  & f1    & em    & avg.  & acc.  & f1    & em    & avg.  & acc.  & f1    & em    & avg.  & str-em & str-hit & avg.  & acc. \\
    \midrule
    \multicolumn{17}{c}{\textit{Qwen2.5-7B-instruct}} \\
    DeepNote & \cellcolor[rgb]{ .882,  .984,  .992}50.6  & \cellcolor[rgb]{ .882,  .984,  .992}59.2  & \cellcolor[rgb]{ .882,  .984,  .992}48.0  & \cellcolor[rgb]{ .882,  .984,  .992}52.6  & \cellcolor[rgb]{ .882,  .984,  .992}50.0  & \cellcolor[rgb]{ .882,  .984,  .992}51.4  & \cellcolor[rgb]{ .882,  .984,  .992}41.8  & \cellcolor[rgb]{ .882,  .984,  .992}47.7  & \cellcolor[rgb]{ .882,  .984,  .992}14.6  & \cellcolor[rgb]{ .882,  .984,  .992}19.8  & \cellcolor[rgb]{ .882,  .984,  .992}11.6  & \cellcolor[rgb]{ .882,  .984,  .992}15.3  & \cellcolor[rgb]{ .882,  .984,  .992}44.4  & \cellcolor[rgb]{ .882,  .984,  .992}19.4  & \cellcolor[rgb]{ .882,  .984,  .992}26.4  & \cellcolor[rgb]{ .882,  .984,  .992}71.6  \\
    w/o Adap. Retrieval & \cellcolor[rgb]{ .973,  .953,  .984}40.2  & \cellcolor[rgb]{ .973,  .953,  .984}48.3  & \cellcolor[rgb]{ .973,  .953,  .984}37.4  & \cellcolor[rgb]{ .973,  .953,  .984}42.0  & \cellcolor[rgb]{ .973,  .953,  .984}35.8  & \cellcolor[rgb]{ .973,  .953,  .984}39.6  & \cellcolor[rgb]{ .973,  .953,  .984}34.6  & \cellcolor[rgb]{ .973,  .953,  .984}36.7  & \cellcolor[rgb]{ .973,  .953,  .984}8.6  & \cellcolor[rgb]{ .973,  .953,  .984}12.7  & \cellcolor[rgb]{ .973,  .953,  .984}6.2  & \cellcolor[rgb]{ .973,  .953,  .984}9.2  & \cellcolor[rgb]{ .973,  .953,  .984}43.9  & \cellcolor[rgb]{ .973,  .953,  .984}19.0  & \cellcolor[rgb]{ .973,  .953,  .984}24.0  & \cellcolor[rgb]{ .973,  .953,  .984}71.2  \\
    w/o Adap. Retrieval \& Init. Note & \cellcolor[rgb]{ .953,  .886,  1}37.4  & \cellcolor[rgb]{ .953,  .886,  1}44.0  & \cellcolor[rgb]{ .953,  .886,  1}33.6  & \cellcolor[rgb]{ .953,  .886,  1}38.3  & \cellcolor[rgb]{ .953,  .886,  1}33.2  & \cellcolor[rgb]{ .953,  .886,  1}36.3  & \cellcolor[rgb]{ .953,  .886,  1}31.8  & \cellcolor[rgb]{ .953,  .886,  1}33.8  & \cellcolor[rgb]{ .953,  .886,  1}7.6  & \cellcolor[rgb]{ .953,  .886,  1}12.5  & \cellcolor[rgb]{ .953,  .886,  1}5.6  & \cellcolor[rgb]{ .953,  .886,  1}8.6  & \cellcolor[rgb]{ .953,  .886,  1}42.1  & \cellcolor[rgb]{ .953,  .886,  1}15.9  & \cellcolor[rgb]{ .953,  .886,  1}22.2  & \cellcolor[rgb]{ .953,  .886,  1}68.4  \\
    \midrule
    \multicolumn{17}{c}{\textit{Llama3.1-8B-Instruct}} \\
    DeepNote & \cellcolor[rgb]{ .882,  .984,  .992}48.0  & \cellcolor[rgb]{ .882,  .984,  .992}54.3  & \cellcolor[rgb]{ .882,  .984,  .992}41.2  & \cellcolor[rgb]{ .882,  .984,  .992}47.8  & \cellcolor[rgb]{ .882,  .984,  .992}58.0  & \cellcolor[rgb]{ .882,  .984,  .992}58.2  & \cellcolor[rgb]{ .882,  .984,  .992}48.2  & \cellcolor[rgb]{ .882,  .984,  .992}54.8  & \cellcolor[rgb]{ .882,  .984,  .992}17.0  & \cellcolor[rgb]{ .882,  .984,  .992}21.3  & \cellcolor[rgb]{ .882,  .984,  .992}13.2  & \cellcolor[rgb]{ .882,  .984,  .992}17.2  & \cellcolor[rgb]{ .882,  .984,  .992}43.4  & \cellcolor[rgb]{ .882,  .984,  .992}17.9  & \cellcolor[rgb]{ .882,  .984,  .992}26.2  & \cellcolor[rgb]{ .953,  .886,  1}70.8  \\
    w/o Adap. Retrieval & \cellcolor[rgb]{ .973,  .953,  .984}37.6  & \cellcolor[rgb]{ .953,  .886,  1}44.5  & \cellcolor[rgb]{ .953,  .886,  1}33.6  & \cellcolor[rgb]{ .953,  .886,  1}38.6  & \cellcolor[rgb]{ .973,  .953,  .984}39.6  & \cellcolor[rgb]{ .973,  .953,  .984}41.2  & \cellcolor[rgb]{ .973,  .953,  .984}38.0  & \cellcolor[rgb]{ .973,  .953,  .984}39.6  & \cellcolor[rgb]{ .973,  .953,  .984}8.4  & \cellcolor[rgb]{ .953,  .886,  1}11.9  & \cellcolor[rgb]{ .953,  .886,  1}5.8  & \cellcolor[rgb]{ .973,  .953,  .984}8.7  & \cellcolor[rgb]{ .973,  .953,  .984}41.3  & \cellcolor[rgb]{ .973,  .953,  .984}16.6  & \cellcolor[rgb]{ .973,  .953,  .984}22.2  & \cellcolor[rgb]{ .882,  .984,  .992}72.2  \\
    w/o Adap. Retrieval \& Init. Note & \cellcolor[rgb]{ .973,  .953,  .984}37.6  & \cellcolor[rgb]{ .973,  .953,  .984}46.4  & \cellcolor[rgb]{ .973,  .953,  .984}35.0  & \cellcolor[rgb]{ .973,  .953,  .984}39.7  & \cellcolor[rgb]{ .953,  .886,  1}33.4  & \cellcolor[rgb]{ .953,  .886,  1}36.3  & \cellcolor[rgb]{ .953,  .886,  1}32.0  & \cellcolor[rgb]{ .953,  .886,  1}33.9  & \cellcolor[rgb]{ .953,  .886,  1}6.8  & \cellcolor[rgb]{ .973,  .953,  .984}12.1  & \cellcolor[rgb]{ .973,  .953,  .984}6.0  & \cellcolor[rgb]{ .953,  .886,  1}8.3  & \cellcolor[rgb]{ .953,  .886,  1}39.3  & \cellcolor[rgb]{ .953,  .886,  1}13.3  & \cellcolor[rgb]{ .953,  .886,  1}20.3  & \cellcolor[rgb]{ .973,  .953,  .984}71.4  \\
    \midrule
    \multicolumn{17}{c}{\textit{GPT-4o-mini}} \\
    DeepNote & \cellcolor[rgb]{ .882,  .984,  .992}56.8  & \cellcolor[rgb]{ .882,  .984,  .992}64.3  & \cellcolor[rgb]{ .882,  .984,  .992}50.2  & \cellcolor[rgb]{ .882,  .984,  .992}57.1  & \cellcolor[rgb]{ .882,  .984,  .992}66.2  & \cellcolor[rgb]{ .882,  .984,  .992}63.7  & \cellcolor[rgb]{ .882,  .984,  .992}52.6  & \cellcolor[rgb]{ .882,  .984,  .992}60.8  & \cellcolor[rgb]{ .882,  .984,  .992}24.8  & \cellcolor[rgb]{ .882,  .984,  .992}31.3  & \cellcolor[rgb]{ .882,  .984,  .992}18.4  & \cellcolor[rgb]{ .882,  .984,  .992}24.8  & \cellcolor[rgb]{ .882,  .984,  .992}48.6  & \cellcolor[rgb]{ .882,  .984,  .992}23.1  & \cellcolor[rgb]{ .882,  .984,  .992}32.2  & \cellcolor[rgb]{ .882,  .984,  .992}76.4  \\
    w/o Adap. Retrieval & \cellcolor[rgb]{ .973,  .953,  .984}47.0  & \cellcolor[rgb]{ .973,  .953,  .984}54.6  & \cellcolor[rgb]{ .973,  .953,  .984}41.4  & \cellcolor[rgb]{ .973,  .953,  .984}47.7  & \cellcolor[rgb]{ .973,  .953,  .984}46.2  & \cellcolor[rgb]{ .973,  .953,  .984}48.8  & \cellcolor[rgb]{ .973,  .953,  .984}43.4  & \cellcolor[rgb]{ .973,  .953,  .984}46.1  & \cellcolor[rgb]{ .973,  .953,  .984}14.2  & \cellcolor[rgb]{ .973,  .953,  .984}20.8  & \cellcolor[rgb]{ .973,  .953,  .984}10.8  & \cellcolor[rgb]{ .973,  .953,  .984}15.3  & \cellcolor[rgb]{ .973,  .953,  .984}47.1  & \cellcolor[rgb]{ .973,  .953,  .984}21.0  & \cellcolor[rgb]{ .973,  .953,  .984}27.8  & \cellcolor[rgb]{ .973,  .953,  .984}74.8  \\
    w/o Adap. Retrieval \& Init. Note & \cellcolor[rgb]{ .953,  .886,  1}44.0  & \cellcolor[rgb]{ .953,  .886,  1}52.2  & \cellcolor[rgb]{ .953,  .886,  1}40.0  & \cellcolor[rgb]{ .953,  .886,  1}45.4  & \cellcolor[rgb]{ .953,  .886,  1}40.4  & \cellcolor[rgb]{ .953,  .886,  1}44.4  & \cellcolor[rgb]{ .953,  .886,  1}39.2  & \cellcolor[rgb]{ .953,  .886,  1}41.3  & \cellcolor[rgb]{ .953,  .886,  1}10.6  & \cellcolor[rgb]{ .953,  .886,  1}17.3  & \cellcolor[rgb]{ .953,  .886,  1}7.6  & \cellcolor[rgb]{ .953,  .886,  1}11.8  & \cellcolor[rgb]{ .953,  .886,  1}44.3  & \cellcolor[rgb]{ .953,  .886,  1}17.5  & \cellcolor[rgb]{ .953,  .886,  1}24.5  & \cellcolor[rgb]{ .953,  .886,  1}71.2  \\
    \midrule
    \multicolumn{17}{c}{\textit{Llama3.1-70B-Instruct}} \\
    DeepNote & \cellcolor[rgb]{ .882,  .984,  .992}59.2  & \cellcolor[rgb]{ .882,  .984,  .992}67.2  & \cellcolor[rgb]{ .882,  .984,  .992}54.2  & \cellcolor[rgb]{ .882,  .984,  .992}60.2  & \cellcolor[rgb]{ .882,  .984,  .992}72.4  & \cellcolor[rgb]{ .882,  .984,  .992}67.1  & \cellcolor[rgb]{ .882,  .984,  .992}55.8  & \cellcolor[rgb]{ .882,  .984,  .992}65.1  & \cellcolor[rgb]{ .882,  .984,  .992}32.6  & \cellcolor[rgb]{ .882,  .984,  .992}35.0  & \cellcolor[rgb]{ .882,  .984,  .992}23.0  & \cellcolor[rgb]{ .882,  .984,  .992}30.2  & \cellcolor[rgb]{ .882,  .984,  .992}44.2  & \cellcolor[rgb]{ .882,  .984,  .992}16.6  & \cellcolor[rgb]{ .882,  .984,  .992}30.3  & \cellcolor[rgb]{ .882,  .984,  .992}75.4  \\
    w/o Adap. Retrieval & \cellcolor[rgb]{ .953,  .886,  1}42.6  & \cellcolor[rgb]{ .953,  .886,  1}51.0  & \cellcolor[rgb]{ .953,  .886,  1}39.8  & \cellcolor[rgb]{ .953,  .886,  1}44.5  & \cellcolor[rgb]{ .953,  .886,  1}38.8  & \cellcolor[rgb]{ .953,  .886,  1}40.0  & \cellcolor[rgb]{ .953,  .886,  1}36.6  & \cellcolor[rgb]{ .953,  .886,  1}38.5  & \cellcolor[rgb]{ .973,  .953,  .984}12.6  & \cellcolor[rgb]{ .953,  .886,  1}16.3  & \cellcolor[rgb]{ .973,  .953,  .984}10.4  & \cellcolor[rgb]{ .973,  .953,  .984}13.1  & \cellcolor[rgb]{ .973,  .953,  .984}42.4  & \cellcolor[rgb]{ .973,  .953,  .984}15.5  & \cellcolor[rgb]{ .973,  .953,  .984}23.7  & \cellcolor[rgb]{ .973,  .953,  .984}73.8  \\
    w/o Adap. Retrieval \& Init. Note & \cellcolor[rgb]{ .973,  .953,  .984}44.6  & \cellcolor[rgb]{ .973,  .953,  .984}53.6  & \cellcolor[rgb]{ .973,  .953,  .984}42.2  & \cellcolor[rgb]{ .973,  .953,  .984}46.8  & \cellcolor[rgb]{ .973,  .953,  .984}45.2  & \cellcolor[rgb]{ .973,  .953,  .984}47.0  & \cellcolor[rgb]{ .973,  .953,  .984}42.8  & \cellcolor[rgb]{ .973,  .953,  .984}45.0  & \cellcolor[rgb]{ .953,  .886,  1}11.6  & \cellcolor[rgb]{ .973,  .953,  .984}17.5  & \cellcolor[rgb]{ .953,  .886,  1}9.2  & \cellcolor[rgb]{ .953,  .886,  1}12.8  & \cellcolor[rgb]{ .953,  .886,  1}42.0  & \cellcolor[rgb]{ .953,  .886,  1}15.3  & \cellcolor[rgb]{ .953,  .886,  1}23.4  & \cellcolor[rgb]{ .973,  .953,  .984}73.8  \\
    \bottomrule
    \bottomrule
    \end{tabular}%
    \caption{\textbf{All results (\%) of ablation study.}
    \colorbox[rgb]{ .882,  .984,  .992}{"Blue"}, \colorbox[rgb]{ .973,  .953,  .984}{"light purple"} and \colorbox[rgb]{ .953,  .89,  1}{"dark purple"} represent the highest, second highest, and lowest values among the results of different top-$k$, respectively.
    }
  \label{tab-app:All results of Ablation Study}%
\end{table*}%

\begin{table*}[ht!]
  \centering
    \setlength{\tabcolsep}{2pt}
    \fontsize{8}{9}\selectfont
    \begin{tabular}{c|c|cccc|cccc|cccc|ccc|c}
    \toprule
    \toprule
    \multirow{2}[2]{*}{\textbf{Top-$\mathbf{k}$}} & \multirow{2}[2]{*}{\textbf{Methods}} & \multicolumn{4}{c|}{\textbf{HotpotQA}} & \multicolumn{4}{c|}{\textbf{2WikiMQA}} & \multicolumn{4}{c|}{\textbf{MusiQue}} & \multicolumn{3}{c|}{\textbf{ASQA}} & \textbf{StrategyQA} \\
\cmidrule{3-18}          &       & acc.  & f1    & em    & avg.  & acc.  & f1    & em    & avg.  & acc.  & f1    & em    & avg.  & str-em & str-hit & avg.  & acc. \\
    \midrule
    \multirow{2}[2]{*}{Top-3} & Vanilla RAG & 42.8  & 51.0  & 39.0  & 44.3  & 39.2  & 43.0  & 38.2  & 40.1  & 10.2  & 15.9  & 7.2   & 11.1  & 41.8  & 16.5  & 23.1  & 68.8  \\
          & DeepNote & \cellcolor[rgb]{ .953,  .886,  1}\textbf{56.6} & \cellcolor[rgb]{ .953,  .89,  1}\textbf{64.1} & \cellcolor[rgb]{ .953,  .89,  1}\textbf{49.6} & \cellcolor[rgb]{ .953,  .89,  1}\textbf{56.8} & \cellcolor[rgb]{ .953,  .89,  1}\textbf{61.4} & \cellcolor[rgb]{ .953,  .89,  1}\textbf{61.0} & \cellcolor[rgb]{ .973,  .953,  .984}\textbf{51.0} & \cellcolor[rgb]{ .953,  .89,  1}\textbf{57.8} & \cellcolor[rgb]{ .953,  .89,  1}\textbf{21.6} & \cellcolor[rgb]{ .953,  .89,  1}\textbf{27.9} & \cellcolor[rgb]{ .953,  .89,  1}\textbf{16.2} & \cellcolor[rgb]{ .953,  .89,  1}\textbf{21.9} & \cellcolor[rgb]{ .953,  .89,  1}\textbf{48.1} & \cellcolor[rgb]{ .953,  .89,  1}\textbf{21.7} & \cellcolor[rgb]{ .953,  .89,  1}\textbf{30.6} & \cellcolor[rgb]{ .953,  .89,  1}\textbf{73.2} \\
    \midrule
    \multirow{2}[2]{*}{Top-5} & Vanilla RAG & 44.0  & 52.2  & 40.0  & 45.4  & 40.4  & 44.4  & 39.2  & 41.3  & 10.6  & 17.3  & 7.6   & 11.8  & 44.3  & 17.5  & 24.5  & 71.2  \\
          & DeepNote & \cellcolor[rgb]{ .973,  .953,  .984}\textbf{56.8} & \cellcolor[rgb]{ .973,  .953,  .984}\textbf{64.3} & \cellcolor[rgb]{ .973,  .953,  .984}\textbf{50.2} & \cellcolor[rgb]{ .973,  .953,  .984}\textbf{57.1} & \cellcolor[rgb]{ .882,  .984,  .992}\textbf{66.2} & \cellcolor[rgb]{ .882,  .984,  .992}\textbf{63.7} & \cellcolor[rgb]{ .882,  .984,  .992}\textbf{52.6} & \cellcolor[rgb]{ .882,  .984,  .992}\textbf{60.8} & \cellcolor[rgb]{ .882,  .984,  .992}\textbf{24.8} & \cellcolor[rgb]{ .882,  .984,  .992}\textbf{31.3} & \cellcolor[rgb]{ .973,  .953,  .984}\textbf{18.4} & \cellcolor[rgb]{ .882,  .984,  .992}\textbf{24.8} & \cellcolor[rgb]{ .973,  .953,  .984}\textbf{48.6} & \cellcolor[rgb]{ .882,  .984,  .992}\textbf{23.1} & \cellcolor[rgb]{ .973,  .953,  .984}\textbf{32.2} & \cellcolor[rgb]{ .882,  .984,  .992}\textbf{76.4} \\
    \midrule
    \multirow{2}[2]{*}{Top-7} & Vanilla RAG & 45.2  & 54.1  & 41.8  & 47.0  & 39.4  & 43.4  & 38.2  & 40.3  & 10.8  & 16.3  & 7.2   & 11.4  & 45.4  & 18.1  & 25.0  & 69.8  \\
          & DeepNote & \cellcolor[rgb]{ .882,  .984,  .992}\textbf{59.0} & \cellcolor[rgb]{ .882,  .984,  .992}\textbf{66.0} & \cellcolor[rgb]{ .882,  .984,  .992}\textbf{51.4} & \cellcolor[rgb]{ .882,  .984,  .992}\textbf{58.8} & \cellcolor[rgb]{ .973,  .953,  .984}\textbf{65.0} & \cellcolor[rgb]{ .973,  .953,  .984}\textbf{62.8} & \cellcolor[rgb]{ .953,  .89,  1}\textbf{50.4 } & \cellcolor[rgb]{ .973,  .953,  .984}\textbf{59.4} & \cellcolor[rgb]{ .882,  .984,  .992}\textbf{24.8} & \cellcolor[rgb]{ .973,  .953,  .984}\textbf{30.4} & \cellcolor[rgb]{ .882,  .984,  .992}\textbf{19.2} & \cellcolor[rgb]{ .882,  .984,  .992}\textbf{24.8} & \cellcolor[rgb]{ .882,  .984,  .992}\textbf{50.0} & \cellcolor[rgb]{ .973,  .953,  .984}\textbf{22.4} & \cellcolor[rgb]{ .882,  .984,  .992}\textbf{32.4} & \cellcolor[rgb]{ .973,  .953,  .984}\textbf{74.2} \\
    \bottomrule
    \bottomrule
    \end{tabular}%
    \caption{\textbf{Results (\%) on different Top-$\mathbf{k}$.} We present the results of DeepNote using GPT-4o-mini as the backbone model.
    \colorbox[rgb]{ .882,  .984,  .992}{"Blue"}, \colorbox[rgb]{ .973,  .953,  .984}{"light purple"} and \colorbox[rgb]{ .953,  .89,  1}{"dark purple"} represent the highest, second highest, and lowest values among the results of different top-$k$, respectively. "\textbf{Bold}" means the higher value between Vanilla RAG and DeepNote under the same top-$k$ setting.
    }
  \label{tab-app:Top-k. GPT-4o-mini}%
\end{table*}%

\begin{table*}[ht!]
  \centering
    \setlength{\tabcolsep}{2pt}
    \fontsize{8}{9}\selectfont
    \begin{tabular}{c|c|cccc|cccc|cccc|ccc|c}
    \toprule
    \toprule
    \multirow{2}[2]{*}{\textbf{Top-$\mathbf{k}$}} & \multirow{2}[2]{*}{\textbf{Methods}} & \multicolumn{4}{c|}{\textbf{HotpotQA}} & \multicolumn{4}{c|}{\textbf{2WikiMQA}} & \multicolumn{4}{c|}{\textbf{MusiQue}} & \multicolumn{3}{c|}{\textbf{ASQA}} & \textbf{StrategyQA} \\
\cmidrule{3-18}          &       & acc.  & f1    & em    & avg.  & acc.  & f1    & em    & avg.  & acc.  & f1    & em    & avg.  & str-em & str-hit & avg.  & acc. \\
    \midrule
    \multirow{2}[2]{*}{Top-3} & Vanilla RAG & 42.2  & 50.8  & 40.8  & 44.6  & 42.8  & 44.3  & 40.0  & 42.4  & 11.4  & 17.4  & 8.8   & 12.5  & 40.1  & 13.4  & 22.0  & 73.2  \\
          & DeepNote & \cellcolor[rgb]{ .953,  .89,  1}\textbf{56.2} & \cellcolor[rgb]{ .953,  .89,  1}\textbf{64.1} & \cellcolor[rgb]{ .953,  .89,  1}\textbf{51.4} & \cellcolor[rgb]{ .953,  .89,  1}\textbf{57.2} & \cellcolor[rgb]{ .953,  .89,  1}\textbf{65.4} & \cellcolor[rgb]{ .953,  .89,  1}\textbf{61.6} & \cellcolor[rgb]{ .953,  .89,  1}\textbf{53.0} & \cellcolor[rgb]{ .953,  .89,  1}\textbf{60.0} & \cellcolor[rgb]{ .953,  .89,  1}\textbf{30.2} & \cellcolor[rgb]{ .953,  .89,  1}\textbf{32.7} & \cellcolor[rgb]{ .953,  .89,  1}\textbf{22.4} & \cellcolor[rgb]{ .953,  .89,  1}\textbf{28.4} & \cellcolor[rgb]{ .953,  .89,  1}\textbf{43.4} & \textbf{16.7} & \cellcolor[rgb]{ .953,  .89,  1}\textbf{29.5} & \cellcolor[rgb]{ .953,  .89,  1}\textbf{73.8} \\
    \midrule
    \multirow{2}[2]{*}{Top-5} & Vanilla RAG & 44.6  & 53.6  & 42.2  & 46.8  & 45.2  & 47.0  & 42.8  & 45.0  & 11.6  & 17.5  & 9.2   & 12.8  & 42.0  & 15.3  & 23.4  & 73.8  \\
          & DeepNote & \cellcolor[rgb]{ .973,  .953,  .984}\textbf{59.2} & \cellcolor[rgb]{ .973,  .953,  .984}\textbf{67.2} & \cellcolor[rgb]{ .973,  .953,  .984}\textbf{54.2} & \cellcolor[rgb]{ .973,  .953,  .984}\textbf{60.2} & \cellcolor[rgb]{ .973,  .953,  .984}\textbf{72.4} & \cellcolor[rgb]{ .973,  .953,  .984}\textbf{67.1} & \cellcolor[rgb]{ .973,  .953,  .984}\textbf{55.8} & \cellcolor[rgb]{ .973,  .953,  .984}\textbf{65.1} & \cellcolor[rgb]{ .886,  .984,  .992}\textbf{32.6} & \cellcolor[rgb]{ .886,  .984,  .992}\textbf{35.0} & \cellcolor[rgb]{ .973,  .953,  .984}\textbf{23.0} & \cellcolor[rgb]{ .886,  .984,  .992}\textbf{30.2} & \cellcolor[rgb]{ .973,  .953,  .984}\textbf{44.2} & \cellcolor[rgb]{ .953,  .89,  1}\textbf{16.6 } & \cellcolor[rgb]{ .973,  .953,  .984}\textbf{30.3} & \cellcolor[rgb]{ .886,  .984,  .992}\textbf{75.4} \\
    \midrule
    \multirow{2}[2]{*}{Top-7} & Vanilla RAG & 45.4  & 54.7  & 43.4  & 47.8  & 45.8  & 47.9  & 44.0  & 45.9  & 10.4  & 17.5  & 9.0   & 12.3  & 43.6  & 15.8  & 23.9  & 76.4  \\
          & DeepNote & \cellcolor[rgb]{ .886,  .984,  .992}\textbf{59.8} & \cellcolor[rgb]{ .886,  .984,  .992}\textbf{67.5} & \cellcolor[rgb]{ .886,  .984,  .992}\textbf{55.0} & \cellcolor[rgb]{ .886,  .984,  .992}\textbf{60.8} & \cellcolor[rgb]{ .886,  .984,  .992}\textbf{75.4} & \cellcolor[rgb]{ .886,  .984,  .992}\textbf{69.9} & \cellcolor[rgb]{ .886,  .984,  .992}\textbf{58.8} & \cellcolor[rgb]{ .886,  .984,  .992}\textbf{68.0} & \cellcolor[rgb]{ .973,  .953,  .984}\textbf{31.8} & \cellcolor[rgb]{ .973,  .953,  .984}\textbf{34.6} & \cellcolor[rgb]{ .886,  .984,  .992}\textbf{23.2} & \cellcolor[rgb]{ .973,  .953,  .984}\textbf{29.9} & \cellcolor[rgb]{ .886,  .984,  .992}\textbf{46.4} & \cellcolor[rgb]{ .886,  .984,  .992}\textbf{18.0} & \cellcolor[rgb]{ .886,  .984,  .992}\textbf{31.4} & \cellcolor[rgb]{ .973,  .953,  .984}\textbf{74.4} \\
    \bottomrule
    \bottomrule
    \end{tabular}%
 \caption{\textbf{Results (\%) on different Top-$\mathbf{k}$.} We present the results of DeepNote using Llama3.1-70B-Instruct as the backbone model.
    \colorbox[rgb]{ .882,  .984,  .992}{"Blue"}, \colorbox[rgb]{ .973,  .953,  .984}{"light purple"} and \colorbox[rgb]{ .953,  .89,  1}{"dark purple"} represent the highest, second highest, and lowest values among the results of different top-$k$, respectively.
    "\textbf{Bold}" means the higher value between Vanilla RAG and DeepNote under the same top-$k$ setting.
    }
  \label{tab-app:Top-k. Llama3.1-70B-Instruct}%
\end{table*}%

\begin{table*}[ht!]
  \centering
    \setlength{\tabcolsep}{2pt}
    \fontsize{8}{9}\selectfont
    \begin{tabular}{c|c|cccc|cccc|cccc}
    \toprule
    \toprule
    \multirow{2}[2]{*}{\textbf{Top-$\mathbf{k}$}} & \multirow{2}[2]{*}{\textbf{Retrievers}} & \multicolumn{4}{c|}{\bf HotpotQA} & \multicolumn{4}{c|}{\bf 2WikiMQA} & \multicolumn{4}{c}{\bf MusiQue} \\
\cmidrule(lr){3-6}\cmidrule(lr){7-10}   \cmidrule(lr){11-14}             &       & acc.  & f1    & em    & avg.  & acc.  & f1    & em    & avg.  & acc.  & f1    & em    & avg. \\
    \midrule
    \multirow{2}[2]{*}{Top-5} & BM25  & \multicolumn{1}{c}{59.2} & \multicolumn{1}{c}{67.2} & \multicolumn{1}{c}{54.2} & 60.2  & \multicolumn{1}{c}{72.4} & \multicolumn{1}{c}{67.1} & \multicolumn{1}{c}{55.8} & 65.1  & \multicolumn{1}{c}{32.6} & \multicolumn{1}{c}{35.0} & \multicolumn{1}{c}{23.0} & 30.2  \\
\cmidrule(lr){2-2}\cmidrule(lr){3-6}\cmidrule(lr){7-10}   \cmidrule(lr){11-14} 
     & bge-base-en-v1.5 & \textbf{61.6} & \textbf{68.4} & \textbf{55.0} & \textbf{61.7} & \textbf{75.8} & \textbf{69.7} & \textbf{59.6} & \textbf{68.4} & \textbf{33.4} & \textbf{37.1} & \textbf{26.0} & \textbf{32.2} \\
    \bottomrule
    \bottomrule
    \end{tabular}%
 \caption{\textbf{Results (\%) of different retrievers.} We present the results of DeepNote on Llama3.1-70B-Instruct.}
  \label{tab-app: Impacts of Different retrievers. Llama3.1-70B-Instruct}%
\end{table*}%

\clearpage

\clearpage

\begin{table*}[ht!]
  \centering
    \setlength{\tabcolsep}{2pt}
    \fontsize{8}{9}\selectfont
    \begin{tabular}{c|ccccc}
    \toprule
    \toprule
    \multicolumn{1}{c|}{\textbf{Settings}} & \multicolumn{1}{c}{\textbf{HotpotQA}} & \multicolumn{1}{c}{\textbf{2WikiMQA}} & \multicolumn{1}{c}{\textbf{MusiQue}} & \multicolumn{1}{c}{\textbf{ASQA}} & \multicolumn{1}{c}{\textbf{StrategyQA}} \\
    \midrule
    \rowcolor[rgb]{ .851,  .851,  .851}\multicolumn{6}{c}{\textit{Dataset statistics}}\\
    \multicolumn{1}{c|}{\#\,\,Samples used for evaluation} & \multicolumn{1}{c}{500} & \multicolumn{1}{c}{500} & \multicolumn{1}{c}{500} & \multicolumn{1}{c}{948} & \multicolumn{1}{c}{500} \\
    \midrule
    \rowcolor[rgb]{ .851,  .851,  .851}\multicolumn{6}{c}{\textit{Evaluation settings}}\\
    \multicolumn{1}{c|}{Metric} & \multicolumn{1}{c}{Accuracy,F1,EM} & \multicolumn{1}{c}{Accuracy,F1,EM} & \multicolumn{1}{c}{Accuracy,F1,EM} & \multicolumn{1}{c}{Accuracy,F1,EM} & \multicolumn{1}{c}{Accuracy} \\
    \midrule
    \rowcolor[rgb]{ .851,  .851,  .851}\multicolumn{6}{c}{\textit{Retrieval settings}}\\
    \multicolumn{1}{c|}{Corpus} & \multicolumn{1}{c}{\citealp{DBLP:conf/acl/TrivediBKS23}} & \multicolumn{1}{c}{\citealp{DBLP:conf/acl/TrivediBKS23}} & \multicolumn{1}{c}{\citealp{DBLP:conf/acl/TrivediBKS23}} & \multicolumn{1}{c}{Wikipedia-2018} 
    & \multicolumn{1}{c}{Wikipedia-2018} \\
    \#\,\,Documents in Corpus & 5233329 & 139416 & 430139 & 21015324 & 21015324 \\
    Retriever & BM25, Dense  & BM25, Dense & BM25,Dense& \multicolumn{1}{c}{Dense} & \multicolumn{1}{c}{Dense} \\
    Top-$k$ & 3,5,7 & 3,5,7 & 3,5,7 & 3,5,7 & 3,5,7 \\
    \bottomrule
    \bottomrule
    \end{tabular}%
    \caption{\textbf{All experimental settings.} We use bge-base-en-v1.5 as the dense retriever.}
  \label{tab-app:All Experimental Settings}%
\end{table*}%
\clearpage

\begin{table*}[ht!]
\begin{tcolorbox}[title={Prompt of the Note Initialization Process},coltitle=mypurple!5!white,colback=mypurple!5!white,colframe=mypurple!50!black]
{\bf Instructions}
\\
\\
Based on the provided document content, write a note.
\\
\\
The note should integrate all relevant information
from the original text that can help answer the specified question and form a coherent paragraph.
Please ensure that the note includes all original text information useful for answering the question.
\tcblower
Question to be answered: \texttt{\{query\}}\\
Document content: \texttt{\{refs\}}\\

Please provide the note you wrote:

\end{tcolorbox}
\caption{{\bf Prompt of the note initialization process.}}
\label{tab-app:Prompt of the Note Initialization Stage}
\end{table*}

\begin{table*}[ht!]
\begin{tcolorbox}[title={Prompt of the Query Refinement Stage},coltitle=mypurple!5!white,colback=mypurple!5!white,colframe=mypurple!50!black]
{\bf Instructions}
\\
\\
Task: Based on the notes, propose two new questions.
\\
\\
These new questions will be used to retrieve
documents to supplement the notes and help answer the original question.
The new questions should be
concise and include keywords that facilitate retrieval. The new questions should avoid duplication with
the existing question list.
\tcblower

Original question: \texttt{\{query\}} \\
Notes: \texttt{\{note\}} \\

Existing question list: \texttt{\{query\_log\}}\\

Please provide the note you wrote:

\end{tcolorbox}
\caption{{\bf Prompt of the query refinement stage.}}
\label{tab-app:Prompt of the Query Refinement Stage}
\end{table*}
\clearpage

\begin{table*}[ht!]
\begin{tcolorbox}[title={Prompt of the Knowledge Accumulation Stage},coltitle=mypurple!5!white,colback=mypurple!5!white,colframe=mypurple!50!black]
{\bf Instructions}
\\
\\
Task: Based on the retrieved documents, supplement the notes with content not yet included but useful for
answering the question.
\\
\\
The supplement should use the original text from the retrieved documents. The
added content should include as much information from the retrieved documents as possible.
\tcblower

Question: \texttt{\{query\}}\\
Retrieved document: \texttt{\{refs\}}\\
Notes: \texttt{\{note\}}\\

Please provide the note you wrote:\\

\end{tcolorbox}
\caption{{\bf Prompt of the knowledge accumulation stage.}}
\label{tab-app:Prompt of the Knowledge Accumulation Stage}
\end{table*}

\begin{table*}[ht!]
\begin{tcolorbox}[title={Prompt of the Adaptive Retrieval Decision Stage},coltitle=mypurple!5!white,colback=mypurple!5!white,colframe=mypurple!50!black]
{\bf Instructions}
\\
\\
Task: Please help me determine which note is better based on the following evaluation criteria:
\\
1. Contains key information directly related to the question.
\\
2. Completeness of Information: Does it cover all relevant aspects and details?
\\
3. Level of Detail: Does it provide enough detail to understand the issue in depth?
\\
4. Practicality: Does the note offer practical help and solutions?
\\
\\
Please make your judgment adhering strictly to the following rules:
\\
- If Note 2 does not add new meaningful content on top of Note 1, or only adds redundant information, return \texttt{json \{\{"status":"False"\}\}} directly.
\\
- If Note 2 has significant improvements over Note 1 based on the above criteria, return \texttt{json \{\{"status":"True"\}\}} directly; otherwise, return \texttt{json \{\{"status":"False"\}\}}.
\\
\tcblower
Question: \texttt{\{query\}}
\\
Provided Note 1: \texttt{\{best\_note\}}
\\
Provided Note 2: \texttt{\{new\_note\}}
\\
\\
Based on the above information, make your judgment without explanation and return the result directly.

\end{tcolorbox}
\caption{{\bf Prompt of the adaptive retrieval decision stage.}}
\label{tab-app:Prompt of the Adaptive Retrieval Decision Stage}
\end{table*}
\clearpage

\begin{table*}[ht!]
\begin{tcolorbox}[title={Prompt of the Note-Informed Answer Generation Process (\textbf{Multi-hop QA})},coltitle=lightgrey!50!white,colback=lightgrey!50!white,colframe=lightgrey!10!black]
{\bf Instructions}
\\
\\
Answer the question based on the given notes. Only give me the answer and do not output any other words.
\tcblower
The following are given notes:\\
\texttt{\{note\}}
\\
Question: \texttt{\{query\}}
\\
\\
Answer:
\end{tcolorbox}
\caption{{\bf Prompt of the note-informed answer generation process (multi-hop QA).}}
\label{Prompts of the Note-Informed Answer Generation Process (Multi-hop QA)}
\end{table*}

\begin{table*}[ht!]
\begin{tcolorbox}[title={Prompts of the Note-Informed Answer Generation Process (\textbf{ASQA})},coltitle=lightgrey!50!white,colback=lightgrey!50!white,colframe=lightgrey!10!black]
{\bf Instructions}
\\
\\
Write an accurate, engaging, and concise answer for the given question using only the provided notes. Use an unbiased and journalistic tone.
\tcblower
Question: \texttt{\{query\}}
\\
Notes: \texttt{\{note\}}
\end{tcolorbox}
\caption{{\bf Prompt of the note-informed answer generation process (ASQA).}}
\label{tab-app:Prompts of the Note-Informed Answer Generation Process (ASQA)}
\end{table*}

\begin{table*}[ht!]
\begin{tcolorbox}[title={Prompts of the Note-Informed Answer Generation Process (\textbf{StrategyQA})},coltitle=lightgrey!50!white,colback=lightgrey!50!white,colframe=lightgrey!10!black]
{\bf Instructions}
\\
\\
Answer the question based on the given notes.
\\
Only give me "yes" or "no" as your answer and do not output any other words.
\tcblower
The following are given notes: \texttt{\{note\}}
\\
Question: \texttt{\{query\}}
\\
\\
Answer:

\end{tcolorbox}
\caption{{\bf Prompt of the note-informed answer generation process (StrategyQA).}}
\label{tab-app:Prompts of the Prompts of the Note-Informed Answer Generation Process (StrategyQA)}
\end{table*}
\clearpage

\begin{table*}[ht!]
\begin{tcolorbox}[title={Prompts of Note Initialization Stage for DPO},coltitle=red!5!white,colback=red!5!white,colframe=red!10!black]
{\bf Instructions}
\\
\\
Task: You will receive a list of notes generated based on a given document content and question. 
Your task is to evaluate and score these notes based on their quality. Quality refers to: relevance, coherence, completeness in answering the specified question, and accuracy of information. 
\tcblower
Question to be answered: \{query\}
\\
Document content: \{refs\}
\\
Generated notes: \{notes\}
\\
Note format: Each note contains "\_id" and "content" fields. 
\\
\\
Evaluate the generated notes. The highest-scoring note must be factually correct based on the document. If no note is correct, or if there is minimal quality difference between notes, use the same \_id for both best and worst.
\\
\\
Output in the following JSON format:
\texttt{json \{\{"best\_id": <\_id of the highest-scoring note>, "worst\_id": <\_id of the lowest-scoring note>\}\}}
\\
\\
Do not include any explanations or additional text.

\end{tcolorbox}
\caption{{\bf Prompt of note initialization stage for DPO.}}
\label{tab-app:Prompts of Note Initialization Stage for DPO}
\end{table*}
\clearpage

\begin{table*}[ht!]
\begin{tcolorbox}[title={Prompts of Query Refinement Stage for DPO},coltitle=red!5!white,colback=red!5!white,colframe=red!10!black]
{\bf Instructions}
\\
\\
Task: You will receive a list of new questions generated based on some notes and an existing question list to supplement a given original question.
Your task is to evaluate these new questions based on their quality. Quality refers to: relevance, specificity, keyword richness, and non-redundancy. The goal is to identify questions that can retrieve useful information to help answer the original question.
\tcblower
Notes: \{notes\}
\\
\\
Original question: \{query\}
\\
\\
Existing question list: \{query\_log\}
\\
\\
New question list: \{new\_querys\}
\\
\\
Question format: Each question contains "\_id" and "content" fields. 
\\
\\
Evaluate the new question list. The highest-scoring new question must be able to help retrieve relevant information to answer the original question. If no new question can help get useful information, or if there is minimal quality difference between new questions, use the same \texttt{\_id} for both \texttt{best\_id} and \texttt{worst\_id}.
\\
\\
Output in the following format:
\texttt{json \{\{"best\_id": <\_id of the highest-scoring question>, "worst\_id": <\_id of the lowest-scoring question>\}\}}
\\
\\
Do not include any explanations or additional text.

\end{tcolorbox}
\caption{{\bf Prompt of query refinement stage for DPO.}}
\label{tab-app:Prompts of Query Refinement Stage for DPO}
\end{table*}
\clearpage

\begin{table*}[t!]
\renewcommand{\arraystretch}{1.2}
\setlength{\tabcolsep}{2pt}
    \fontsize{9}{10}\selectfont
    \centering
    \begin{tabular}{p{15cm}}
\toprule
\toprule
{\bf Question: Where was the place of death of Anna Of Pomerania's father?}
\\\bottomrule
{\bf Process of DeepNote:} \\
\ding{43} \, \textbf{Init note:} Unfortunately, the provided document content does not contain any information about Anna Of Pomerania's father or his place of death. The document content appears to be unrelated to Anna Of Pomerania and instead discusses various topics such as a British silent adventure film, the concept of "sins of the father," and the definitions of "place of birth" and "place of origin."
\\
\ding{43} \, {\bf New queries:}
1. Who was Anna Of Pomerania's father?\,\,\,\,2. What is known about the life and death of Anna Of Pomerania's father?\\
\ding{43} \, {\bf Updated note:}
Unfortunately, the provided document content \textcolor{myblue}{\bf does not contain any information about Anna Of Pomerania's father or his place of death.} $\dots$ Sophie of Pomerania, Duchess of Pomerania, we can learn that Sophie's father was Bogislaw IX, Duke of Pomerania, and that he died in 1446. $\dots$ It is worth noting that the documents provided do not mention Anna Of Pomerania at all, and the information about Sophie of Pomerania and her family is not directly related to Anna Of Pomerania $\dots$
\,\,\ding{220}\,\,\textbf{[Knowledge growth: False]}
\\
\ding{43} \, {\bf New queries:}
1. What is the name of Anna Of Pomerania's father and where did he die?\,\,\,\,2. Who was the parent of Anna Of Pomerania and what was his place of death?
\\
\ding{43} \, {\bf Updated note:}
Based on the retrieved documents, I was \textcolor{myblue}{\bf able to find some information related to Anna of Pomerania's family, but not directly about her father's place of death.}
$\dots$
From the document "Anna of Pomerania",
we know that \textcolor{myblue}{\hlgreen{\bf Anna of Pomerania's father was Bogislaw XIII, Duke of Pomerania}}. Unfortunately, the document does not mention his place of death.
$\dots$\,\,\ding{220}\,\,\textbf{[Knowledge growth: True]}
\\
\ding{43} \, {\bf New queries:}
1. What is known about the death of Bogislaw XIII, Duke of Pomerania?\,\,\,\,2. Where did Bogislaw XIII, Duke of Pomerania, die?
\\
\ding{43} \, {\bf Updated note:}
From the document "Bogislaw XIII, Duke of Pomerania", we know that \textcolor{myblue}{\hlgreen{\bf Anna of Pomerania's father was Bogislaw XIII, Duke of Pomerania.}} According to the document, \textcolor{myblue}{\hlgreen{\bf Bogislaw XIII died on 7 March 1606, in Stettin (also known as Szczecin in Polish).}}
$\dots$
Bogislaw XIII was a prince of Stettin and Wolgast, and a member of the Griffins. He was $\dots$\,\,\ding{220}\,\,\textbf{[Knowledge growth: True]}\\
\ding{43} \, {\bf Stop the iteration process and output the answer}
\\\midrule
{\bf Vanilla RAG:} \hlred{No information} is given in the passages about Anna Of Pomerania's father.\,\,\textcolor{red}{\ding{55}}
\\
{\bf Self-RAG:} 
\hlred{Not mentioned.}\,\,\textcolor{red}{\ding{55}}
\\\midrule
{\bf Answer of DeepNote:} \hlgreen{Stettin (also known as Szczecin in Polish)}\,\textcolor{green}{\ding{51}}
\\\midrule
{\bf Golden Answer:} Stettin
\\\bottomrule
\bottomrule
 \end{tabular}
    \caption{{\bf An example of DeepNote on the 2WikiMQA dataset ($\text{max step}=3, \text{max failure}=2$).}
    \textcolor{myblue}{\bf Bold blue text} represents all segments related to solving the problem (including both relevant and correct segments as well as relevant but incorrect segments). \hlgreen{Green segments} denote those used to support correct answers. \hlred{Red segments} represent key incorrect information and erroneous responses collected. The "\textcolor{green}{\ding{51}}" indicates a correct answer, while "\textcolor{red}{\ding{55}}" signifies an incorrect answer.
    }
    \label{tab-app: DeepNote examples on the 2WikiMQA dataset}
\end{table*}
\clearpage

\begin{table*}[t!]
\renewcommand{\arraystretch}{1.2}
\setlength{\tabcolsep}{2pt}
    \fontsize{9}{10}\selectfont
    \centering
    \begin{tabular}{p{15cm}}
\toprule
\toprule
{\bf Question: What city was the author of Neighbors born in?}
\\\bottomrule
{\bf Process of DeepNote:} \\
\ding{43} \, \textbf{Init note:} Unfortunately, the provided document content \textcolor{myblue}{\bf does not contain information about the birth city of the author of "Neighbors".} However, based on the available information, I can provide a note that includes the relevant details about the author of "Neighbors".
Note: \textcolor{myblue}{\hlgreen{\bf The author of the short story "Neighbors" is Raymond Carver.}} Unfortunately, the provided document content does not mention his birth city. However, it does provide information about the publication history of the short story, which was first published in "Esquire magazine" in 1971 and later included in several collections.
\\
\ding{43} \, {\bf New queries:}
1. What is the birth city of American short story writer Raymond Carver?\,\,\,\,2. Where was Raymond Carver, the author of the short story "Neighbors", born?\\
\ding{43} \, {\bf Updated note:}
Based on the retrieved documents, I have supplemented the notes with the following content: Note: \textcolor{myblue}{\hlgreen{\bf The author of the short story "Neighbors" is Raymond Carver.}} According to the available information, \textcolor{myblue}{\hlgreen{\bf Raymond Carver was born in Clatskanie, Oregon,}} a mill town on the Columbia River. \textcolor{myblue}{\bf This information is found in the document titled "Raymond Carver", which provides details about his birthplace and upbringing.} Although the document "Neighbors (short story)" does not mention his birth city, it does provide information about the publication history of the short story, which was first published in "Esquire magazine"$\dots$
\,\,\ding{220}\,\,\textbf{[Knowledge growth: True]}
\\
\ding{43} \, {\bf New queries:}
1. What is the birthplace of the American author of the short story collection "Will You Please Be Quiet, Please?"?\,\,\,\,2. In which city was the author of the short story "Neighbors" raised?
\\
\ding{43} \, {\bf Updated note:}
Based on the retrieved documents, I have supplemented the notes with the following content: Note: The author of the short story "Neighbors" is Raymond Carver. Unfortunately, the retrieved documents do not provide information about his birth city. However, the document "Neighbors (short story)" does provide information about the publication history of the short story, which was first published in "Esquire magazine" in 1971 and later included in several collections, including "Will You Please Be Quiet, Please?" in 1976, "Where I'm Calling From" in 1989, and "Short Cuts" in 1993. Since the retrieved documents \textcolor{myblue}{\bf do not provide information about Raymond Carver's birth city,} I was unable to find the answer to the question. However, I can suggest that further research may be needed to find the correct answer. \,\,\ding{220}\,\,\textbf{[Knowledge growth: False]}\\
\ding{43} \, {\bf Stop the iteration process and output the answer}
\\\midrule
{\bf Vanilla RAG:} \hlred{Not mentioned}\,\,\textcolor{red}{\ding{55}}
\\
{\bf ReAct:} 
\hlred{Raymond Carver}\,\,\textcolor{red}{\ding{55}}
\\\midrule
{\bf Answer of DeepNote:} \hlgreen{Clatskanie}\,\textcolor{green}{\ding{51}}
\\\midrule
{\bf Golden Answer:} Clatskanie
\\\bottomrule
\bottomrule
 \end{tabular}
    \caption{{\bf An example of DeepNote on the MusiQue dataset ($\text{max step}=3, \text{max failure}=1$)}.
    \textcolor{myblue}{\bf Bold blue text} represents all segments related to solving the problem (including both relevant and correct segments as well as relevant but incorrect segments). \hlgreen{Green segments} denote those used to support correct answers. \hlred{Red segments} represent key incorrect information and erroneous responses collected. The "\textcolor{green}{\ding{51}}" indicates a correct answer, while "\textcolor{red}{\ding{55}}" signifies an incorrect answer.
    }
    \label{tab-app: DeepNote examples on the MusiQue dataset}
\end{table*}
\clearpage

\begin{table*}[t!]
\renewcommand{\arraystretch}{1.2}
\setlength{\tabcolsep}{2pt}
    \fontsize{9}{10}\selectfont
    \centering
    \begin{tabular}{p{15cm}}
\toprule
\toprule
{\bf Question: A man who played in the 1986 FIFA world cup played for what team during the 1982 Scottish League Cup Final?}
\\\bottomrule
{\bf Process of DeepNote:} \\
\ding{43} \, \textbf{Init note:} Unfortunately, the provided document content \textcolor{myblue}{\bf does not contain information about a specific player who played in the 1986 FIFA World Cup and also played in the 1982 Scottish League Cup Final.} However, based on the available information, we can make an educated guess. Labid Khalifa played in the 1986 FIFA World Cup, but there is no information about him playing in the 1982 Scottish League Cup Final. On the other hand, \textcolor{myblue}{\bf Hamish McAlpine played for }\textcolor{myblue}{\hlred{\bf Dundee United}} for twenty years, \textcolor{myblue}{\bf between 1966 and 1986, and won the Scottish League Cup twice}, \textcolor{myblue}{\bf but it is not specified if he played in the 1986 FIFA World Cup.} However, since he played for Dundee United until 1986, it is possible that he played in the 1986 FIFA World Cup, but this \textcolor{myblue}{\bf is not confirmed}. If we assume that Hamish McAlpine played in the 1986 FIFA World Cup, then based on the information provided, he played for Dundee United during the 1982 Scottish League Cup Final.
\\
\ding{43} \, {\bf New queries:}
1. Which players from the 1986 FIFA World Cup participated in the 1982 Scottish League Cup Final?\,\,\,\,2. Who were the Dundee United players that played in the 1986 FIFA World Cup?\\
\ding{43} \, {\bf Updated note:}
Based on the retrieved documents, I have supplemented the notes with content not yet included but useful for answering the question. Unfortunately, the provided document content \textcolor{myblue}{\bf still does not contain direct information about a specific player} who played in the 1986 FIFA World Cup and also played in the 1982 Scottish League Cup Final. However, I have added more information from the retrieved documents to provide a more comprehensive answer.
\textcolor{myblue}{\bf Labid Khalifa played in the 1986 FIFA World Cup, but there is no information about him playing in the 1982 Scottish League Cup Final.} On the other hand, \textcolor{myblue}{\bf Hamish McAlpine played for Dundee United} for twenty years, \textcolor{myblue}{\bf between 1966 and 1986}, and won the Scottish League Cup twice, but it is \textcolor{myblue}{\bf not specified if he played in the 1986 FIFA World Cup.} However, since he played for Dundee United until 1986, it is possible that he played in the 1986 FIFA World Cup, but this is not confirmed. \textcolor{myblue}{\bf Additionally, the 1986 FIFA World Cup squads document may contain information about players who played in the 1986 FIFA World Cup, but the provided text does not specify the players.} If we had access to the full squads document, we might be able to find a player who played in the 1986 FIFA World Cup and also played in the 1982 Scottish League Cup Final. $\dots$ If we assume that Hamish McAlpine played in the 1986 FIFA World Cup, then based on the information provided, he played for Dundee United during the 1982 Scottish League Cup Final. \textcolor{myblue}{\bf However, this is still a speculative answer, and we would need more information to confirm it.}$\dots$
\,\,\ding{220}\,\,\textbf{[Knowledge growth: False]}
\\
\ding{43} \, {\bf New queries:}
1. Which players participated in both the 1986 FIFA World Cup and the 1982 Scottish League Cup Final for Dundee United?\,\,\,\,2. Did Hamish McAlpine play for Scotland in the 1986 FIFA World Cup?
\\
\ding{43} \, {\bf Updated note:}
Based on the retrieved documents, the supplemented notes are as follows: Unfortunately, the provided document content \textcolor{myblue}{\bf does not contain information about a specific player} who played in the 1986 FIFA World Cup and also played in the 1982 Scottish League Cup Final. However, \textcolor{myblue}{\bf based on the available information}, we can make an educated guess. Labid Khalifa played in the 1986 FIFA World Cup, but there is no information about him playing in the 1982 Scottish League Cup Final. On the other hand, Hamish McAlpine played for Dundee United for twenty years, between 1966 and 1986, and won the Scottish League Cup twice, \textcolor{myblue}{\bf but it is not specified if he played in the 1986 FIFA World Cup.} However, \textcolor{myblue}{\bf since he played for} \textcolor{myblue}{\hlred{\bf Dundee United}} until 1986, it is \textcolor{myblue}{\bf possible that he played in the 1986 FIFA World Cup, but this is not confirmed.} As per the document "Hamish McAlpine", it is mentioned that "He played for Dundee United for twenty years, between 1966 and 1986. Whilst with the club he won the Scottish Football League title once and the Scottish League Cup twice." Although it does not specifically mention the 1982 Scottish League Cup Final, it does mention that he won the Scottish League Cup twice, which suggests that he might have played in the 1982 Scottish League Cup Final. $\dots$ It is worth noting that \textcolor{myblue}{\bf the document "1986 FIFA World Cup squads" could potentially provide more information} about the players who played in the 1986 FIFA World Cup, \textcolor{myblue}{\bf but unfortunately, the content of this document is not available.} \,\,\ding{220}\,\,\textbf{[Knowledge growth: False]}\\
\ding{43} \, {\bf Stop the iteration process and output the answer}
\\\midrule
{\bf Answer of DeepNote:} \hlred{Dundee United}\,\textcolor{red}{\ding{55}}
\\\midrule
{\bf Golden Answer:} Celtic
\\\bottomrule
\bottomrule
 \end{tabular}
    \caption{{\bf Badcase analysis of DeepNote on the HotpotQA dataset ($\text{max step}=3, \text{max failure}=2$)}.
    \textcolor{myblue}{\bf Bold blue text} represents all segments related to solving the problem (including both relevant and correct segments as well as relevant but incorrect segments). \hlgreen{Green segments} denote those used to support correct answers. \hlred{Red segments} represent key incorrect information and erroneous responses collected. The "\textcolor{green}{\ding{51}}" indicates a correct answer, while "\textcolor{red}{\ding{55}}" signifies an incorrect answer.
    }
    \label{tab-app: Badcase Analysis of DeepNote on the HotpotQA dataset}
\end{table*}
\clearpage

\end{document}